\documentclass[10pt]{article}
\usepackage[margin=0.75in]{geometry}
\usepackage{graphicx}
\usepackage{amsmath}
\usepackage{amssymb}
\usepackage{mathtools}
\usepackage{bm}
\usepackage{tikz, pgfplots}
\usetikzlibrary{fit, positioning, calc}
\usetikzlibrary{shapes.geometric, arrows.meta, positioning}
\usetikzlibrary{positioning}
\usepackage{subcaption}
\usepackage{xcolor}
\definecolor{BrightBlue}{HTML}{007FFF}
\usepackage{url}
\usepackage{booktabs}
\usepackage{multicol}
\usepackage[square,numbers]{natbib}
\usepackage[colorlinks=true,linkcolor=blue,citecolor=blue]{hyperref}
\usepackage{float}
\usepackage[linesnumbered,ruled,noline]{algorithm2e}
\SetKwFor{For}{for}{do}{end for}

\def\bv{\bm{v}}
\def\by{\bm{y}}
\def\bz{\bm{z}}
\def\bw{\bm{w}}
\def\bphi{\bm{\phi}}
\def\bmu{\bm{\mu}}
\def\bsigma{\bm{\sigma}}
\def\bepsilon{\bm{\epsilon}}
\DeclareMathOperator*{\argmin}{argmin} 
\DeclareMathOperator*{\argmax}{argmax}

\title{Model synthesis and identifiability analysis of stiff chemical reaction systems with inVAErt networks}
\author{Sreejata Dey$^{1}$, Guoxiang Grayson Tong$^{2}$, 
Jonathan~F.~MacArt$^{3}$, Daniele~E.~Schiavazzi$^{1}$} 
\date{\small
$^{1}$ Department of Applied and Computational Mathematics and Statistics, 
University of Notre Dame\\
$^{2}$ Department of Pediatrics, Stanford University\\
$^{3}$ Department of Aerospace and Mechanical Engineering, University of Notre Dame\\
}

\begin{document}

\maketitle

\begin{abstract}
\noindent We consider the problem of learning data-driven replicas for stiff systems of ordinary differential equations arising in chemical kinetics that can be evaluated with high computational efficiency. 
We first focus on training emulators for families of reaction equations under varying reaction rates, using conditional residual networks or long-short term memory architectures. 
We then apply a recently proposed data-driven framework known as ``inVAErt networks'' to address the ill-posed inverse problem of inferring reaction rates, integration time, and possibly initial conditions from a target set of species concentrations - a problem that has received relatively little attention in the literature. 
The proposed approach is demonstrated on chemical systems with reversible and irreversible kinetics, spanning 2 to 20 differential equations, 3 to 20 chemical species, and 3 to 25 reaction rate parameters.
Relative root mean squared errors produced by the proposed emulators range from $10^{-5}$ for lower-dimensional systems to $10^{-4}$ and $10^{-3}$ for an air pollution model and a hydrogen-air reaction system, respectively. Manifolds of non-identifiable reaction rates recovered by the proposed approach can be analytically verified for simple systems and are consistent with local identifiability analysis in higher dimensions.
\end{abstract}

% =====================================
\section{Introduction}\label{sec:intro}
% =====================================

%
The widely disparate time scales induced by common physical models of chemical kinetics
remains the primary computational bottleneck in reactive flow simulations, motivating extensive research into neural network-based acceleration or imputation methods. In spatially homogeneous systems, these disparate scales induce stiff systems of ordinary differential equations (ODEs).
Four major methodological frameworks have emerged over the past decade to solve such systems: Neural Ordinary Differential Equations (Neural ODEs), Physics-Informed Neural Networks (PINNs), operator learning methods, and graph-based architectures. 

Neural ODEs, exemplified by ChemNODE~\cite{owoyele2022chemnode} and the foundational work on stiff neural ODEs in~\cite{kim2021stiff}, integrate neural network predictions during training to ensure stability when coupled with ODE solvers, achieving speedups up to factors of 70 to 100 on hydrogen-air combustion systems. 
Physics-constrained extensions~\cite{kumar2024physics} incorporate elemental mass conservation directly into loss functions, improving long-time integration stability. PINNs use automatic differentiation for weakly enforcing satisfaction of the governing equations in the loss function, providing a regularization mechanism that enables learning even under \emph{limited data} regimes. 
However, in their original formulation, PINNs fail on stiff systems due to extreme scale separations between species concentrations~\cite{ji2021stiff}, and approaches such as Stiff-PINN~\cite{ji2021stiff} or Multiscale PINNs (MPINNs~\cite{weng2022multiscale}) were developed to separate fast from slow timescales in the evolution of species concentrations. 

Surrogate modeling approaches, which have evolved significantly since early attempts in the 1990s~\cite{christo1996,blasco1998}, now emphasize physical consistency as a critical requirement. 
Recent work demonstrates that traditional error metrics can be misleading -- networks with low training errors may still violate conservation laws and diverge during long-time integration~\cite{wang2025enforcing}. 
Use of structural (through the network architecture) rather than weak (e.g., loss augmentation) constraints proves essential for robust predictions. 
Multi-scale sampling strategies~\cite{zhang2022multiscale} address data imbalance by targeting underrepresented transient regions, while innovative approaches like learning rates of rate-determining steps~\cite{doppel2022efficient} achieve greater accuracy up to factors of 50 to 200 than conventional methods with minimal training data. 
Recent advances include Kolmogorov-Arnold Networks (ChemKANs~\cite{koenig2025chemkans}), which leverage chemistry-specific information flow and achieve speedups of a factor of 40 with strong noise robustness, and global reaction neural networks~\cite{kircher2023global} that embed thermodynamic and stoichiometric constraints for autonomous kinetic model learning from reactor data.
\citet{vijayarangan2024data} uses a data-driven reduced order model for stiff chemical reaction systems combining a nonlinear autoencoder with a neural ODE to learn a latent space that inherently removes stiffness, enabling larger integration time steps than physics-based alternatives.

Operator learning methods, particularly Deep Operator Networks (DeepONets~\cite{goswami2024learning,echekki2024combustion}) and Fourier Neural Operators (FNO~\cite{weng2025extended}), represent a paradigm shift by learning solution operators that map initial conditions to time-evolved states rather than learning point-wise functions. 
Extended DeepONets achieve speedups up to factors of 100 to 1000 by combining autoencoder-based dimensionality reduction with operator learning in latent space, exploiting the manifold structure of combustion chemistry. 
The Adaptive Multi-Output Operator Network (AMORE) framework~\cite{jagtap2024amore} advances scalability by using a single operator for all thermochemical variables with adaptive loss functions and automatic mass conservation. 
Graph neural networks provide complementary capabilities through atom-conserving Chemical Reaction Neural Networks (CRNNs) and condensed graph of reaction (CGR) representations~\cite{schwaller2019molecular,heid2022machine}, enabling mechanism discovery and interpretable kinetic modeling. 

In this study, in addition to the focus of the cited references on \emph{emulation}, i.e. determining the time-varying concentrations from known reaction rate parameters, we also consider the less-investigated problem of estimating the reaction rate parameters in an \emph{amortized} fashion (i.e., without retraining our inference engine) from multiple sets of concentrations. Additionally, unlike many of the previous studies, we learn \emph{families of admissible solutions}, under uniformly distributed priors of kinetic rate constants.
Inference is important since not all species can be measured experimentally: many intermediate species are either (1) too short-lived to be detected, (2) below detection limits of available instrumentation, (3) inaccessible without perturbing the system, or (4) require specialized, expensive equipment not routinely available~\cite{gasparyan2023parameter}.
Additionally, intermediate species play a crucial role in combustion chemistry, but their molecules are typically present at an extremely low number density. These intermediates are critical for validating mechanisms, yet comprehensive measurement is impossible. Researchers must infer unmeasured reaction rates from the limited species they can observe~\cite{kiefer2011laser}. 
Finally, estimation of reaction rate constants is important to formulate hypotheses on the underlying chemical reaction mechanisms, and to reduce the stiffness in the chemical mechanism producing known concentrations.
In this study, we demonstrate the effectiveness of inVAErt networks, a recently proposed data-driven architecture approach for model synthesis and identifiability analysis to learn the forward and inverse response of hard-to-solve, stiff dynamical systems. 

The paper is organized as follows. Section~\ref{sec:chem_odes} introduces the chemical reaction systems we choose as the focus of our investigation.
Specifically, we discuss a characteristic solution, equilibrium and stability of the Robertson problem in Section~\ref{subsec:rober}, and for the POLLU problem in Section~\ref{subsec:pollu}.
We then investigate two systems characterized by reversible and irreversible kinetics, respectively, in Section~\ref{subsec:rev-irr}, and a hydrogen-air reaction system in Section~\ref{sec:hydro-air}.
Training data is generated first by numerical integration, followed by re-sampling. In Section~\ref{sec:data_gen} we discuss three re-sampling strategies whose combination allows us to capture the initial transient, sudden changes of pressure and temperature, and long-run steady concentrations. 
Section~\ref{sec:invaert} introduces inVAErt networks, the workhorse for model synthesis and parameters estimation in this study, discussing two possible data-driven surrogates for the forward problem, i.e., a residual and a LSTM-type network.
Results are presented in Section~\ref{subsec:results}, first for the emulation task (Section~\ref{sec:emu_res}), followed by inference of rate constants (Section~\ref{sec:inference}).
Finally, we discuss the results from this study and offer some conclusions in Section~\ref{sec:discussion}.

% ===================================================
\section{Chemical reaction ODEs}\label{sec:chem_odes}
% ===================================================

The \emph{law of mass action}~\cite{law-of-mass-action, law2006combustion} provides a standard framework for modeling chemical reaction systems as dynamical systems, and hereby acts as a bridge between the chemistry and a mathematical model. 
All systems considered in this study were formalized accordingly. The law of mass action models the system with \emph{rate constants} as parameters which govern the dynamics (as well as the stiffness) of the reaction system. The chemical concentrations at the beginning of the reaction serve as the initial conditions of the dynamical system.

Chemical reaction systems when modeled as dynamical systems often present themselves as stiff systems, a property that arises from the presence of widely separated time scales.
In practice, the Arrhenius rate constant parametrization contributes to the stiffness by introducing a dependence on temperature, which dynamically amplifies reaction rates at events such as ignition.
Thus, the central focus of this work is to characterize the stiffness of such systems and to investigate its effects on both the resulting dynamics and the identifiability of rate constants.
We have worked with benchmark problems that are widely used in the stiff ODEs literature -- we begin with the Robertson problem and subsequently investigate the POLLU problem, which presents a higher dimensional and more challenging test case.
We additionally investigate emulation of identifiability in cases where the initial conditions are included as system parameters. 
To this end, we have considered two systems characterized by a reversible and irreversible kinetics, respectively.
Finally, we test our setup with a realistic problem involving a hydrogen-air reaction mechanism.

% ================================================
\subsection{Robertson problem}\label{subsec:rober}
% ================================================

\begin{figure}[!ht]
\centering
\begin{subfigure}{0.30\textwidth}
    \centering
    \includegraphics[width=\textwidth]{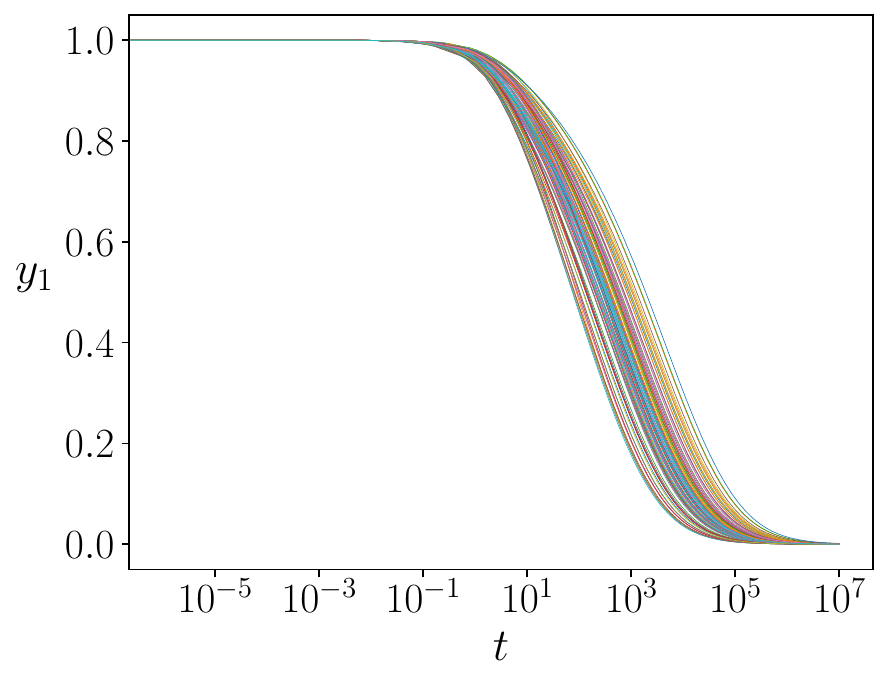}
\end{subfigure}
\begin{subfigure}{0.35\textwidth}
    \centering
    \includegraphics[width=\textwidth]{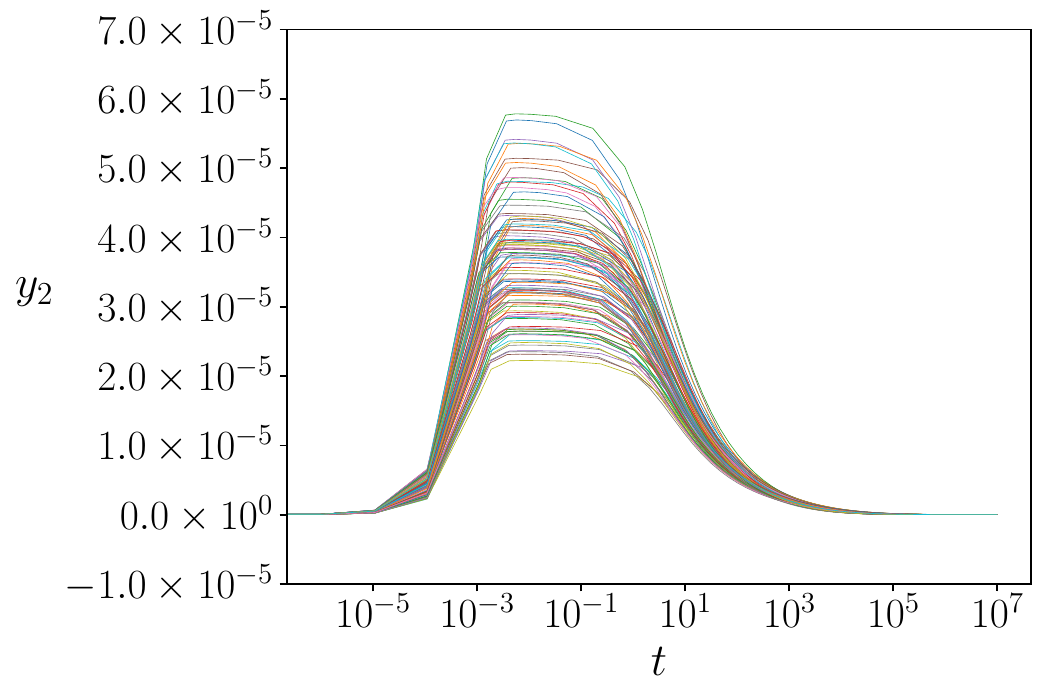}
\end{subfigure}
\begin{subfigure}{0.30\textwidth}
    \centering
    \includegraphics[width=\textwidth]{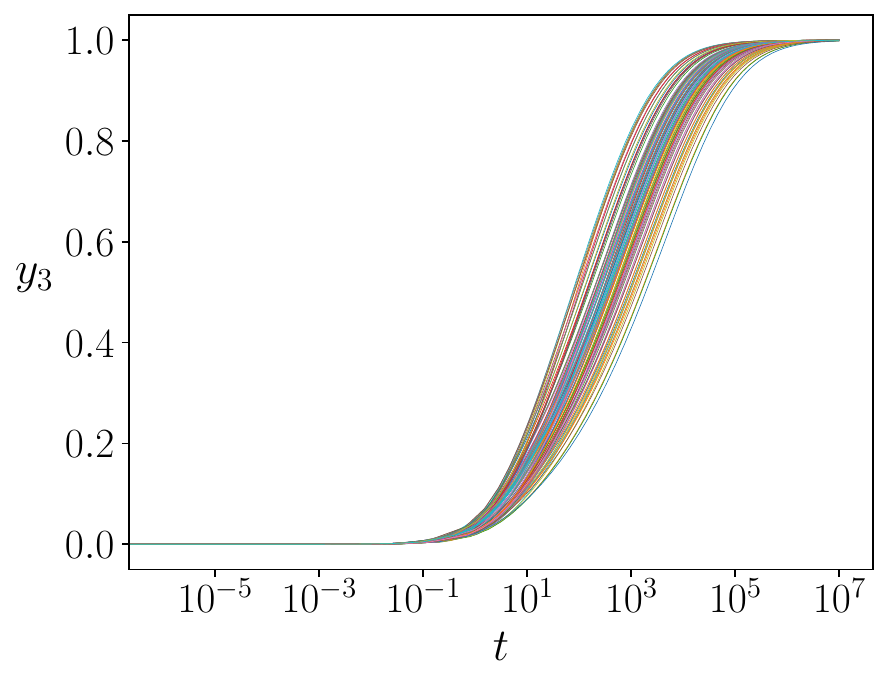}
\end{subfigure}
\caption{Solution plots for the Robertson problem: parameters perturbed 100 times in range $\pm 50\%$ of nominal parameters.}\label{fig:rober_perturbed}
\end{figure}

The problem consists of a stiff system of 3 non-linear ODEs, first proposed by Robertson~\cite{rober} and named \emph{ROBER} by Hairer \& Wanner~\cite{wanner1996solving}.
The chemical reaction system is parametrized in terms of the reaction rate constants $\{k_1, k_2, k_3\}$. If the time-varying mass fractions
% concentrations
of chemicals $A, B, C$ are denoted by $y_1(t), y_2(t), y_3(t)$ respectively for $t\in[0,10^2]$, then, by the law of mass action~\cite{law-of-mass-action}, we may describe the time evolution of this system as follows
\begin{equation}\label{equ:rober}
\begin{aligned}
A &\xrightarrow{k_1} B 
& & \qquad \frac{dy_1}{dt} = -k_1 y_1 + k_3 y_2 y_3, \\[0.5em]
B + B &\xrightarrow{k_2} B + C 
& \qquad \boldsymbol{\to} \qquad & \qquad \frac{dy_2}{dt} = k_1 y_1 - k_2 y_2^2 - k_3 y_2 y_3, \\[0.5em]
B + C &\xrightarrow{k_3} A + C 
& &\qquad \frac{dy_3}{dt} = k_2 y_2^2,
\end{aligned}
\end{equation}
where initial values are assumed equal to 
\begin{equation}\label{equ:rober_ic}
y_1(t=0)=1,\,\, y_2(t=0)=0,\,\, y_3(t=0)=0.
\end{equation}
Notice that $d(y_1(t)+y_2(t)+y_3(t))/dt= 0$, so $y_1(t)+y_2(t)+y_3(t)=y_1(0)+y_2(0)+y_3(0)=1$ for all $t$. Additionally, nominal values for the reaction rate parameters are assumed equal to 
\begin{equation}\label{equ:rober_nominal_k}
k_1=4\times10^{-2}, k_2=3\times 10^{7}, k_3=1\times 10^{4}.
\end{equation}
Figure~\ref{fig:rober_perturbed} shows the same quantities when varying the reaction rate constants within a $\pm 50 \%$ range of their nominal values in \eqref{equ:rober_nominal_k}.

If the Robertson system is written in the form $d\bm{y}/dt=\bm{\mathcal{F}}(\bm{y})$, the stiffness for this system can be analytically computed from the Jacobian matrix $\bm{J}=\partial \bm{\mathcal{F}} / \partial \bm{y}$ where
\[
\bm{J} = \begin{bmatrix}
-k_1 & k_3 y_3(t) & k_3 y_2(t) \\
k_1 & -2 k_2 y_2(t) - k_3 y_3(t) & -k_3 y_2(t) \\
0 & 2 k_2 y_2(t) & 0
\end{bmatrix},
\] and the dependence on time is made explicit for $y_1, y_2, y_3$. The eigenvalues of $\bm{J}$ are
\[
\begin{split}
\lambda_1=0,\,\,&\lambda_2=\frac{1}{2} \left( -k_1 - 2 k_2 y_2 - k_3 y_3 + \sqrt{-4 (2 k_1 k_2 y_2 + 2 k_2 k_3 y_2^2) + (k_1 + 2 k_2 y_2 + k_3 y_3)^2} \right),\\
& \text{ and } \lambda_3=\frac{1}{2} \left( -k_1 - 2 k_2 y_2 - k_3 y_3 - \sqrt{-4 (2 k_1 k_2 y_2 + 2 k_2 k_3 y_2^2) + (k_1 + 2 k_2 y_2 + k_3 y_3)^2} \right).
\end{split}
\]
Under the initial conditions \eqref{equ:rober_ic}, both nonzero eigenvalues $\lambda_2,\lambda_3$ are negative in sign. Plots of these eigenvalues versus time show that one decays in magnitude to 0 while the other grows in magnitude asymptotically to $-k_1-k_3 \approx -10^4$. Equilibrium is attained at $y_1=0, y_2=0, y_3=1$ as all mass eventually transfers to the monotonically increasing \emph{sink} $y_3$, with eigenvalues at equilibrium equal to $(0, 0, -k_1-k_3)$.

\begin{figure}[!ht]
\centering
\begin{subfigure}[t]{0.31\textwidth}
    \centering
    \includegraphics[width=\textwidth]{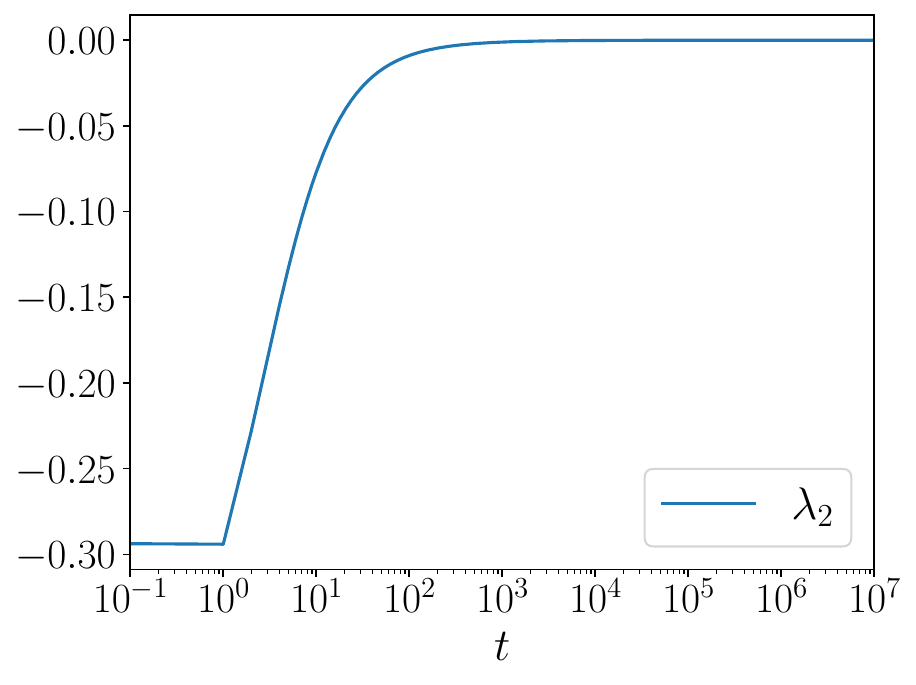}
    \caption{Non-zero eigenvalue with minimum magnitude $(\lambda_2)$ vs. time.}
\end{subfigure}
\begin{subfigure}[t]{0.32\textwidth}
    \centering
    \includegraphics[width=\textwidth]{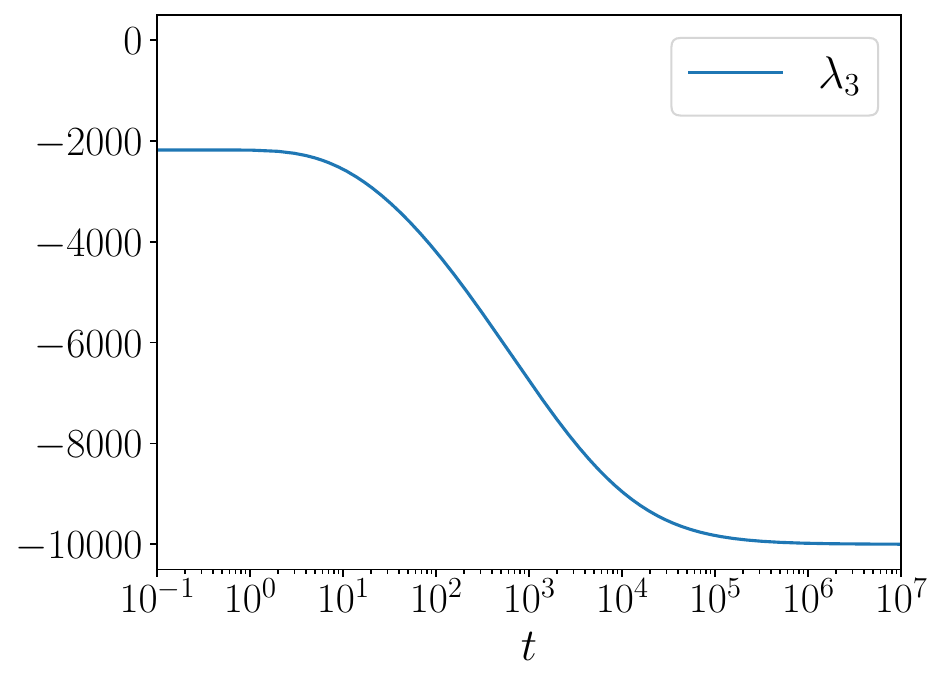}
    \caption{Non-zero eigenvalue with maximum magnitude $(\lambda_3)$ vs. time.}
\end{subfigure}
\begin{subfigure}[t]{0.30\textwidth}
    \centering
    \includegraphics[width=\textwidth]{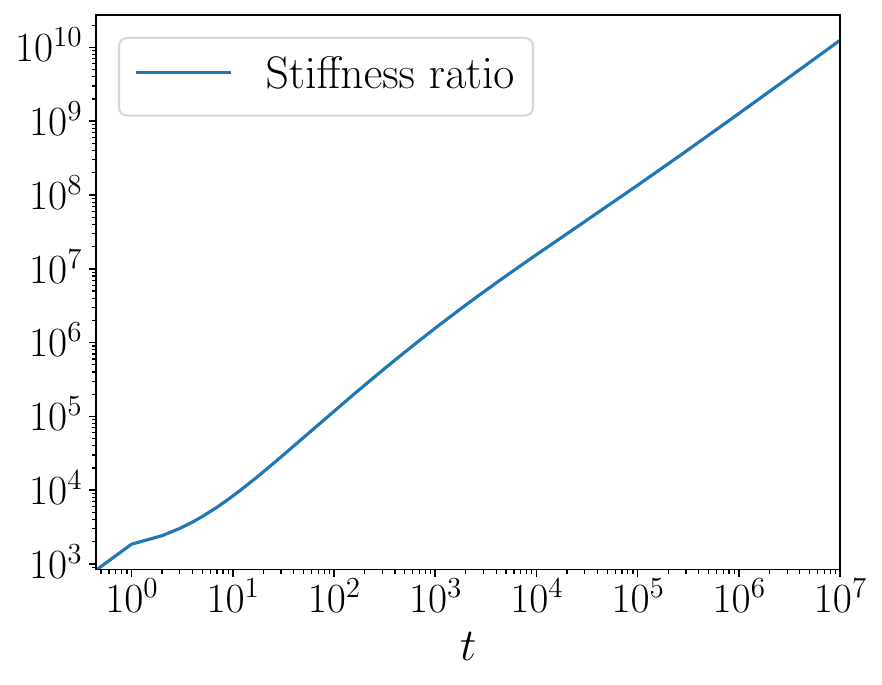}
    \caption{Stiffness ratio vs. time plot: $R$ keeps increasing as $\lambda_2$ approaches 0 while $\lambda_3$ approaches -10000.04.}
\end{subfigure}
\caption{Eigenvalues and stiffness ratio for the Robertson problem.}
\label{fig:rober_stiff}
\end{figure}

The stiffness ratio is defined as $R=\text{max}(\text{Re}(\lambda))/\text{min}(\text{Re}(\lambda))$ where $\lambda$ is a non-zero eigenvalue of the Jacobian matrix. 
The time evolution of $\lambda_{2}$, $\lambda_{3}$ and the stiffness ratio $R$ are shown in Figure~\ref{fig:rober_stiff}. 
$R$ grows as $t$ increases, and $\lambda_2 \rightarrow 0$. Since the eigenvalues differ from each other considerably in scale, the resulting system is very stiff, which is reflected in the value of $R$. As a consequence, if we were to use a constant time-step numerical solver, very small time steps would be required to resolve steep gradients in the Robertson system's response. Thus, only numerical integrators with adaptive time-stepping are able to solve this system efficiently, maintaining a relative coarse underlying time discretization.

% ============================================
\subsection{POLLU problem}\label{subsec:pollu}
% ============================================

The POLLU problem is a stiff initial value problem with 20 ODEs describing the evolution of 20 chemical species that evolve under the influence of 25 chemical reactions.
It is part of an air pollution model discussed by Verwer in~\cite{verwer1994gauss}. 
The problem is of the form
\[
\frac{d\bm{y}}{dt}=\bm{\mathcal{F}}(\bm{y}),\,\,\bm{y}(0)=\bm{y}_0,
\]
on $t\in[0,60]$ where $\bm{y} = [y_{1},\dots,y_{20}]^{T}\in \mathbb{R}^{20}$, and the initial condition is defined as 
\begin{equation}\label{equ:pollu_ic}
\bm{y}_0 = [0, 0.2, 0, 0.04, 0, 0, 0.1, 0.3, 0.01, 0, 0, 0, 0, 0, 0, 0, 0.007, 0, 0, 0]^T .
\end{equation}
Consider $R_{1} = \{1, 10, 14, 23, 24\}$, and $R_{2} = \{2, 3, 9, 11, 12, 22, 25\}$, the function $\bm{\mathcal{F}}:\mathbb{R}^{20}\to\mathbb{R}^{20}$ is defined as

\vspace{3pt}

\begin{tabular}{p{5.3cm} p{5.5cm} p{5.7cm}}
$\mathcal{F}_{1}(\bm{y}) = - \sum_{j\in R_{1}} r_j + \sum_{j\in R_{2}} r_j$, & $\mathcal{F}_{2}(\bm{y}) = -r_2 -r_3 -r_9 -r_{12} + r_1 + r_{21}$ & $\mathcal{F}_{3}(\bm{y}) = -r_{15} + r_1 + r_{17} + r_{19} + r_{22}$,\\
$\mathcal{F}_{4}(\bm{y}) = -r_2 - r_{16} - r_{17} - r_{23} + r_{15}$, & $\mathcal{F}_{5}(\bm{y}) = -r_3 + 2r_4 + r_6 + r_7 + r_{13} + r_{20}$, & $\mathcal{F}_{6}(\bm{y}) = -r_6 - r_8 - r_{14} - r_{20} + r_3 + 2r_{18}$,\\
$\mathcal{F}_{7}(\bm{y}) = -r_4 - r_5 - r_6 + r_{13}$, & $\mathcal{F}_{8}(\bm{y}) = r_4 + r_5 + r_6 + r_7$, & $\mathcal{F}_{9}(\bm{y}) = -r_7 -r_8$,\\
$\mathcal{F}_{10}(\bm{y}) = -r_{12} + r_7 + r_9$, & $\mathcal{F}_{11}(\bm{y}) = -r_9 - r_{10} + r_8 + r_{11}$, & $\mathcal{F}_{12}(\bm{y}) = r_9$,\\
$\mathcal{F}_{13}(\bm{y}) = -r_{11} +r_{10}$, & $\mathcal{F}_{14}(\bm{y}) = -r_{13} +r_{12}$, & $\mathcal{F}_{15}(\bm{y}) = r_{14}$,\\
$\mathcal{F}_{16}(\bm{y}) = -r_{18} -r_{19} + r_{16}$, & $\mathcal{F}_{17}(\bm{y}) = -r_{20}$, & $\mathcal{F}_{18}(\bm{y}) = r_{20}$,\\
$\mathcal{F}_{19}(\bm{y}) = -r_{21} - r_{22} -r_{24} +r_{23} + r_{25}$, & $\mathcal{F}_{20}(\bm{y}) = -r_{25} + r_{24}$.
\end{tabular}
with coefficients $r_{j},\,i=1\dots,25$ defined in Table~\ref{tab:pollu_coeffs}.

We present the time evolution of a subset of the chemical species resulting from uniformly sampling the reaction rate constants independently in an interval obtained using $\pm 50 \%$ of the nominal values in Table~\ref{tab:params_pollu}.

\begin{table}[H]
\centering

\begin{tabular}{| c c | c c | c c | c c | c c |}
\hline
\textbf{$\bm{r_i}$} & \textbf{Definition} &
\textbf{$\bm{r_i}$} & \textbf{Definition} &
\textbf{$\bm{r_i}$} & \textbf{Definition} &
\textbf{$\bm{r_i}$} & \textbf{Definition} &
\textbf{$\bm{r_i}$} & \textbf{Definition} \\
\hline

$r_1$  & $k_1 y_1$ &
$r_6$  & $k_6 y_7 y_6$ &
$r_{11}$ & $k_{11} y_{13}$ &
$r_{16}$ & $k_{16} y_4$ &
$r_{21}$ & $k_{21} y_{19}$ \\

$r_2$  & $k_2 y_2 y_4$ &
$r_7$  & $k_7 y_9$ &
$r_{12}$ & $k_{12} y_{10} y_2$ &
$r_{17}$ & $k_{17} y_4$ &
$r_{22}$ & $k_{22} y_{19}$ \\

$r_3$  & $k_3 y_5 y_2$ &
$r_8$  & $k_8 y_9 y_6$ &
$r_{13}$ & $k_{13} y_{14}$ &
$r_{18}$ & $k_{18} y_{16}$ &
$r_{23}$ & $k_{23} y_1 y_4$ \\

$r_4$  & $k_4 y_7$ &
$r_9$  & $k_9 y_{11} y_2$ &
$r_{14}$ & $k_{14} y_1 y_6$ &
$r_{19}$ & $k_{19} y_{16}$ &
$r_{24}$ & $k_{24} y_{19} y_1$ \\

$r_5$  & $k_5 y_7$ &
$r_{10}$ & $k_{10} y_{11} y_1$ &
$r_{15}$ & $k_{15} y_3$ &
$r_{20}$ & $k_{20} y_{17} y_6$ &
$r_{25}$ & $k_{25} y_{20}$ \\

\hline
\end{tabular}
\caption{Contributions to the right-hand-side for the POLLU system.}
\label{tab:pollu_coeffs}
\end{table}

\begin{table}[H]
\centering

\begin{tabular}{| c c | c c | c c | c c | c c |}
\hline
$\bm{k_i}$ & \textbf{Value} &
$\bm{k_i}$ & \textbf{Value} &
$\bm{k_i}$ & \textbf{Value} &
$\bm{k_i}$ & \textbf{Value} &
$\bm{k_i}$ & \textbf{Value} \\
\hline

$k_1$  & $0.350$ &
$k_6$  & $0.150 \times 10^5$ &
$k_{11}$ & $0.220 \times 10^{-1}$ &
$k_{16}$ & $0.350 \times 10^{-3}$ &
$k_{21}$ & $0.210 \times 10^1$ \\

$k_2$  & $0.266 \times 10^2$ &
$k_7$  & $0.130 \times 10^{-3}$ &
$k_{12}$ & $0.120 \times 10^5$ &
$k_{17}$ & $0.175 \times 10^{-1}$ &
$k_{22}$ & $0.578 \times 10^1$ \\

$k_3$  & $0.123 \times 10^5$ &
$k_8$  & $0.240 \times 10^5$ &
$k_{13}$ & $0.188 \times 10^1$ &
$k_{18}$ & $0.100 \times 10^9$ &
$k_{23}$ & $0.474 \times 10^{-1}$ \\

$k_4$  & $0.860 \times 10^{-3}$ &
$k_9$  & $0.165 \times 10^5$ &
$k_{14}$ & $0.163 \times 10^5$ &
$k_{19}$ & $0.444 \times 10^{12}$ &
$k_{24}$ & $0.178 \times 10^4$ \\

$k_5$  & $0.820 \times 10^{-3}$ &
$k_{10}$ & $0.900 \times 10^4$ &
$k_{15}$ & $0.480 \times 10^7$ &
$k_{20}$ & $0.124 \times 10^4$ &
$k_{25}$ & $0.312 \times 10^1$ \\

\hline
\end{tabular}
\caption{Reaction rate constants for the POLLU system.}
\label{tab:params_pollu}
\end{table}

\begin{figure}[!ht]
    \centering
    \includegraphics[width=\linewidth]{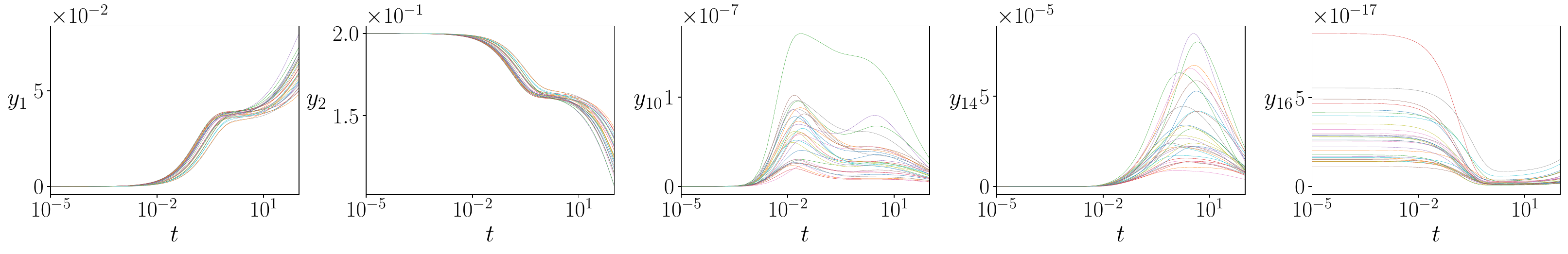}
    \caption{POLLU system solutions under 30 different parameter perturbations within a $\pm 50\%$ range of their nominal values.}
    \label{fig:pollu_perturbed}
\end{figure}

\begin{figure}[!ht]
    \centering
    \begin{subfigure}[c]{0.40\textwidth}
        \centering
        \includegraphics[width=\textwidth]{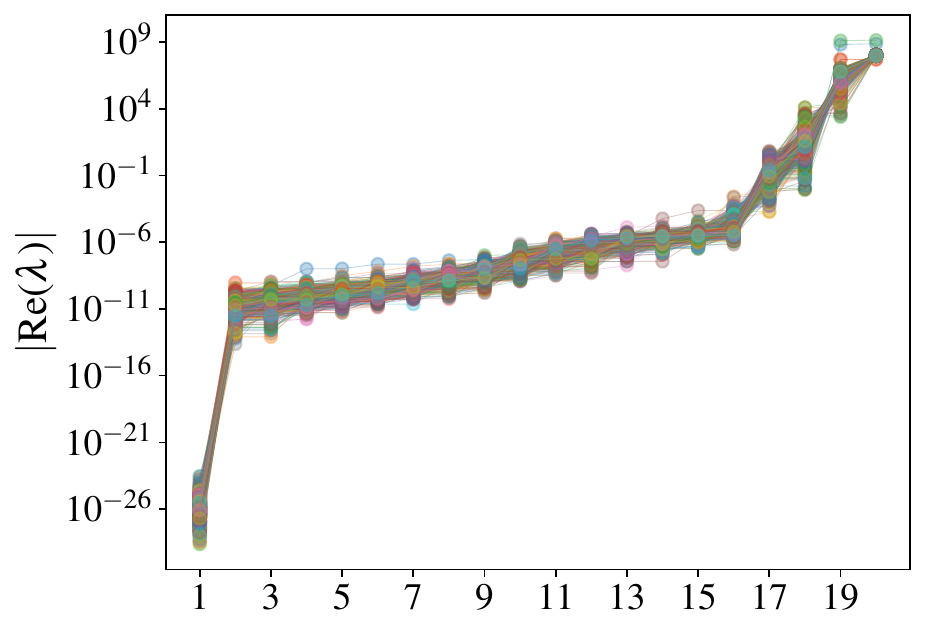}
        \caption{Absolute real component of the Jacobian eigenvalues. Note that each line is computed at a different location along the time axis.}
    \end{subfigure}
    \begin{subfigure}[c]{0.40\textwidth}
        \centering
        \includegraphics[trim=0 -22pt 0 0, clip, width=\textwidth]{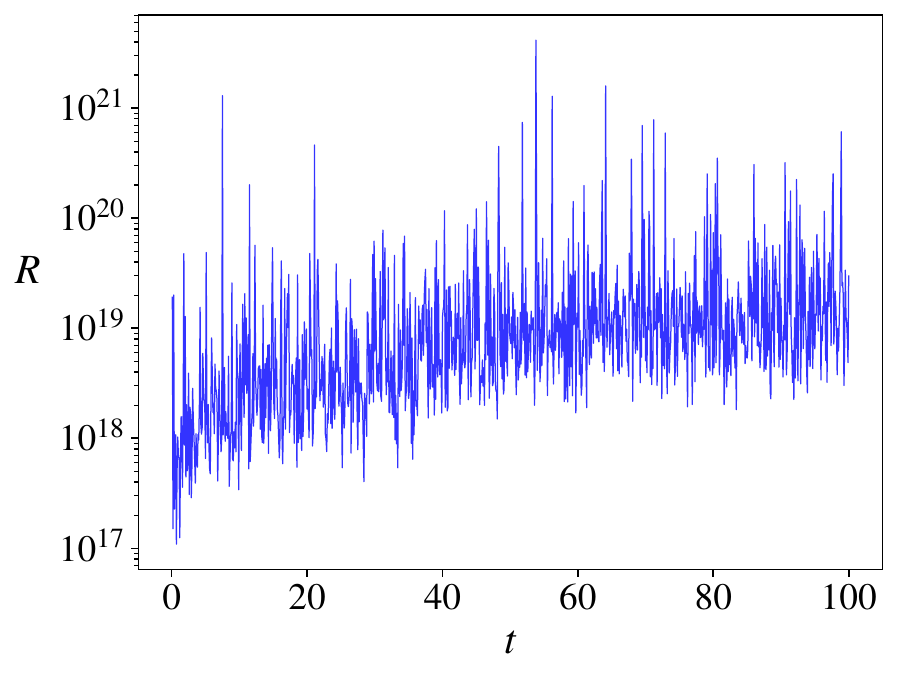}
        \caption{Stiffness ratio vs. time plot.}
    \end{subfigure}
    \caption{Eigenvalues and stiffness ratio for the POLLU system.}
    \label{fig:pollu-stiffness}
\end{figure}

The stiffness analysis for this system reveals the presence of widely separated timescales. Similar to the Robertson problem, we have computed the Jacobian $\bm{J}=\partial \bm{\mathcal{F}} / \partial \bm{y}$. However considering the high dimensionality of the system, we have done so numerically, using central differences, where $\Delta y_j=10^{-7}\times y_j$. Figure~\ref{fig:pollu-stiffness} presents the eigenvalues and stiffness ratio for the POLLU system. We have considered the smallest eigenvalue as a numerical zero since we expect one zero eigenvalue on account of conservation of mass and computed the ratio of the largest to smallest magnitude eigenvalues among the rest.

% ==================================================================================
\subsection{Systems with Reversible and Irreversible Kinetics}\label{subsec:rev-irr}
% ==================================================================================

The following systems originally presented  in~\cite{powers2016combustion}, represent a reversible system and the corresponding irreversible system of reactions, each having three species of reactants of identical molecular mass, and identical initial concentrations.

% ====================================================
\subsubsection{Reversible System}\label{subsubsec:rev}
% ====================================================

This system is modeled as
\begin{equation}
\begin{array}{rcl}
A \xrightleftharpoons[K_{c,1}]{k_1} B 
& & \displaystyle\frac{dy_1}{dt} = \left[-k_1-k_3\left(1+K_{c,1}\,K_{c,2}\right)\right]\,y_1 + \left(\frac{k_1}{K_{c,1}}-k_3\right)y_2 +1,\\[1.1em]
B \xrightleftharpoons[K_{c,2}]{k_2} C 
& \quad\boldsymbol{\to}\quad & \displaystyle\frac{dy_2}{dt} = \left(k_1-\frac{k_2}{K_{c,2}}\right)y_1 + \left[-\frac{k_1}{K_{c,1}}-k_2\left(1+\frac{1}{K_{c,2}}\right)\right]y_2 +1, \\[1.1em]
C \xrightleftharpoons[1/(K_{c,1}K_{c,2})]{k_3} A
& & \displaystyle\text{by conservation of mass: } y_3 = \left(y_1(0)+y_2(0)+y_3(0)\right)-(y_1+y_2),
\end{array}
\end{equation}
with initial conditions
\begin{equation}\label{equ:rev_ic}
y_1(0) = 1/3, y_2(0)=1/3, y_3(0)=1/3,
\end{equation} 
along with reaction rate constants $k_i$ and equilibrium constants $K_{c,i}$ defined as
\begin{equation}\label{equ:rev_parameters}
k_1=1, k_2=2, k_3=1, K_{c,1}=2, K_{c,2}=2.
\end{equation} 

The solution plots \ref{fig:rev-nominal} show that the system attains equilibrium at concentrations: $y_{1,eq} = 1/7, y_{2,eq}=2/7, y_{3,eq}=4/7$. The equilibrium concentrations can be obtained by solving the algebraic equations corresponding to $dy_1/dt=0$ and $dy_2/dt=0$, $y_3=\left(y_1(0)+y_2(0)+y_3(0)\right)-(y_1+y_2)$.
Figure~\ref{fig:rev-perturb} also shows the time evolution of the system corresponding to reaction rate parameters uniformly and independently sampled within a $\pm 30 \%$ range of the nominal values~\eqref{equ:rev_parameters} corresponding to three different choices of initial conditions.

% ======================================================
\subsubsection{Irreversible System}\label{subsubsec:irr}
% ======================================================

This system is defined and modeled as follows
\begin{equation}\label{equ:irr_eq}
\begin{array}{rcl}
A \xrightarrow{k_1} B
& & \displaystyle\frac{dy_1}{dt} = \left(-k_1-k_3\right)y_1 + \left(-k_3\right)y_2 +1, \\[1.1em]
B \xrightarrow{k_2} C
& \quad\rightarrow\quad &\displaystyle \frac{dy_2}{dt} = k_1y_1 + \left(-k_2\right)y_2, \\[1.1em]
C \xrightarrow{k_3} A
& & \text{by conservation of mass: } y_3 = \left(y_1(0)+y_2(0)+y_3(0)\right)-(y_1+y_2)
\end{array}
\end{equation}
with initial conditions
\begin{equation}\label{equ:irr_ic}
y_1(0) = 1/3, y_2(0)=1/3, y_3(0)=1/3,
\end{equation}
and reaction rate constants
\begin{equation}\label{equ:irr_parameters}
k_1=1, k_2=2, k_3=1.
\end{equation}

\begin{figure}[!ht]
\centering
    \begin{subfigure}[t]{0.365\textwidth}
        \centering
        \includegraphics[width=\textwidth]{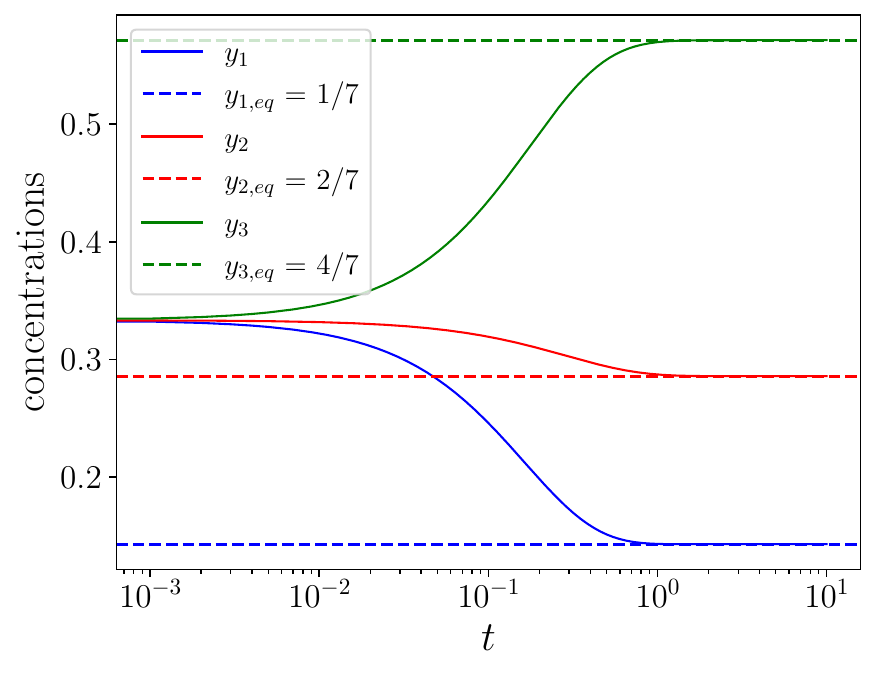}
        \caption{Trajectories for the reversible system, for rate parameters as in~\eqref{equ:rev_parameters}, and initial conditions as in~\eqref{equ:rev_ic}.}
        \label{fig:rev-nominal}
    \end{subfigure}
    $\quad\quad$
    \begin{subfigure}[t]{0.37\textwidth}
        \centering
        \includegraphics[width=\textwidth]{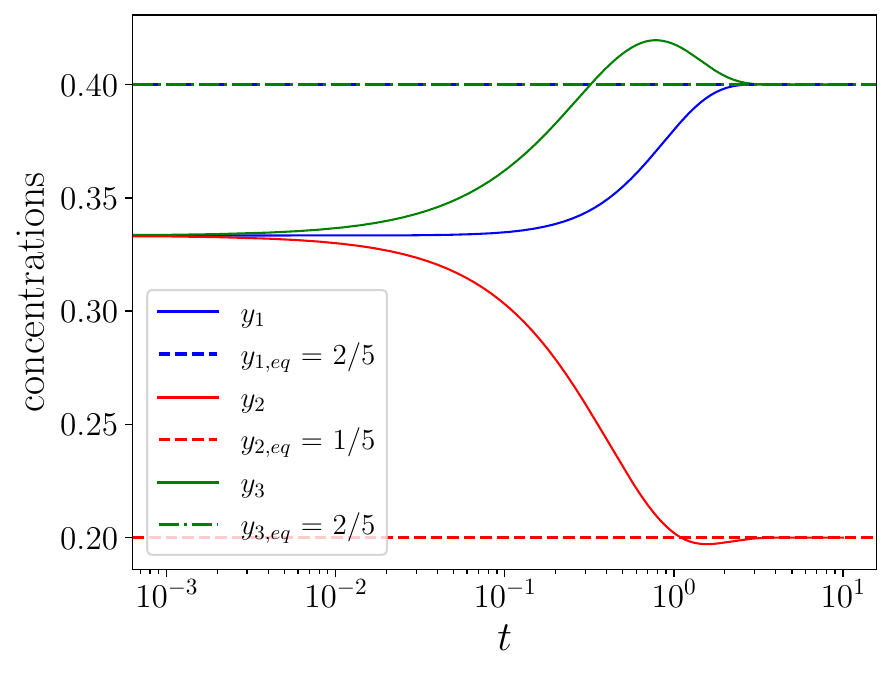}
        \caption{Trajectories for the irreversible system, for rate parameters as in~\eqref{equ:irr_parameters}, and initial conditions as in~\eqref{equ:irr_ic}.}
        \label{fig:irr-nominal}
    \end{subfigure}
    \caption{Baseline solutions for the reversible and irreversible systems.}
\end{figure}

\begin{figure}[!ht]
\begin{subfigure}[t]{\textwidth}
\centering
    \begin{subfigure}[t]{0.32\textwidth}
        \centering
        \includegraphics[width=\textwidth]{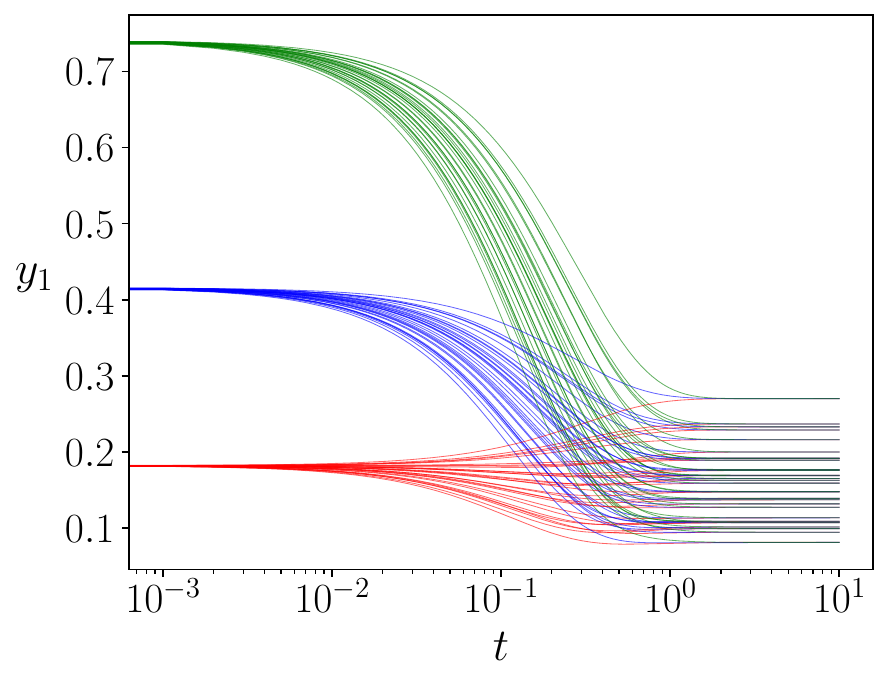}
    \end{subfigure}
    \begin{subfigure}[t]{0.32\textwidth}
        \centering
        \includegraphics[width=\textwidth]{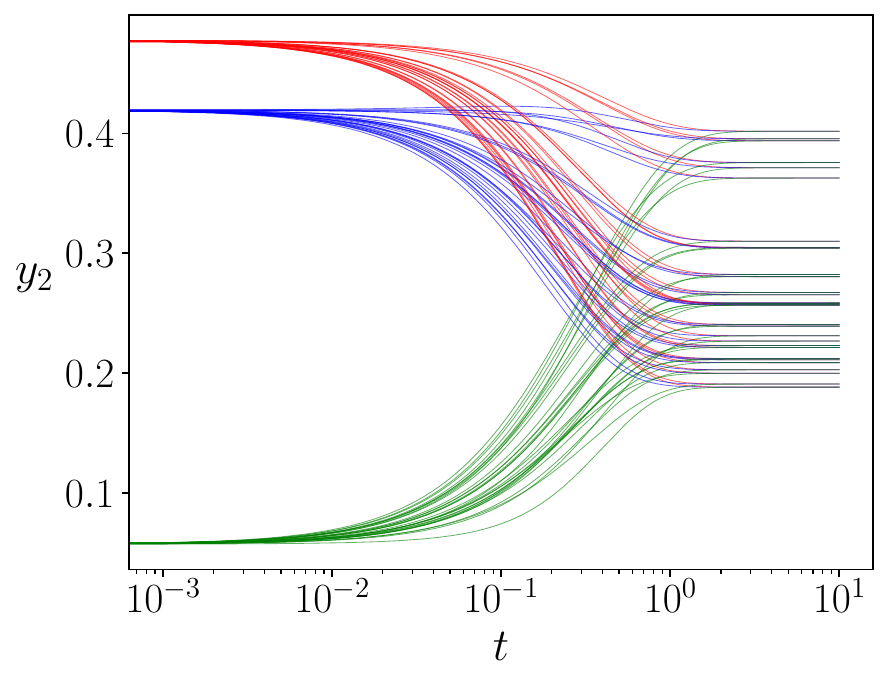}
    \end{subfigure}
        \begin{subfigure}[t]{0.32\textwidth}
        \centering
        \includegraphics[width=\textwidth]{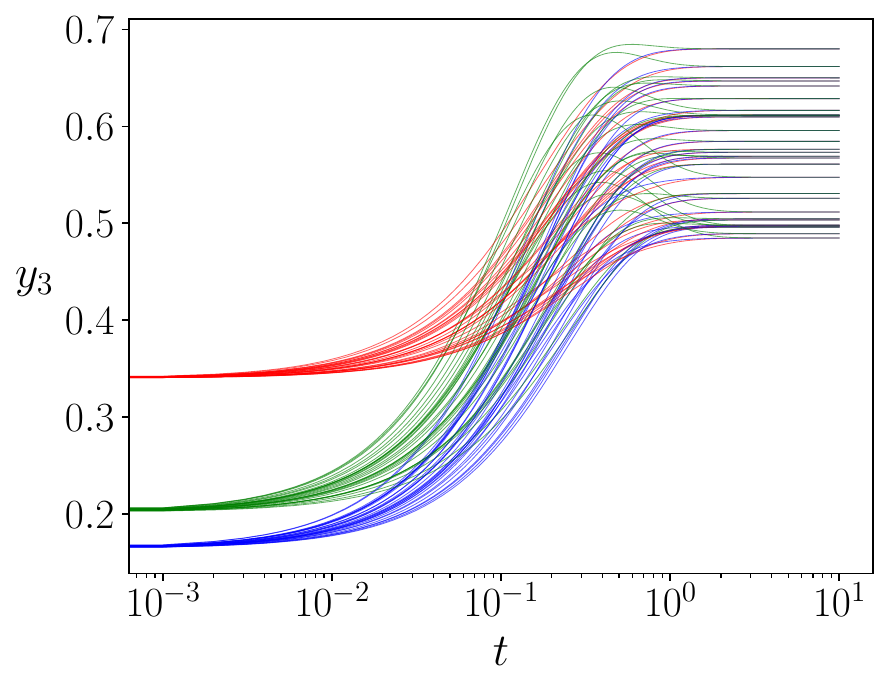}
    \end{subfigure}
    \caption{Reversible system.}
    \label{fig:rev-perturb}
\end{subfigure}

\begin{subfigure}[t]{\textwidth}
\centering
    \begin{subfigure}[t]{0.30\textwidth}
        \centering
        \includegraphics[width=\textwidth]{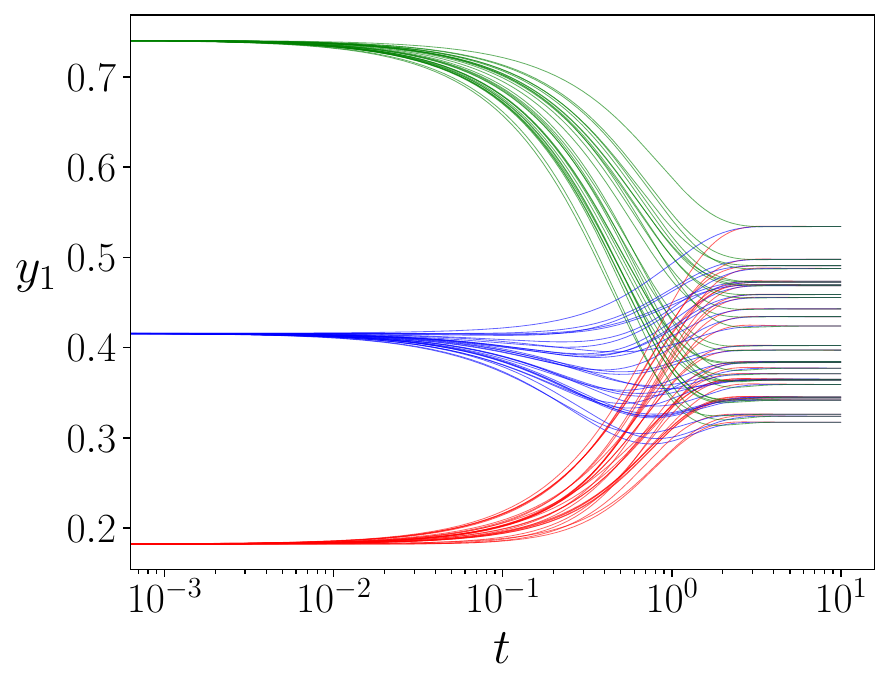}
    \end{subfigure}
    \begin{subfigure}[t]{0.30\textwidth}
        \centering
        \includegraphics[width=\textwidth]{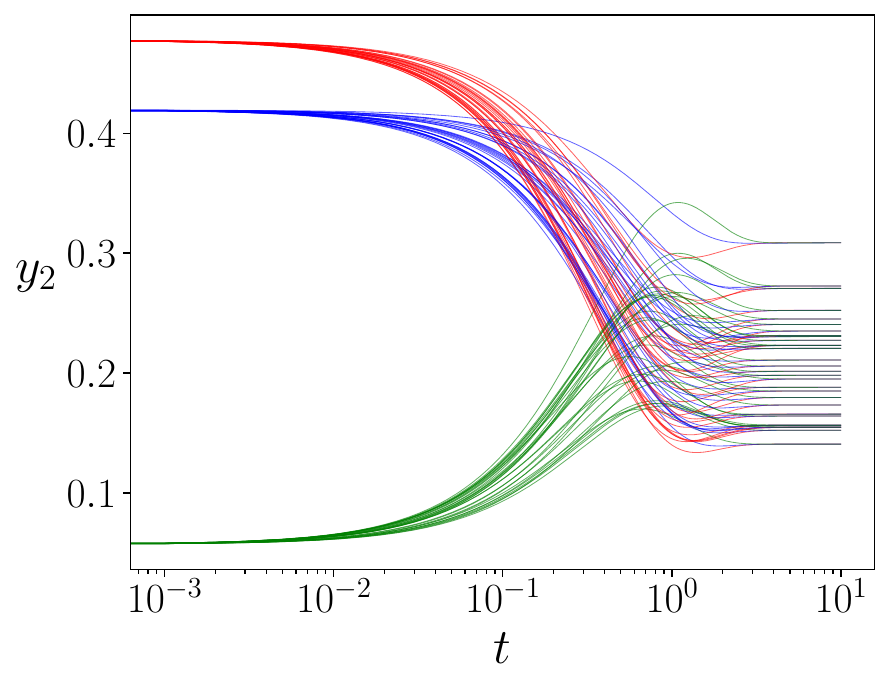}
    \end{subfigure}
        \begin{subfigure}[t]{0.32\textwidth}
        \centering
        \includegraphics[width=\textwidth]{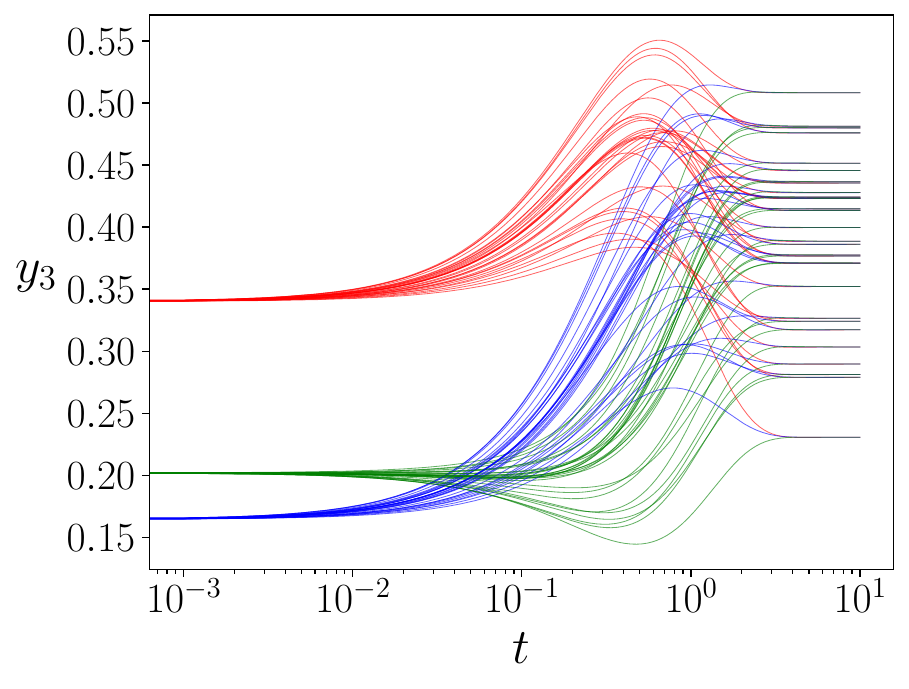}
    \end{subfigure} 
    \caption{Irreversible system.}
    \label{fig:irr-perturb}
\end{subfigure}
\caption{Solutions for 30 perturbed trajectories corresponding to each of three different initial conditions for the (a) reversible and (b) irreversible systems. Initial conditions are randomly chosen from $(0,1)$ while preserving the sum of the three components to be 1, and the parameters are uniformly sampled over intervals obtained by perturbing their nominal values by $\pm$30\%.}
\end{figure}

The solution plots \ref{fig:irr-nominal} show that the system attains equilibrium at concentrations: $y_{1,eq} = 2/5, y_{2,eq}=1/5, y_{3,eq}=2/5$. The time evolution of the system corresponding to the reaction rate parameters uniformly and independently perturbed in a $\pm 30 \%$ range of the nominal values~\eqref{equ:irr_parameters} corresponding to three different choices of initial conditions is shown in Figure~\ref{fig:irr-perturb}. These examples exhibit different behavior on their approaches to equilibrium. The reversible system approaches with real negative eigenvalues.  The irreversible system has complex eigenvalues with a negative real part, and the complex part induces the oscillatory behavior.

% =============================================================
\subsection{Hydrogen-air kinetics problem}\label{sec:hydro-air}
% =============================================================

\begin{figure}[!ht]
    \centering
    \includegraphics[width=0.6\linewidth]{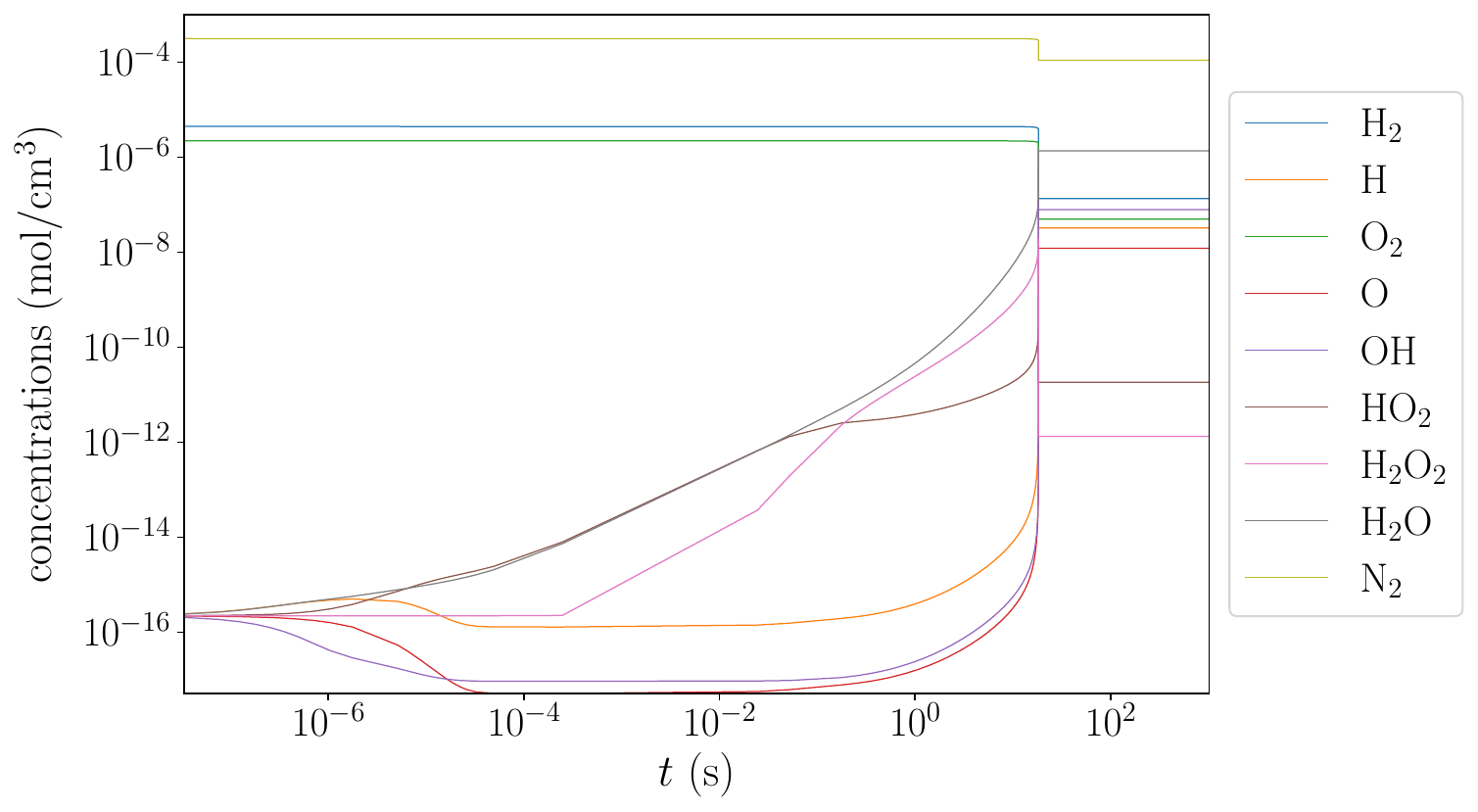}
    \caption{Solution plots for the hydrogen-air system, with initial conditions as in \eqref{equ:water_ic}, and reaction rates as in Table~\ref{tab:water-def}.}
    \label{fig:water_default}
\end{figure}

The hydrogen-air kinetics problem, a detailed kinetics mechanism under adiabatic and isobaric conditions, is given in~\cite{jmp-water} and is described in Table~\ref{tab:water-def}. Temperature is solved such that the internal energy of the system is conserved (according to the first law of thermodynamics), with specific heats obtained from standard polynomial fits to experimental data. It consists of 19 reversible chemical reactions that involve 9 chemical species, namely, H$_2$, H,  O$_2$, O, OH, HO$_2$, H$_2$O$_2$, H$_2$O, and N$_2$, where N$_2$ acts as an inert diluent for the mixture. Figure~\ref{fig:water_default} shows the time evolution of molar concentrations of the different species corresponding to the nominal parameters shown in Table~\ref{tab:water-def}. Note that the molar concentration of N$_2$ changes despite it being an inert participant of the reaction. This is due to the decrease in mass density of the isobaric system with increase in specific volume as temperature rises. By conservation of mass, the denominator of the molar concentration units (mol/cm$^3$) increases. Hence the change in the molar concentration of N$_2$ is essentially a proxy for the change of the mass density (inverse specific volume). The time domain is chosen to be $[0,100]$, and initial conditions are $T(0)=800$~K, $P(0)=1$~atm together with
\begin{equation}\label{equ:water_ic}
    \bm{y}(0)= [2.00\times10^{-5},\; 10^{-15},\; 10^{-5},\; 10^{-15},\;  10^{-15},\; 10^{-15},\; 10^{-15},\; 10^{-15},\; 3.76 \times 10^{-5}]^T.
\end{equation}
The chemical concentrations are measured in mol/cm$^3$ and time is measured in s. The different components of InVAErt are trained on log values of the data, hence there is a need to clip pure zeros. Keeping that in mind, the zero initial conditions are initialized as $10^{-15}$ instead. Note that the $(A, b, E)$ values in Table~\ref{tab:rober-emu-single-step} parametrize the respective rate constants $k$ via the Arrhenius parametrization $k=A\,T^b\,e^{-E/RT}$.

\begin{table}[H]
\centering
\renewcommand{\arraystretch}{1.2}
\begin{tabular}{| c l | c c c |}
\hline
$\boldsymbol{j}$ & {\bf Reaction} &
$\boldsymbol{A_j}$ (mol/cm$^3)^{1 - \sum_{i=1}^{N} \nu '_{ij}}/s/{K}^{b_j}$ &
$\boldsymbol{b_j}$ &
$\boldsymbol{E_j}$ (cal/mol) \\
\hline
1  & H$_2$ + O$_2$ $\rightleftharpoons$ OH + OH            & 1.70$\times$10$^{13}$ & 0.0  & 47780 \\
2  & OH + H$_2$ $\rightleftharpoons$ H$_2$O + H           & 1.17$\times$10$^{9}$  & 1.3  & 3626 \\
3  & H + O$_2$ $\rightleftharpoons$ OH + O                & 5.13$\times$10$^{16}$ & $-0.816$ & 16507 \\
4  & O + H$_2$ $\rightleftharpoons$ OH + H                & 1.80$\times$10$^{10}$ & 1.0  & 8826 \\
5  & H + O$_2$ + M $\rightleftharpoons$ HO$_2$ + M$^a$    & 2.10$\times$10$^{18}$ & $-1.0$ & 0 \\
6  & H + O$_2$ + O$_2$ $\rightleftharpoons$ HO$_2$ + O$_2$ & 6.70$\times$10$^{19}$ & $-1.42$ & 0 \\
7  & H + O$_2$ + N$_2$ $\rightleftharpoons$ HO$_2$ + N$_2$ & 6.70$\times$10$^{19}$ & $-1.42$ & 0 \\
8  & OH + HO$_2$ $\rightleftharpoons$ H$_2$O + O$_2$      & 5.00$\times$10$^{13}$ & 0.0  & 1000 \\
9  & H + HO$_2$ $\rightleftharpoons$ OH + OH              & 2.50$\times$10$^{14}$ & 0.0  & 1900 \\
10 & O + HO$_2$ $\rightleftharpoons$ O$_2$ + OH           & 4.80$\times$10$^{13}$ & 0.0  & 1000 \\
11 & OH + OH $\rightleftharpoons$ O + H$_2$O              & 6.00$\times$10$^{8}$  & 1.3  & 0 \\
12 & H$_2$ + M $\rightleftharpoons$ H + H + M$^b$         & 2.23$\times$10$^{12}$ & 0.5  & 92600 \\
13 & O$_2$ + M $\rightleftharpoons$ O + O + M             & 1.85$\times$10$^{11}$ & 0.5  & 95560 \\
14 & H + OH + M $\rightleftharpoons$ H$_2$O + M$^c$       & 7.50$\times$10$^{23}$ & $-2.6$ & 0 \\
15 & H + HO$_2$ $\rightleftharpoons$ H$_2$ + O$_2$        & 2.50$\times$10$^{13}$ & 0.0  & 700 \\
16 & HO$_2$ + HO$_2$ $\rightleftharpoons$ H$_2$O$_2$ + O$_2$ & 2.00$\times$10$^{12}$ & 0.0 & 0 \\
17 & H$_2$O + M $\rightleftharpoons$ OH + OH + M          & 1.30$\times$10$^{17}$ & 0.0  & 45500 \\
18 & H$_2$O$_2$ + H $\rightleftharpoons$ HO$_2$ + H$_2$   & 1.60$\times$10$^{12}$ & 0.0  & 3800 \\
19 & H$_2$O$_2$ + OH $\rightleftharpoons$ H$_2$O + HO$_2$ & 1.00$\times$10$^{13}$ & 0.0  & 1800 \\
\hline
\end{tabular}

\vspace{2pt}
\begin{minipage}{0.85\textwidth}
\raggedright
\textit{Note.} The third-body efficiency $\alpha_{j,i}$ is the relative effectiveness of species $i$  in transferring energy during the three-molecule reaction $j$, normalized against $\alpha = 1$. Unless otherwise specified, the third body collision efficiency coefficients are unity, $\alpha=1$.\\
The non-unity third-body collision efficiency coefficients are:\\
$^a$ for reaction 5, $\alpha_{5,\mathrm{H_2}} = 3.3$, $\alpha_{5,\mathrm{H_2O}} = 21$; \\
$^b$ for reaction 12, $\alpha_{12,\mathrm{H_2}} = 3$, $\alpha_{12,\mathrm{H_2O}} = 6$, $\alpha_{12,\mathrm{H}} = 2$; \\
$^c$ for reaction 14, $\alpha_{14,\mathrm{H_2O}} = 20$.\\
\end{minipage}
\caption{Chemical reactions and Arrhenius parameters for the H$_2$-air kinetics problem.}
\label{tab:water-def}

\end{table}

\begin{figure}[!ht]
    \centering
    \includegraphics[width=0.38\linewidth]{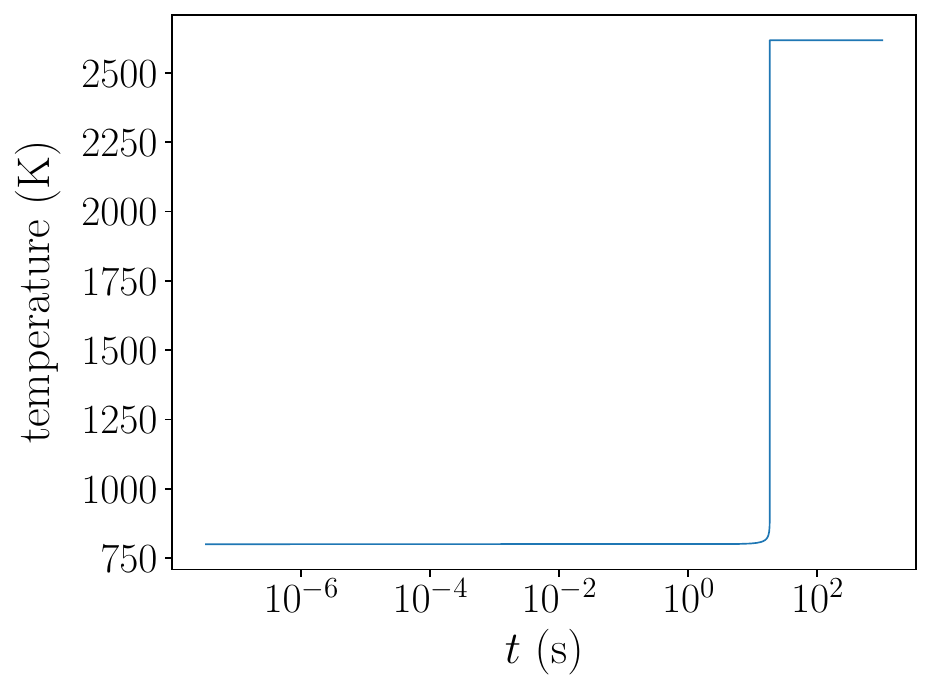}
    \caption{Evolution of the temperature of the hydrogen-air system, with initial conditions as in \eqref{equ:water_ic}, and reaction rates as in Table~\ref{tab:water-def}.}
    \label{fig:water-temp}
\end{figure}

\begin{figure}[!ht]
    \centering
    \includegraphics[width=\linewidth]{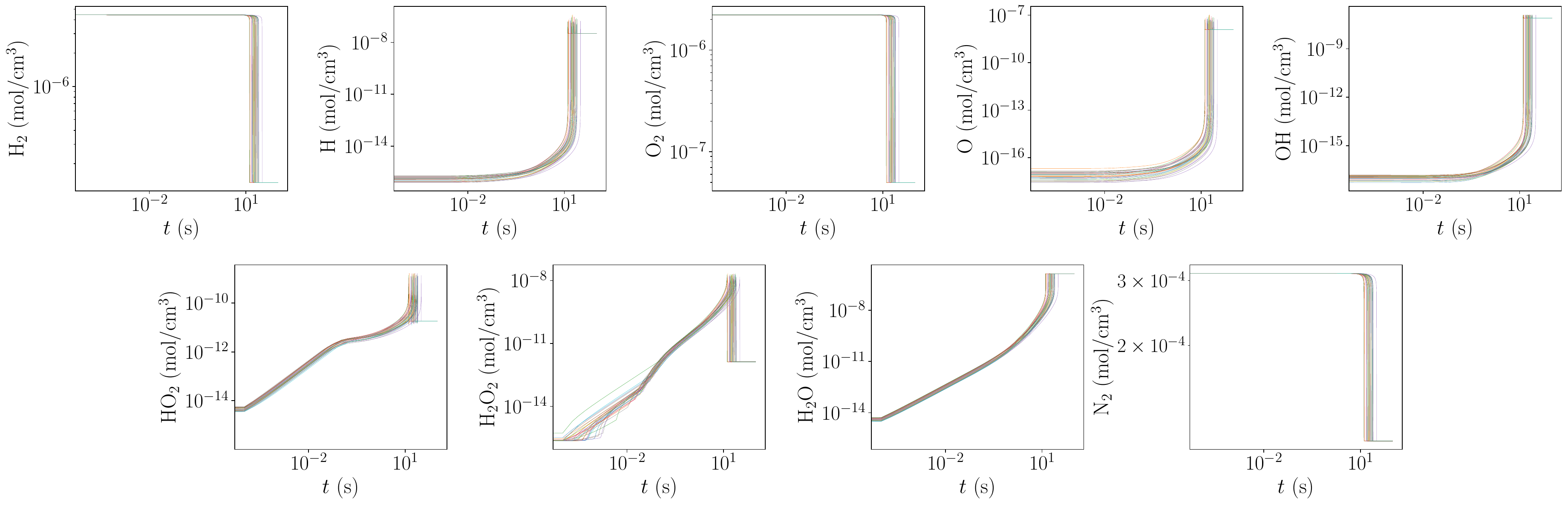}
    \caption{Solution plots for a parametric family of hydrogen-air systems. Rate parameters are perturbed 30 times such that $(A, T_0^b, e^{-E/RT_0})$ vary in a $\pm$ 50\% range of their nominal values.}
    \label{fig:water_perturbed}
\end{figure}

% ==============================================
\section{Dataset Generation}\label{sec:data_gen}
% ==============================================

 For the purpose of data generation, we have simulated the solution of each system using the \emph{Radau} (Implicit Runge-Kutta method of the Radau IIA family of order 5~\cite[Section IV.8.]{wanner1996solving}) method of the
\emph{scipy.integrate.solve\_ivp} function, with the exception of the hydrogen-air system, for which we have used the DLSODE solution method in \emph{Chemkin}~\cite{kee1989chemkin}. 
Simulations were performed corresponding to different perturbed parameter values, and the target number of training data points was sampled from each simulation based on the chosen method of time-sampling, as described in the next section. 

% ================================================================
\subsection{Time-Sampling Strategies}\label{subsec: time-sampling}
% ================================================================

 The stiffness inherent in our systems results in solution trajectories that evolve over widely diverse time scales. 
Our investigation indicates that the distribution of time samples across each simulation is critical to the quality of the resulting training data.
In particular, the time-sampling strategy must be consistent with the temporal evolution of the solution trajectories. 
In this section, we first introduce different sampling strategies with an example. 
Table~\ref{tab:ratios-time-sampling} then reports the choice of the sampling approach we found to be optimal for each system we analyzed.

\begin{figure}[!ht]
\centering
\begin{subfigure}{0.32\textwidth}
    \centering
    \includegraphics[width=\textwidth]{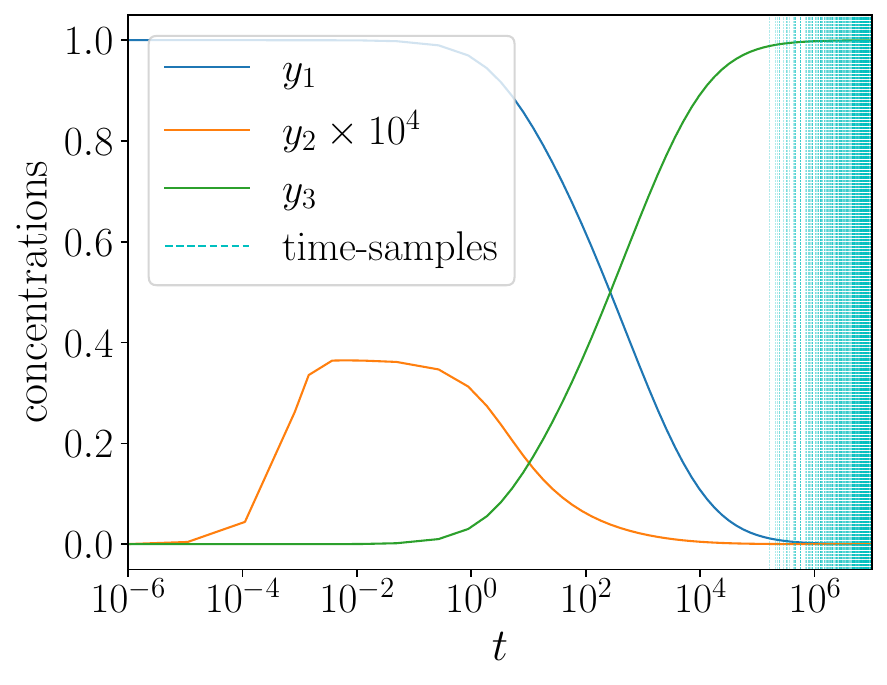}
    \caption{Uniform time sampling.}
     \label{fig:rober-uni}
\end{subfigure}
\begin{subfigure}{0.32\textwidth}
    \centering
    \includegraphics[width=\textwidth]{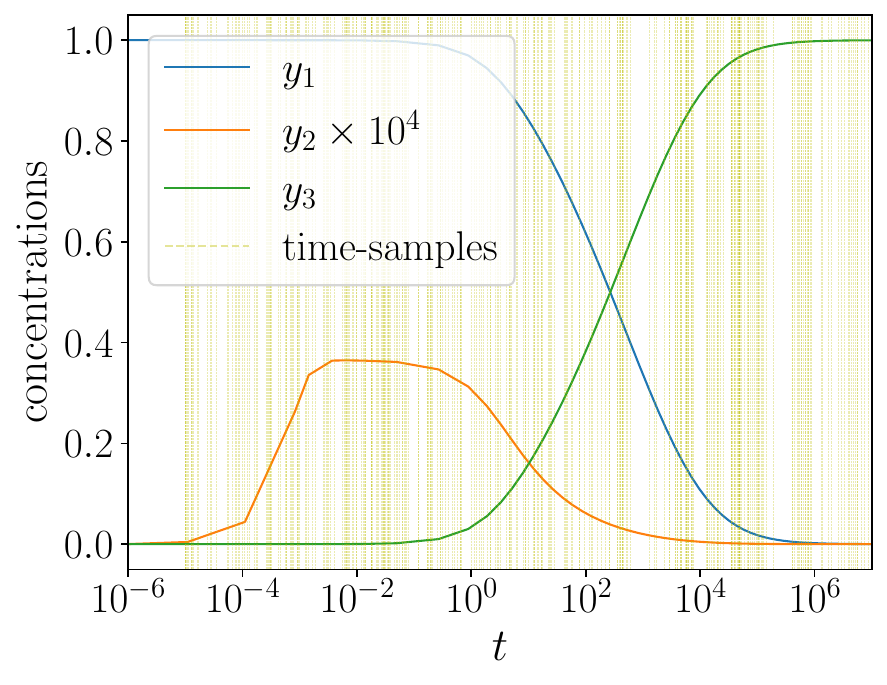}
    \caption{Logarithmic time sampling.}
     \label{fig:rober-log}
\end{subfigure}
\begin{subfigure}{0.32\textwidth}
    \centering
    \includegraphics[width=\textwidth]{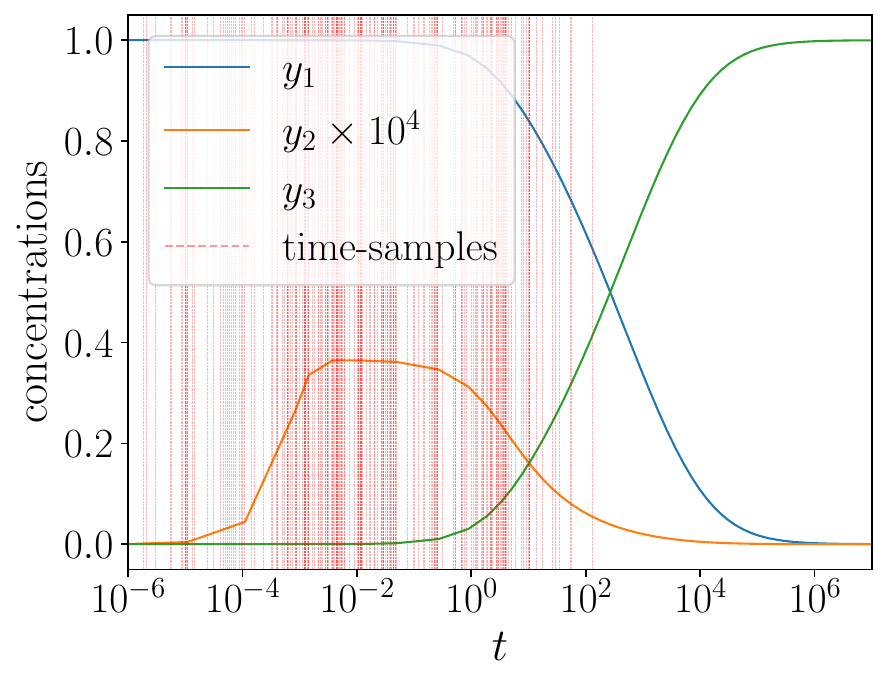}
    \caption{Inverse time sampling.}
    \label{fig:rober-inv}
\end{subfigure}
\caption{Time sampling strategies illustrated using the Robertson problem.}
\end{figure}

\begin{enumerate}
    \item \textbf{Uniform time-sampling:} We simply select solutions at times that are  uniformly distributed over the entire simulation duration. We call this~\emph{uniform} time-sampling. An application of this strategy to the Robertson problem in Figure~\ref{fig:rober-uni} results in a redundant number of training points towards the end of the logarithmic simulation time, despite most of the interesting dynamics occurring at earlier stages. 
    Hence this approach generates a sub-optimal training dataset for this system.

    \item \textbf{Logarithmic time-sampling:} Time instances are collected uniformly at random from $\log_{10} t$ for the entire simulation duration, then converted to actual time samples using $10^{t}$. Figure~\ref{fig:rober-log} indicates that log-samples are able to better capture the dynamics of the Robertson system than uniform sampling.
    
    \item \textbf{Inverse time sampling:} The core idea is to distribute time samples preferentially to intervals where the solution state exhibits rapid temporal change. 
    This is implemented by first computing the time derivatives either for a selected state variable, or in general, by taking the maximum time derivative across all species concentrations at each point in time. 
    We then construct a monotonically increasing cumulative metric by integrating the magnitude of these gradients over time. 
    Uniform time samples are finally drawn through inverse cumulative projections (an approach commonly referred to as \emph{inverse transform sampling}~\cite{devroye2006nonuniform}) while the corresponding concentration are determined via interpolation.
    Figure~\ref{fig:rober-inv} shows the sample distribution determined through inverse time sampling for the Robertson system. Since all important features in the dynamical response occur early in time, we can expect inverse time and log sampling to generate equally appropriate training datasets for this problem.
\end{enumerate}

\begin{figure}[!ht]
    \centering
    \includegraphics[width=0.5\linewidth]{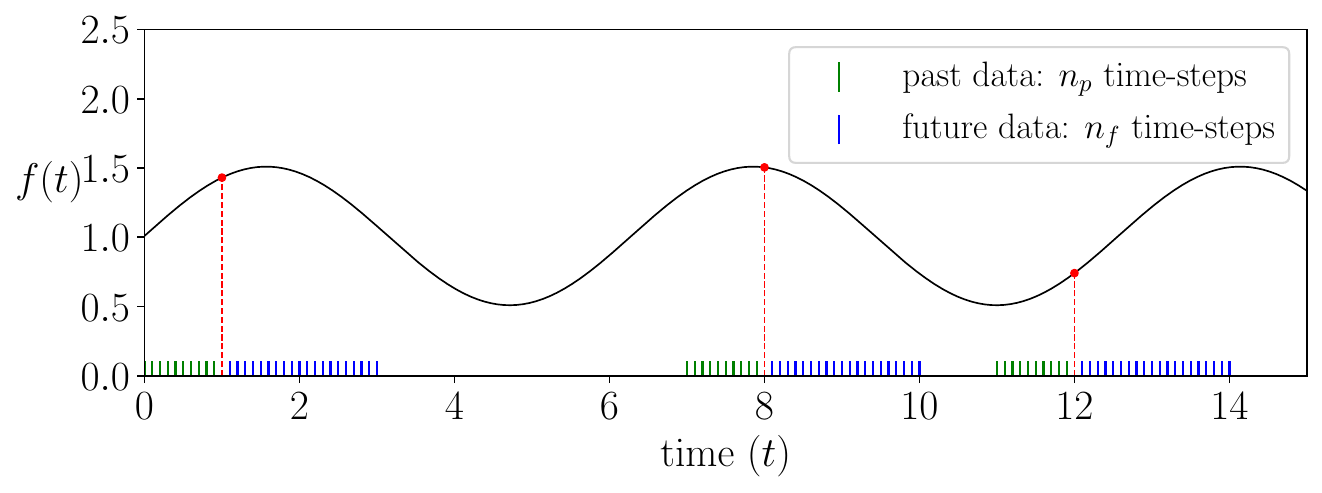}
    \caption{Training dataset generation for the emulator. The time instances in red are selected using one of the time-sampling methods (uniform, log or inverse) for an arbitrary function $f(t)$. Using $\Delta t=10^{-1}, n_p=10, n_f=20$, we train the emulator to predict the $n_f$ (future) solutions at the times marked in blue, from the $n_p$ function values (past and present) marked in green.}
    \label{fig:delta-t-explanation}
\end{figure}

In practice, any combination of these techniques can be employed to generate training data, depending on the problem at hand.
For each simulation, the time-sampling strategies give us the required number of locations along the time domain at which we sample training data.
Preceding each such location $t$, we extract $n_p$ time-steps of data separated by a constant time-step $\Delta t$ chosen based on the problem, to be used as auxiliary data. 
When training an emulator, we predict the next $n_f$ time-steps, each separated by the same interval $\Delta t$, as illustrated in Figure~\ref{fig:delta-t-explanation}.
Finally, for all models, the logarithm of the species concentration is computed before feeding the data to the network for training.
As shown in Table~\ref{tab:ratios-time-sampling}, we have chosen logarithmic sampling for most of our systems since it has proved to be sufficient to reach satisfactory accuracy for each of these. However, for the hydrogen-air reaction, inverse time sampling is adopted to more effectively resolve the sharp, time-localized ignition peaks.

\begin{table}[!ht]
    \centering
    \begin{tabular}{|c|c|c|c|}
    \hline
         \textbf{Problem} & \textbf{Uniform (\%)} & \textbf{Logarithmic (\%)} & \textbf{Inverse (\%)} \\
         \hline
         Robertson & 0 & 100 & 0 \\
         POLLU & 0 & 100 & 0 \\
         Reversible & 0 & 100 & 0 \\
         Irreversible & 0 & 100 & 0 \\
         Hydrogen-air & 5 & 50 & 45 \\
         \hline
    \end{tabular}
    \caption{Source of training data by sampling technique for each problem considered in Section~\ref{sec:chem_odes}.}
    \label{tab:ratios-time-sampling}
\end{table}

% ========================================================================
\section{Emulation and inference with inVAErt networks}\label{sec:invaert}
% ========================================================================

% =====================
\subsection{Background}
% =====================

In this study, we consider a \emph{model} $\bm{f}$ as a map describing the change in a number of \emph{outputs} $\bm{y}\in\mathcal{Y}$ as a function of a set of \emph{inputs} $\bm{v}\in\mathcal{V}$ and, possibly, space $\bm{x}$ and time $t\in\bm{\mathcal{T}}=[0,T]$, according to a number of physically justified driving principles. 
Formally, evaluating $\bm{f}:\bm{\mathcal{V}} \mapsto \bm{\mathcal{Y}}$ (an \emph{input-to-output map}) given an input, space and time (here we assume $\bm{\mathcal{V}}$ includes all three) is referred to as solving a \emph{forward problem}, while determining its inverse (an \emph{output-to-input} map) $\bm{f}^{-1}:\bm{\mathcal{Y}} \mapsto \bm{\mathcal{V}}$ corresponds to the solution of an \emph{inverse problem}.

While the forward problem generates a single output under a deterministic model $\bm{f}$, and multiple outputs in the presence of irreducible aleatoric uncertainty (see, e.g.,~\cite{de2025quantification}), the inverse map, in many cases, is \emph{ill-posed}, as multiple sets of inputs can give rise to the same outputs $\bm{y}$. 
In such a case, the model inputs $\bm{v}$ are said to be \emph{non-identifiable} from the outputs $\bm{y}$, and a \emph{manifold} of non-identifiable inputs $\mathcal{M}_{\bm{y}}$ is defined as $\{\bm{v}: \bm{f}(\bm{v}) = \bm{y}\}$.

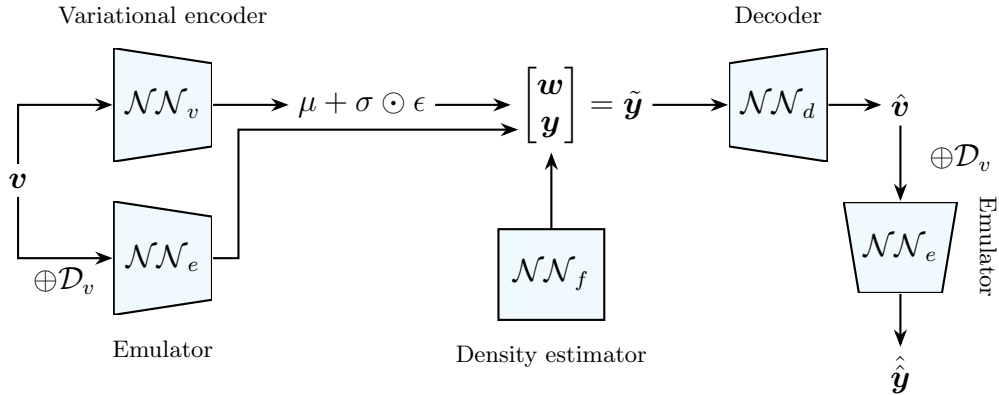
\begin{figure}[!ht]
\centering
\begin{tikzpicture}[
    >=Stealth,
    node distance=1.8cm,
    every node/.style={font=\large},
    nn/.style={
  thick,
  align=center,
  minimum height=1.5cm,
  minimum width=1.3cm,
  path picture={
    \path[fill=cyan!5] 
      (path picture bounding box.north west) --
      (path picture bounding box.south west) --
      ([yshift=0.20cm]path picture bounding box.south east) --
      ([yshift=-0.20cm]path picture bounding box.north east) --
      cycle;
    \draw[thick]
      (path picture bounding box.north west) --
      (path picture bounding box.south west) --
      ([yshift=0.20cm]path picture bounding box.south east) --
      ([yshift=-0.20cm]path picture bounding box.north east) --
      cycle;
  }},
    nnflip/.style={
  thick,
  align=center,
  minimum height=1.5cm,
  minimum width=1.3cm,
  path picture={
    \path[fill=cyan!5] 
      ([yshift=-0.20cm]path picture bounding box.north west) --
      ([yshift=0.20cm]path picture bounding box.south west) --
      (path picture bounding box.south east) --
      (path picture bounding box.north east) --
      cycle;
    \draw[thick]
      ([yshift=-0.20cm]path picture bounding box.north west) --
      ([yshift=0.20cm]path picture bounding box.south west) --
      (path picture bounding box.south east) --
      (path picture bounding box.north east) --
      cycle;
  }},
    nndown/.style={
  thick,
  align=center,
  minimum height=1.2cm,
  minimum width=1.5cm,
  path picture={
    \path[fill=cyan!5] 
      (path picture bounding box.north west) --
      ([xshift=0.20cm]path picture bounding box.south west) --
      ([xshift=-0.20cm]path picture bounding box.south east) --
      (path picture bounding box.north east) --
      cycle;
    \draw[thick]
      (path picture bounding box.north west) --
      ([xshift=0.20cm]path picture bounding box.south west) --
      ([xshift=-0.20cm]path picture bounding box.south east) --
      (path picture bounding box.north east) --
      cycle;
  }},
    block/.style={draw, rectangle, minimum height=1cm, minimum width=1.4cm, thick},
    arrow/.style={->, thick}
]

% --- Left side (encoder + emulator) ---
\node (v) {$\boldsymbol{v}$};

\node[nn, right=1cm of v, yshift=1cm] (nnv) {$\mathcal{NN}_v$};
\node[nn, below=0.5cm of nnv] (nne1) {$\mathcal{NN}_e$};

\node[left=0.1cm of nne1, yshift=-3mm] (Dv) {$\oplus \mathcal{D}_v$};

\draw[arrow] (v) |- (nnv);
\draw[arrow] (v) |- (nne1);

% --- Latent sampling ---
\node[right=1cm of nnv] (sample) {$\mu + \sigma \odot \epsilon$};

\draw[arrow] (nnv) -- (sample);

% --- Latent concatenation ---
\node[right=1cm of sample] (latent) {$
\begin{bmatrix}
\boldsymbol{w}\\
\boldsymbol{y}
\end{bmatrix}=\tilde{\boldsymbol{y}}
$};

\draw[arrow] (sample) -- (latent);
\draw[arrow] (nne1) 
  -- ++(1cm,0)
  -- ++ (0, 1.7cm)
  -- ++ (3.7cm, 0);

% --- Density estimator ---
\node[rectangle, fill=cyan!5, thick,
  align=center,
  minimum height=1.2cm,
  minimum width=1.4cm, draw, below=1cm of latent,
  xshift=-0.4cm] (nnf) {$\mathcal{NN}_f$};
\node[below=0.2cm of nnf] {\small Density estimator};

\draw[arrow] (nnf.north) -- ++(0, 1cm); %(latent);

% --- Decoder ---
\node[nnflip, right=1cm of latent] (nnd) {$\mathcal{NN}_d$};

\draw[arrow] (latent) -- (nnd);

% --- Emulator (right) ---
\node[right=0.7cm of nnd] (vhat) {$\hat{\boldsymbol{v}}$};
\node[nndown, below=1cm of vhat] (nne2) {$\mathcal{NN}_e$};
\node[below=0.7cm of nne2] (yhat) {$\hat{\hat{\boldsymbol{y}}}$};

\draw[arrow] (nnd) -- (vhat);
\draw[arrow] (vhat) -- (nne2);
\draw[arrow] (nne2) -- (yhat);

\node[right=0.1mm of vhat, yshift=-0.7cm] {$\oplus \mathcal{D}_v$};

% --- Section labels ---
\node[above=0.2cm of nnv] {\small Variational encoder};
\node[above=0.2cm of nnd] {\small Decoder};
\node[below=0.2cm of nne1] {\small Emulator};
\node[right=0.4cm of nne2, yshift=0.8cm, rotate=-90] {\small Emulator};

\end{tikzpicture}
\caption{Schematic of all components of an inVAErt network and their interactions.}
\label{fig:invert-maps}
\end{figure}

An inVAErt network~\cite{tong2024invaert} provides a data-driven approach for fast evaluation of both the forward and inverse maps, with the ability to characterize manifolds of non-indentifiabile inputs in ill-posed inverse problems.
As shown in Figure~\ref{fig:invert-maps}, an inVAErt network consists of an emulator $\mathcal{NN}_e: \bm{\mathcal{V}}\mapsto \bm{\mathcal{Y}}$ as a surrogate for the input-to-output map $\bm{f}$, a $L$-layer flow-based density estimator $\mathcal{NN}_f: \bm{\mathcal{Z}}\mapsto \bm{\mathcal{Y}}$ for the output distribution $p(\bm{y})$, and a variational encoder $\mathcal{NN}_v: \bm{\mathcal{V}}\mapsto \bm{\mathcal{W}}$ generating an input-dependent latent space, capturing the lack of bijectivity between inputs and outputs. 
Additionally, a decoder $\mathcal{NN}_d: \bm{\mathcal{Y}} \times \bm{\mathcal{W}}\mapsto \bm{\mathcal{V}}$ approximates a bijective inverse map from concatenated outputs and latent variables to model inputs.
Given a dataset $\mathcal{D} = \big\{(\bv^{(j)}, \by^{(j)})\big\}_{j=1}^{N}$, the optimal weights and biases for each inVAErt network component are obtained by minimizing loss functions of the form
\begin{equation}
\begin{split}
     \bphi_e^{\text{opt}} &= \displaystyle \argmin\limits_{\bphi_e} \sum\limits_{j=1}^N \Big\|\by_j - \mathcal{NN}_e(\bv_j; \bphi_e) \Big\|_2^2 \ ,\quad
    \bphi_f^{\text{opt}} = \displaystyle \argmax\limits_{\bphi_f} \sum\limits_{j=1}^N \left( \log \pi_0(\bz^{(0)}_j) - \sum\limits_{l=1}^L \log\left|\det{\left(\frac{d\bz_j^{(l)}}{d\bz_j^{(l-1)}}\right)}\right| \right) \ , \\
    \vspace{-0.2cm}\\
    \bphi_v^{\text{opt}}, \bphi_d^{\text{opt}} &= \displaystyle  \argmin\limits_{\bphi_v, \bphi_d} \Bigg[ \beta_d \sum\limits_{j=1}^N \Big\|\bv_j - \mathcal{NN}_d(\by_j, \bw_j; \bphi_d)\Big\|_2^2 + \frac{\beta_v}{2} \sum\limits_{j=1}^N\sum\limits_{k=1}^{\dim (\bw)} \Big( \mu_{jk}^2 +\sigma_{jk}^2 - \log(\sigma_{jk}^2) - 1 \Big) \\
    & \displaystyle  + \beta_r \sum\limits_{j=1}^N \Big\|\by_j - \mathcal{NN}_e\Big( \mathcal{NN}_d(\by_j, \bw_j; \bphi_d); \bphi_e^{\text{opt}} \Big) \Big\|_2^2 \Bigg],
\end{split}
    \label{equ: nn parameters opt}
\end{equation}
which combines a MSE loss for the emulator $\mathcal{NN}_e$, a MLE loss for the normalizing flow-based density estimator $\mathcal{NN}_f$, and a weighted sum of MSE and KL-divergence losses for the variational encoder $\mathcal{NN}_v$ and the decoder $\mathcal{NN}_d$ combined.
These losses for the variational encoder-decoder are associated with weights $\beta_v$, $\beta_d$, and $\beta_r$, respectively that are chosen to ensure that each loss is minimized with its due priority irrespective of its magnitude. 

We utilize a Real-NVP (real-valued non-volume preserving transformation) discrete normalizing flow architecture~\cite{dinh2017density}, which consists of $L$ bijections defined as
$$\mathcal{NN}_f(\bz^{(0)}; \bphi_f) = (g_L \circ \cdots \circ g_2 \circ g_1)( \bz^{(0)} ), \quad \bz^{(0)} \sim \pi_0(\bz),$$
where $\bz^{(l)} = g_{l} (\bz^{(l-1)}), \bz^{(l-1)} = g_{l}^{-1} (\bz^{(l)}), l = 1,\dots,L$, and $\bz^{(L)} \approx \by \sim p(\by)$.
Additional details on approaches for discrete and continuous normalizing flows can be found in~\cite{papamakarios2021normalizing,kobyzev2020normalizing}.

Generation of input-dependent latent variables $\bw \sim p(\bw|\bv)$ follows the classical VAE framework~\cite{kingma2013auto}, i.e.,
\begin{equation}
    \mathcal{NN}_v(\bv; \bphi_v) = \big[\bmu, \log \bsigma^2\big]^T, \quad \bw = \bmu + \bepsilon \odot \bsigma, \quad \bepsilon \sim \mathcal{N}(\boldsymbol{0}, \mathbf{I}) \ .
\end{equation}
The input reconstruction loss ensures that the decoded outputs from $\mathcal{NN}_d$ resemble the parameters that were used as inputs to the variational encoder $\mathcal{NN}_v$.
The KL-divergence regularizes the latent space $\bm{\mathcal{W}}$ by encouraging the posterior distribution $p(\bw|\bv)$ to be statistically close to a standard normal prior $p(\bw)$.
Finally, the output reconstruction loss (sometimes referred to as the \emph{re-evaluation} loss~\cite{tong2024invaert}) uses the previously trained emulator, i.e., $\mathcal{NN}_e(\cdot, \bphi_e^{\text{opt}})$, to ensure that the decoded parameters map to the correct outputs.

During inversion, the latent variables $\bm{w}$ are typically sampled directly from a standard Gaussian prior and subsequently decoded.
While this is sufficient for simple problems, it may results in suboptimal parameters estimates in more complex cases, particularly if the prior significantly deviates from a standard Gaussian~\cite{tong2024invaert} (e.g., for relatively small values of the $\beta_{v}$ penalty). 
Improved sampling can be achieved using the \emph{predictor-corrector (PC) method}~\cite{tong2024invaert}, summarized in Algorithm~\ref{alg:pc-sampling}.

\begin{algorithm}[H]
\caption{Predictor-corrector (PC) sampling.}
For a given $\bm{y}^*$, sample $\bm{w}^{[0]} \sim \mathcal{N}(\mathbf{0}, \mathbf{I})$ from the latent space. 

Concatenate and decode to produce: $\hat{\bm{v}}^{[0]} = \mathcal{NN}_d(\bm{y}^*, \bm{w}^{[0]})$ \hfill \tcp{Prediction step}

\For{$r = 1, 2, \dots, R$: \hfill \tcp{Correction loop}}{
    VAE encode: $[\boldsymbol{\mu}^{[r]}, \boldsymbol{\sigma}^{[r]}] = NN_v(\hat{\bm{v}}^{[r-1]})$ \;
    Assign mean: $\bm{w}^{[r]} = \bm{\mu}^{[r]}$
    
    Concatenate and decode: $\hat{\bm{v}}^{[r]} = \mathcal{NN}_d(\bm{w}^{[r]}, \bm{y}^*)$ 
}
\Return{$\hat{\bm{v}}^{[R]}$}
\label{alg:pc-sampling}
\end{algorithm}

% =================================
\subsection{Emulator architectures}
% =================================

Henceforth, we shall consider our forward process to be an autonomous dynamical system of the form
\[
\frac{d\bm{y}}{dt} = \bm{\mathcal{F}}(\bm{y}; \bm{k}); \quad \bm{y}(0)=\bm{y}_0.
\]
The emulator $\mathcal{NN}_e$ serves as a neural network-based surrogate for the forward map $\bm{v}=(\bm{k},t) \mapsto \bm{y}$. 
In addition to $\bm{v}$, we also provide $\boldsymbol{y}_{\text{aux}}\in \boldsymbol{\mathcal{D}}_v$ as auxiliary data, i.e., $n_p$ steps of historical output data $\bm{y}$ (see Figure~\ref{fig:delta-t-explanation}) to the emulator, which is trained to learn the map $\bm{\mathcal{V}} \oplus \bm{\mathcal{D}}_v \rightarrow \bm{\mathcal{Y}}$, where  $\oplus$ indicates concatenation.
In Section~\ref{subsubsec:resnet} and Section~\ref{subsubsec:lstm}, we introduced two types of emulator architectures used in our numerical tests.

% =====================================================
\subsubsection{ResNet emulator}\label{subsubsec:resnet}
\begin{figure}[!ht]
\centering
    \begin{tikzpicture}[
    >=Stealth,
    node distance=1.8cm,
    every node/.style={font=\large},
    nn/.style={
  thick,
  align=center,
  minimum height=1.5cm,
  minimum width=1.3cm,
  path picture={
    \path[fill=cyan!5] 
      (path picture bounding box.north west) --
      (path picture bounding box.south west) --
      ([yshift=0.20cm]path picture bounding box.south east) --
      ([yshift=-0.20cm]path picture bounding box.north east) --
      cycle;
    \draw[thick]
      (path picture bounding box.north west) --
      (path picture bounding box.south west) --
      ([yshift=0.20cm]path picture bounding box.south east) --
      ([yshift=-0.20cm]path picture bounding box.north east) --
      cycle;
  }},
    nnflip/.style={
  thick,
  align=center,
  minimum height=1.5cm,
  minimum width=1.3cm,
  path picture={
    \path[fill=cyan!5] 
      ([yshift=-0.20cm]path picture bounding box.north west) --
      ([yshift=0.20cm]path picture bounding box.south west) --
      (path picture bounding box.south east) --
      (path picture bounding box.north east) --
      cycle;
    \draw[thick]
      ([yshift=-0.20cm]path picture bounding box.north west) --
      ([yshift=0.20cm]path picture bounding box.south west) --
      (path picture bounding box.south east) --
      (path picture bounding box.north east) --
      cycle;
  }},
    nndown/.style={
  thick,
  align=center,
  minimum height=1.2cm,
  minimum width=1.5cm,
  path picture={
    \path[fill=cyan!5] 
      (path picture bounding box.north west) --
      ([xshift=0.20cm]path picture bounding box.south west) --
      ([xshift=-0.20cm]path picture bounding box.south east) --
      (path picture bounding box.north east) --
      cycle;
    \draw[thick]
      (path picture bounding box.north west) --
      ([xshift=0.20cm]path picture bounding box.south west) --
      ([xshift=-0.20cm]path picture bounding box.south east) --
      (path picture bounding box.north east) --
      cycle;
  }},
    block/.style={draw, rectangle, minimum height=1cm, minimum width=1.4cm, thick},
    arrow/.style={->, thick}
]
    
    % Left block: inputs
    \node at (0,0) (v-plus-Dv) {
    $\begin{array}{c}
    \boldsymbol{\mathcal{V}} \\ \oplus \\ \boldsymbol{\mathcal{D}}_v
    \end{array}
    \ni
    \left(
    \begin{array}{c}
    \boldsymbol{v} \\[2pt] \boldsymbol{y}_{aux}
    \end{array}
    \right)$};
    
    % Middle block: decomposition
    \node[right=0.2cm of v-plus-Dv] (middle-block) {
    $=
    \begin{array}{c}
    \left(
    \begin{array}{c}
    \boldsymbol{k} \\[2pt] t
    \end{array}
    \right) \\[4pt]
    \oplus \\[4pt]
    \left(
    \begin{array}{c}
    \boldsymbol{y}(t-n_p\Delta t) \\[2pt]
    \vdots \\[2pt]
    \boldsymbol{y}(t-2\Delta t) \\[2pt]
    \boldsymbol{y}(t-\Delta t)
    \end{array}
    \right)
    \end{array}$};
    
    % Encoder box
    \node[nn, right=1cm of middle-block] (nne) {$\mathcal{NN}_e$};
    \draw[arrow] (middle-block) -- ([xshift=0.3cm]nne);
    
    % Arrow to prediction
    \node[right=1cm of nne] (residual) {$\Delta \hat{\boldsymbol{y}}(t)$};
    \draw[arrow] (nne) -- (residual);
    
    % Vertical and horizontal arrows (feedback)
    \draw[arrow] (middle-block)
    -- ++(0, -2.7cm)
    -- ++(6cm,0);
    
    % Labels for feedback
    \node[below=1.2cm of nne, xshift=-0.5cm] (skip-connection) {$+\ \boldsymbol{y}(t-\Delta t)$};
    \node[below=3cm of residual] (output) {$\hat{\boldsymbol{y}}(t)$};
    \draw[arrow] (residual) -- (output);
    
    \end{tikzpicture}
\caption{Schematic of the ResNet emulator $\mathcal{NN}_e$.}
\label{fig:ResNet_schematic}
\end{figure}
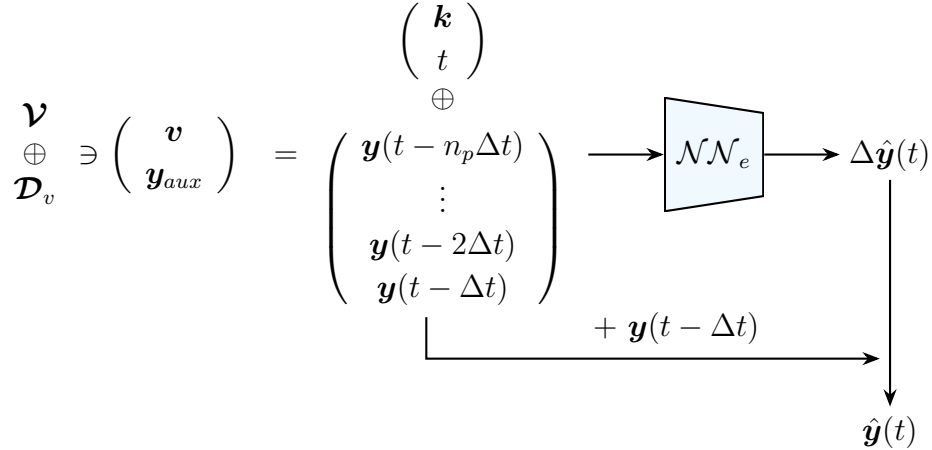

% =====================================================
%  
The emulator based on ResNet~\cite{he2016deep,qin2019data} is essentially a generalization of the forward Euler method, expressed as
\begin{center}
    $\boldsymbol{y}(t+\Delta t) = \boldsymbol{y}(t) + \Delta\bm{y}(t) = \boldsymbol{y}(t) +{\underbrace{\int_t^{t+\Delta t}\boldsymbol{y}'(s)\, ds}_{\text{Residual}}}$,
\end{center}
where the change in $\boldsymbol{y}$ going from time $t$ to time $t+\Delta t$ is known as the \emph{residual} $\Delta\bm{y}(t)$. 
The ResNet emulator technically learns the residual at each time-step conditioned on the parameters $\bm{k}$ of the system, and the predicted residual added to the input gives the prediction for the next time-step. 
The schematic in Figure~\ref{fig:ResNet_schematic} shows the architecture of the ResNet emulator.
Given a dataset $\mathcal{D} = \big\{(\boldsymbol{v}^{(j)}, \boldsymbol{y}^{(j)}(t),\boldsymbol{y}^{(j)}(t-\Delta t))\big\}_{j=1}^{N}$, optimal weights and biases for $\mathcal{NN}_e$ are obtained by minimizing a loss function of the form
\[
\bm{\phi}_e^{\text{opt}} = \displaystyle \operatorname*{arg\,min}\limits_{\bphi_e} \sum\limits_{j=1}^N \Big\|\boldsymbol{y}^{(j)}(t) - ( \boldsymbol{y}^{(j)}(t-\Delta t) + \mathcal{NN}_e(\boldsymbol{v}^{(j)}; \bphi_e)) \Big\|_2^2.
\]

% ==========================================================================
\subsubsection{Long-short term memory (LSTM) emulator}\label{subsubsec:lstm}
% ==========================================================================

The residuals in the emulator are dependent on $\Delta t$, hence we are required to provide data with constant time-step during training. However, on account of the stiffness of our systems, we often have to choose a very small $\Delta t$ to capture the trends in the data. 
Consequently, errors may accumulate as we perform predictions autoregressively (defined as \emph{rollout} in Section~\ref{sec:emu_res}) over a large number of time steps using the ResNet based emulator. 
To address this limitation and improve long-term prediction stability while mitigating issue due to vanishing and exploding gradient, we implement an LSTM encoder-decoder emulator~\cite{hochreiter1996lstm}. 
LSTMs are used primarily for sequence to sequence mapping, that is, we provide $n_p$ steps of historical $\bm{y}$-data, and predict $n_f$ future values at once.

\begin{figure}[!ht]
\centering
    \begin{tikzpicture}[
    box/.style={rectangle, draw=black, fill=cyan!5, thick, minimum height=1.2cm, minimum width=3.2cm, align=center},
    arrow/.style={->, thick},
    arrowdashed/.style={
      arrow,
      dashed
    },
    boxdotted/.style={
      rectangle,
      draw=black,
      dotted,
      thick,
      minimum width=2cm,
      minimum height=2.2cm
    },
    nndown/.style={
      thick,
      align=center,
      minimum height=1.2cm,
      minimum width=3.4cm,
      path picture={
        \path[fill=cyan!5] 
          (path picture bounding box.north west) --
          ([xshift=0.40cm]path picture bounding box.south west) --
          ([xshift=-0.40cm]path picture bounding box.south east) --
          (path picture bounding box.north east) --
          cycle;
        \draw[thick]
          (path picture bounding box.north west) --
          ([xshift=0.40cm]path picture bounding box.south west) --
          ([xshift=-0.40cm]path picture bounding box.south east) --
          (path picture bounding box.north east) --
          cycle;
      }}]
    
    \draw node at (0,0) (input) {$
    \left( \bm{k}, \hspace{2mm}
    \begin{array}{c}
    \boldsymbol{Y}(t-n_p\Delta t) \\[2pt]
    \vdots \\[2pt]
    \boldsymbol{Y}(t-2\Delta t) \\[2pt]
    \boldsymbol{Y}(t-\Delta t)
    \end{array}
    \right) \in \boldsymbol{\mathcal{V}} \oplus \boldsymbol{\mathcal{D}}_v
    $};

    \draw node at (-0.55,1) (node-y) {};

    \node[boxdotted] at (-0.55,0) (dotted-box) {};

    \draw[arrow] (node-y)
                -- ++(0,0.5cm)
                -- ++(3.85cm,0);

    \draw node at (3.6, -0.5) {$\bm{h}$};
    \draw node at (6.2, -0.5) {$\bm{c}$};
    \draw node at (0.9, -1.5) {$\oplus \bm{k}$};
    \draw node at (3.6, -3.2) {$\tilde{\bm{h}}$};
    \draw node at (2.1, -5.05) {$\oplus \bm{k}$};
    \draw node at (2.1, -6.25) {$\oplus \bm{k}$};

    \node[box]  at (5.1,1.2) (Encoder) {Bi-directional LSTM\\Encoder};
    \node[box, below=4.2cm of Encoder] (Decoder) {LSTM Decoder};
    \node[box, right=2.6cm of Decoder] (Decoder2) {LSTM Decoder};
    \node[box, below=2cm of Decoder2] (Decoder3) {LSTM Decoder};

    \node[nndown] at (3.8, -2.3) (encoder-cond) {Hidden state\\conditioning};

    \draw[arrow] (3.8, 0.6) -- (encoder-cond); % h arrow
    \draw[arrow] (6.4, 0.6) -- (6.4, -3.6); % c arrow
    \draw[arrow] (-2.15, -0.3)
        -- ++(0, -1cm)
        -- ++(5.9cm, 0); % hidden state conditioning arrow
    \draw[arrow] (encoder-cond)
        -- ++(0, -1.3cm); % h tilde arrow
    \draw[arrow] (-0.7, -1.3)
        -- ++(0, -2.95cm)
        -- ++(4.15cm, 0); % first decoder conditioning arrow

    \draw node at (1.3, -4) {$\boldsymbol{Y}(t-\Delta t) \oplus \bm{k}$};
    \draw node at (8.3, -4.2) (y-pred-1) {$\widehat{\boldsymbol{Y}}(t)$};
    \node[below=0.8cm of Decoder2] (y-pred-secondlast) {$\widehat{\boldsymbol{Y}}(t+(n_f-2)\Delta t)$};
    \node[below=0.4cm of Decoder3] (y-pred-last) {$\widehat{\boldsymbol{Y}}(t+(n_f-1)\Delta t)$};

    \draw[arrow] (Decoder)--(y-pred-1);
    \draw[arrow] (y-pred-1)--(Decoder2);
    \draw[arrow] (-2.15, -0.3)
    -- ++(0, -5cm)
    -- ++(11.1cm, 0)
    -- ++(0, 1.05cm); % second last decoder conditioning arrow
    \draw[arrow] (-2.15, -0.3)
    -- ++(0, -6.2cm)
    -- ++(13.05cm, 0); % last decoder conditioning arrow

    \draw[arrowdashed] (Decoder2)--++(0, -1.6cm);
    \draw[arrow] (y-pred-secondlast)--++(0, -0.85cm);
    \draw[arrow] (Decoder3)--(y-pred-last);
    
    \end{tikzpicture}
\caption{Schematic of the LSTM emulator $\mathcal{NN}_e$. The quantities $\bm{c}$ and $\bm{h}$ represent cell and hidden states, respectively.}
\label{fig:LSTM_schematic}
\end{figure}
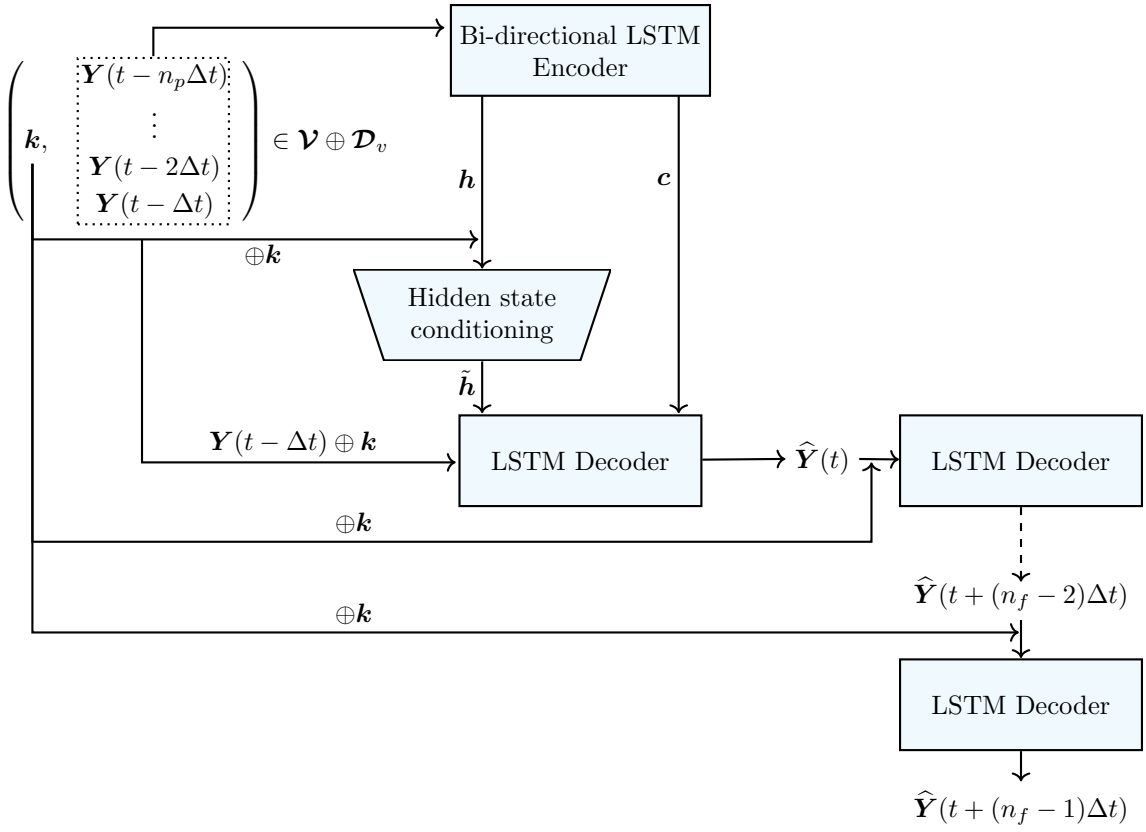

An LSTM encoder-decoder network consists of a conditional encoder and a conditional decoder.
The first is a multi-layer bi-directional LSTM with a feed-forward network that encodes the parameter information into the hidden state, whereas the second is a unidirectional LSTM decoder with additional conditioning. 
The model architecture is motivated by~\cite{lstm-encoder-decoder} and~\cite{grayson-lstm} with hidden state conditioning inspired by~\cite{cond-lstm} and a re-weighted $L^2$ loss function similar to~\cite{jagtap2024amore} to facilitate the conditioning process.
A schematic of this architecture is shown in Figure~\ref{fig:LSTM_schematic}.

The bi-directional LSTM encoder takes in $n_p$ time steps of data in the form of $\bm{Y}=(\bm{y}, t)$ pairs, and passes on the cell state and hidden states onto the LSTM decoder.
Following the last LSTM layer, a conditional predictor is obtained by concatenating the parameters $\bm{k}$ with the encoded hidden state, and encoding this latter into an updated hidden state through a feed-forward network.
The conditional hidden state and cell state are then passed on to the LSTM decoder. 
The decoder takes in the final step data $\bm{y}(t - \Delta t)$ and augments it with the associated $\bm{k}$ to predict $\bm{y}$ for the next $n_f$ time steps, recursively.
Finally, the outputs from the decoder are compared to the true outputs using the following weighted loss function at iteration $N$
\begin{equation}
\mathcal{L}_{N}(\bm{y}, \widehat{\bm{y}}) = \frac{1}{B_{N}\,n_{f}} \sum_{k=1}^{n_{f}} \sum_{j=1}^{B_{N}} \sum_{i=1}^{N_{\bm{y}}} \widetilde{W}_{N,i}\,\left(y^{(j)}_{i,k} - \widehat{y}^{(j)}_{i,k}\right)^2,
\end{equation}
where $N_{\bm{y}}=dim(\bm{y})$, $B_{N}$ the mini-batch size at step $N$, $y^{(j)}_{i,k}$ represents the $i$-th species concentration at the $j$-th sample for the $k$-th future estimate, and the weights are initiated as $\widetilde{W}_{0,i}=1,\,i=1,\dots,N_{\bm{y}}$.
At the end of each mini-batch, the weights are updated based on the values of the  component-wise losses at the previous batch as
\begin{equation}
\widetilde{W}_{N,i}=\frac{[1/(B_{N-1}\,n_{f})]\sum_{k=1}^{n_{f}}\sum_{j=1}^{B_{N-1}} \lvert y_{i,k}^{(j)} - \widehat{y}^{(j)}_{i,k}\rvert}{\sum_{i=1}^{N_{\bm{y}}}\,[1/(B_{N-1}\,n_{f})] \sum_{k=1}^{n_{f}}\sum_{j=1}^{B_{N-1}}  \lvert y^{(j)}_{i,k} - \widehat{y}^{(j)}_{i,k}\rvert}\cdot N_{\bm{y}},
\end{equation}
while preserving the sum of the weights as $\sum_{i=1}^{N_{\bm{y}}}\,\widetilde{W}_{N,i} = N_{\bm{y}}$. The $\mathcal{NN}_e$ model parameters $\bphi_e$ are obtained after optimizing the loss function defined above as
\[
\phi_e^{\text{opt}} = \displaystyle \operatorname*{arg\,min}\limits_{\bphi_e} \mathcal{L}_{N}(\bm{y}, \widehat{\bm{y}};\bphi_e).
\]
Finally, note that this framework is entirely data-driven and does not explicitly enforce any conservation laws.

% =====================================
\section{Results}\label{subsec:results}
% =====================================

% =======================================
\subsection{Emulation}\label{sec:emu_res}
% =======================================

In order to evaluate the performance of the emulators, we use two different metrics.
We first consider a \emph{single-step RMSE}, or the relative root mean-squared errors for single-step prediction obtained from the emulator for the $j$-th validation sample.
Given the $j$-th validation inputs $(\bm{k}^{(j)}, t^{(j)})$, if the true concentration of the $i$-th species is $y_i^{(j)}(t^{(j)};\bm{k}^{(j)})=y_i^{(j)}$ among the $N_{\bm{y}}$ species in the system, and the corresponding model prediction $\widehat{y}_{i}^{(j)}(t^{(j)};\bm{k}^{(j)}) = \widehat{y}_{i}^{(j)}$, then we compute
\begin{equation}\label{rmse_resnet}
\text{RMSE}^{(j)} = \sqrt{\frac{1}{N_{\bm{y}}}\sum_{i=1}^{N_{\bm{y}}}\,\left[y^{(j)}_{i} -\widehat{y}^{(j)}_{i}\right]^2},\,\,\text{relative RMSE}^{(j)} = \frac{\sqrt{\frac{1}{N_{\bm{y}}}\sum_{i=1}^{N_{\bm{y}}}\left[y^{(j)}_{i} -\widehat{y}^{(j)}_{i}\right]^2}}{\sqrt{\frac{1}{N_{\bm{y}}}\sum_{i=1}^{N_{\bm{y}}}\,\left[y^{(j)}_{i}\right]^2}}.
\end{equation}
The expressions~\eqref{rmse_resnet} hold for a ResNet emulator since predictions are limited to a single step in the future. In case of an LSTM emulator, we compute $n_f$ predictions at a time. So we compute instead, the mean RMSE and mean relative RMSE over all $n_f$ predictions. If the emulator predicts $n_f$ steps into the future, and $y^{(j,r)}, \widehat{y}^{(j,r)}$ are true and predicted outputs for input $(t^{(j)}+r\Delta t;\bm{k}^{(j)})$ then, we have
\[
\text{RMSE}^{(j)} = \frac{1}{n_f}\sum_{r=1}^{n_f}\sqrt{\frac{1}{N_{\bm{y}}}\sum_{i=1}^{N_{\bm{y}}}\,\left[y^{(j,r)}_{i} -\widehat{y}^{(j,r)}_{i}\right]^2},\,\,\text{relative RMSE}^{(j)} = \frac{1}{n_f}\sum_{r=1}^{n_f}\frac{\sqrt{\frac{1}{N_{\bm{y}}}\sum_{i=1}^{N_{\bm{y}}}\left[y^{(j,r)}_{i} -\widehat{y}^{(j,r)}_{i}\right]^2}}{\sqrt{\frac{1}{N_{\bm{y}}}\sum_{i=1}^{N_{\bm{y}}}\,\left[y^{(j,r)}_{i}\right]^2}}.
\]
Thereafter, we report the mean RMSE and mean relative RMSE over the entire validation dataset.

\begin{figure}[!ht]
\centering
    \begin{tikzpicture}[
    box/.style={rectangle, draw=black, fill=cyan!5, thick, minimum height=1.2cm, minimum width=3.2cm, align=center},
    arrowcyan/.style={->, draw=BrightBlue, thick},
    linedashed/.style={ -, draw=magenta, thick, dashed },
    boxcyan0/.style={
      rectangle,
      draw=BrightBlue,
      dashed,
      thick,
      minimum width=7cm,
      minimum height=0.7cm
    },
    boxcyan1/.style={
      rectangle,
      draw=BrightBlue,
      dashed,
      thick,
      minimum width=7.8cm,
      minimum height=1.4cm
    },
    boxcyan2/.style={
      rectangle,
      draw=BrightBlue,
      dashed,
      thick,
      minimum width=9cm,
      minimum height=2.1cm
    },]

    \draw node at (0, 0) (y-s) {$
    \boldsymbol{y}(0) \quad
    \boldsymbol{y}(\Delta t) \quad
    \boldsymbol{y}(2\Delta t) \quad
    \cdots \quad
    \boldsymbol{y}((n_p-1)\Delta t) \quad
    \boldsymbol{\hat{y}}(n_p\Delta t) \quad
    \boldsymbol{\hat{y}}((n_p+1)\Delta t) \quad
    \boldsymbol{\hat{y}}((n_p+2)\Delta t) \quad \cdots$};

    \node[circle,fill=BrightBlue,inner sep=0pt,minimum size=5pt] (A) at (-6.8,-0.35) {};
    \node[circle,fill=BrightBlue,inner sep=0pt,minimum size=5pt] (B) at (-5.6,-0.35) {};
    \node[circle,fill=BrightBlue,inner sep=0pt,minimum size=5pt] (C) at (-4.3,-0.35) {};
    \node[circle,fill=BrightBlue,inner sep=0pt,minimum size=5pt] (D) at (-1.4,-0.35) {};
    \node[circle,fill=magenta,inner sep=0pt,minimum size=5pt] (E) at (0.6,-0.35) {};
    \node[circle,fill=magenta,inner sep=0pt,minimum size=5pt] (F) at (2.7,-0.35) {};
    \node[circle,fill=magenta,inner sep=0pt,minimum size=5pt] (G) at (5.2,-0.35) {};

    \draw [->] (A) -- ++(0, -0.5) -- ([yshift=-0.5cm]E.center) -- (E);
    \draw [-] (B) -- ++(0, -0.5);
    \draw [-] (C) -- ++(0, -0.5);
    \draw [-] (D) -- ++(0, -0.5);

    \node[circle,fill=BrightBlue,inner sep=0pt,minimum size=5pt] (B1) at (-5.6,-1) {};
    \node[circle,fill=BrightBlue,inner sep=0pt,minimum size=5pt] (C1) at (-4.3,-1) {};
    \node[circle,fill=BrightBlue,inner sep=0pt,minimum size=5pt] (D1) at (-1.4,-1) {};
    \node[circle,fill=BrightBlue,inner sep=0pt,minimum size=5pt] (E1) at (0.6,-1) {};

    \draw [->] (B1) -- ++(0, -0.5) -- ([yshift=-1.15cm]F.center) -- (F);
    \draw [-] (C1) -- ++(0, -0.5);
    \draw [-] (D1) -- ++(0, -0.5);
    \draw [-] (E1) -- ++(0, -0.5);
    
    \node[circle,fill=BrightBlue,inner sep=0pt,minimum size=5pt] (C2) at (-4.3,-1.65) {};
    \node[circle,fill=BrightBlue,inner sep=0pt,minimum size=5pt] (D2) at (-1.4,-1.65) {};
    \node[circle,fill=BrightBlue,inner sep=0pt,minimum size=5pt] (E2) at (0.6,-1.65) {};
    \node[circle,fill=BrightBlue,inner sep=0pt,minimum size=5pt] (F2) at (2.7,-1.65) {};

    \draw [->] (C2) -- ++(0, -0.5) -- ([yshift=-1.8cm]G.center) -- (G);
    \draw [-] (D2) -- ++(0, -0.5);
    \draw [-] (E2) -- ++(0, -0.5);
    \draw [-] (F2) -- ++(0, -0.5);
    
    % \node[boxcyan0] at (-3.8,0) (input-0) {};
    % \node[below=0.2pt of input-0] {{\color{BrightBlue}Auxiliary window 0}};

    % \node[boxcyan1] at (-2.35,-0.2) (input-1) {};
    % \node[below=0.2pt of input-1] {{\color{BrightBlue}Auxiliary window 1}};

    % \node[boxcyan2] at (-0.55,-0.4) (input-2) {};
    % \node[below=0.2pt of input-2] {{\color{BrightBlue}Auxiliary window 2}};

    % \draw[linedashed] (-0,-0.27) -- node[midway, below=0.6pt] {\color{magenta}Prediction 1} (1.2,-0.27);

    % \draw[linedashed] (1.6,-0.27) -- node[midway, below=0.6pt] {\color{magenta}Prediction 2} (3.75,-0.27);

    % \draw[linedashed] (4.1,-0.27) -- node[midway, below=0.6pt] {\color{magenta}Prediction 3} (6.25,-0.27);
    
    \end{tikzpicture}
\caption{Rollout mechanism for ResNet architecture. Blue and red dots indicate input data and predictions, respectively.}
\label{fig:rollout_schematic_resnet}
\end{figure}
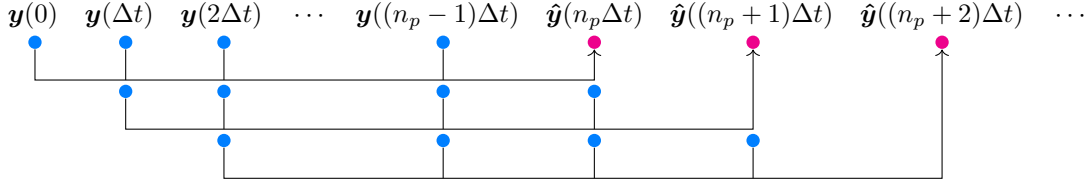

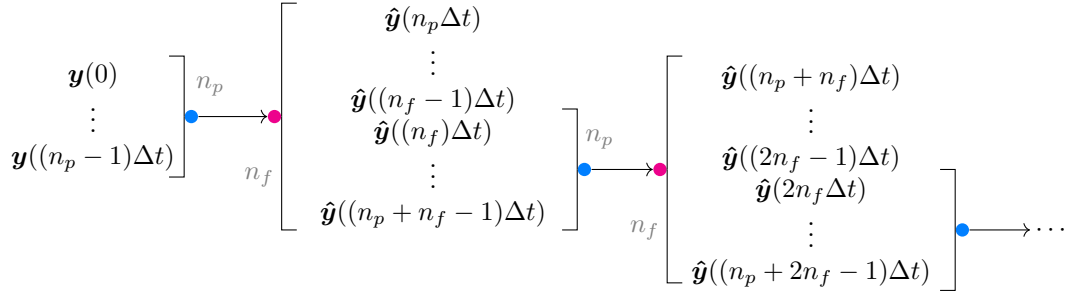
\begin{figure}[!ht]
\centering
    \begin{tikzpicture}[
    box/.style={rectangle, draw=black, fill=cyan!5, thick, minimum height=1.2cm, minimum width=3.2cm, align=center},
    arrowcyan/.style={->, draw=BrightBlue, thick},
    arrowdashed/.style={
      arrow,
      dashed
    },
    boxcyan0/.style={
      rectangle,
      draw=BrightBlue,
      dashed,
      thick,
      minimum width=2.3cm,
      minimum height=2.1cm
    },
    boxcyan1/.style={
      rectangle,
      draw=BrightBlue,
      dashed,
      thick,
      minimum width=3.2cm,
      minimum height=1.6cm
    },
    boxcyan2/.style={
      rectangle,
      draw=BrightBlue,
      dashed,
      thick,
      minimum width=3.3cm,
      minimum height=1.65cm
    },
    boxmagenta1/.style={
      rectangle,
      draw=magenta,
      dashed,
      thick,
      minimum width=3.5cm,
      minimum height=3.3cm
    },
    boxmagenta2/.style={
      rectangle,
      draw=magenta,
      dashed,
      thick,
      minimum width=3.7cm,
      minimum height=3.3cm
    },]

    \draw node at (-7, 0) (step-0) {$\begin{matrix} 
    \boldsymbol{y}(0) \\
    \vdots \\
    \boldsymbol{y}((n_p-1)\Delta t)
    \end{matrix}$};

    \draw node at (-2.5, 0) (step-1) {$\begin{matrix} 
    \boldsymbol{\hat{y}}(n_p\Delta t) \\    
    \vdots \\
    \boldsymbol{\hat{y}}((n_f-1)\Delta t) \\
    \boldsymbol{\hat{y}}((n_f)\Delta t) \\
    \vdots \\
    \boldsymbol{\hat{y}}((n_p+n_f-1)\Delta t)
    \end{matrix}$};

    % \node[boxcyan0] at (-7,0) (input-0) {};
    % \node[below=0.2pt of input-0] {{\color{BrightBlue}Auxiliary window 0}};

    \node[circle,fill=BrightBlue,inner sep=0pt,minimum size=5pt] (A) at (-5.7,0.0) {};

    \draw [-] (A.west) -- ++(0, 0.8) -- ++(-0.2, 0);
    \draw [-] (A.west) -- ++(0, -0.8) -- ++(-0.2, 0);    

    % \draw[arrowcyan] (input-0) -- node[midway, above=0.6pt] {} (step-1);

    % \node[boxcyan1] at (-2.5,-0.77) (input-1) {};
    % \node[below=0.2pt of input-1] {{\color{BrightBlue}Auxiliary window 1}};
    % \node[boxmagenta1] at (-2.5,0) (output-1) {};
    % \node[above=0.2pt of output-1] {{\color{magenta}Prediction window 1}};

    \draw node at (2.5, -0.77) (step-2) {$\begin{matrix} 
    \boldsymbol{\hat{y}}((n_p+n_f)\Delta t) \\    
    \vdots \\
    \boldsymbol{\hat{y}}((2n_f-1)\Delta t) \\
    \boldsymbol{\hat{y}}(2n_f\Delta t) \\
    \vdots \\
    \boldsymbol{\hat{y}}((n_p+2n_f-1)\Delta t)
    \end{matrix}$};

    \node[circle,fill=magenta,inner sep=0pt,minimum size=5pt] (B) at (-4.6,0.0) {};

    \draw [-] (B.east) -- ++(0, 1.5) -- ++(0.2, 0);
    \draw [-] (B.east) -- ++(0, -1.5) -- ++(0.2, 0);

    \draw [->] (A.east) -- (B.west);

    %\draw[arrowcyan] (input-1) -- node[midway, above=0.6pt] {} (step-2);
    %\node[boxcyan2] at (2.5,-1.5) (input-2) {};
    %\node[below=0.2pt of input-2] {{\color{BrightBlue}Auxiliary window 2}};
    %\node[boxmagenta2] at (2.5,-0.77) (output-2) {};
    %\node[above=0.2pt of output-2] {{\color{magenta}Prediction window 2}};

    \node[circle,fill=BrightBlue,inner sep=0pt,minimum size=5pt] (C) at (-0.5,-0.7) {};

    \draw [-] (C.west) -- ++(0, 0.8) -- ++(-0.2, 0);
    \draw [-] (C.west) -- ++(0, -0.8) -- ++(-0.2, 0);  

    \node[circle,fill=magenta,inner sep=0pt,minimum size=5pt] (D) at (0.5,-0.7) {};

    \node[text=gray] at (-0.3,-0.3) (np2) {$n_{p}$};
    \node[text=gray] at (0.3,-1.5) (nf2) {$n_{f}$};

    \draw [-] (D.east) -- ++(0, 1.5) -- ++(0.2, 0);
    \draw [-] (D.east) -- ++(0, -1.5) -- ++(0.2, 0);

    \draw [->] (C.east) -- (D.west);

    \node[circle,fill=BrightBlue,inner sep=0pt,minimum size=5pt] (E) at (4.5,-1.5) {};

    \draw [-] (E.west) -- ++(0, 0.8) -- ++(-0.2, 0);
    \draw [-] (E.west) -- ++(0, -0.8) -- ++(-0.2, 0);  

    \draw [->] (E.east) -- ++(0.8, 0);
    
    \node at (5.7,-1.5) (dots) {$\cdots$};
    \node[text=gray] at (-5.45,0.4) (np1) {$n_{p}$};
    \node[text=gray] at (-4.8,-0.8) (nf1) {$n_{f}$};

    % \draw[arrowcyan] (input-2) -- node[midway, above=0.6pt] {} (dots);

    \end{tikzpicture}
\caption{Rollout mechanism for LSTM architecture.}
\label{fig:rollout_schematic_lstm}
\end{figure}

We also consider the error accumulated during \emph{rollout}, where current predictions are treated as auxiliary data for future predictions, in an autoregressive fashion, to assess the performance of the emulator over longer periods of time.
Consider, for instance, the $S$-th emulation step for the proposed LSTM architecture. During such a step, $n_p=10$ time steps of auxiliary historical data are used as inputs to predict the following $n_f=50$ time steps. 
At the $(S+1)$-th step the last $10$ steps out of these $50$ are used as auxiliary data to predict the next $50$ time steps, and so on and so forth.
In case of a ResNet emulator, essentially $n_f=1$. Hence, at each step, we update the auxiliary data with $n_p$ time steps by concatenating the solution obtained in the past $(n_p-1)$ steps with the solution predicted at step $S$.
Figure~\ref{fig:rollout_schematic_resnet} and Figure~\ref{fig:rollout_schematic_lstm} explain this rollout mechanism for a ResNet and LSTM architecture, respectively.

% ===============================
\subsubsection{Robertson problem}
% ===============================

On the time interval $[0,100]$, we generate 1000 time points per simulation, using log sampling. Additionally, the training dataset for the emulator is created by extracting $n_p+n_f$ solutions at equally spaced time points ($\Delta t=10^{-4}$) around each time suggested by log-sampling, see schematic in Figure~\ref{fig:delta-t-explanation}.
The parameter set $\bm{k}$ is obtained by applying random perturbations to the nominal values $\bm{k}^{*}$ in~\eqref{equ:rober_nominal_k}, with each parameter varied within a $\pm 50\%$ range.
We use an LSTM emulator with $n_p=10$ and $n_f=50$.

\begin{figure}[!ht]
\centering
\begin{subfigure}{\textwidth}
\centering
\includegraphics[width=0.95\textwidth]{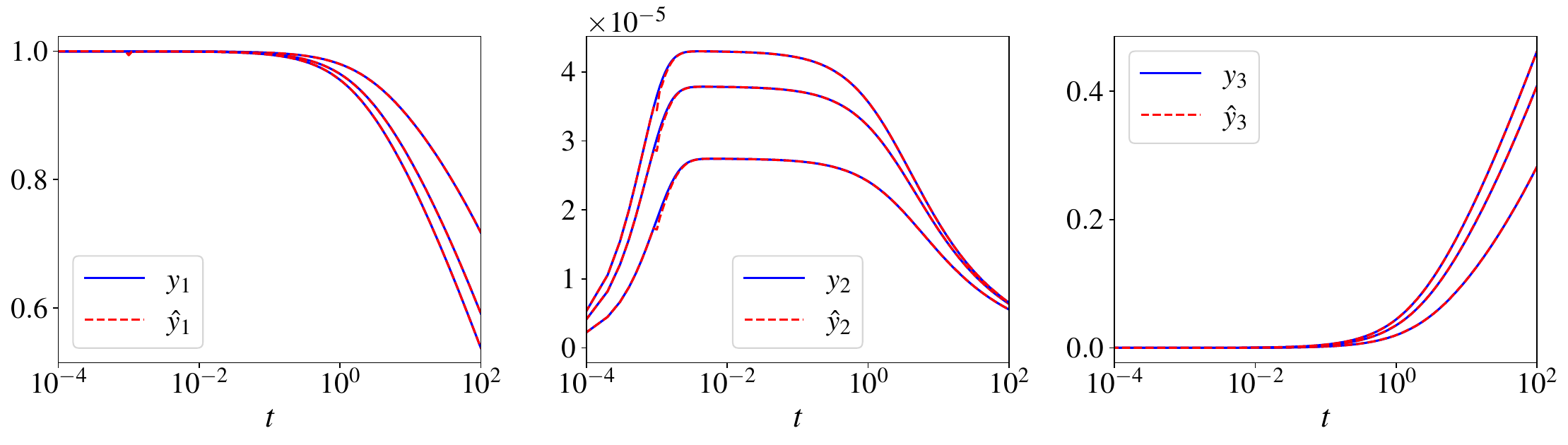}
\caption{Emulator rollout for the Robertson problem for three parameter combinations $\bm{k}$ chosen randomly from the validation set.}
\label{fig:rober-emu-rollout}
\end{subfigure}

\begin{subfigure}{\textwidth}
\centering
\includegraphics[width=\textwidth]{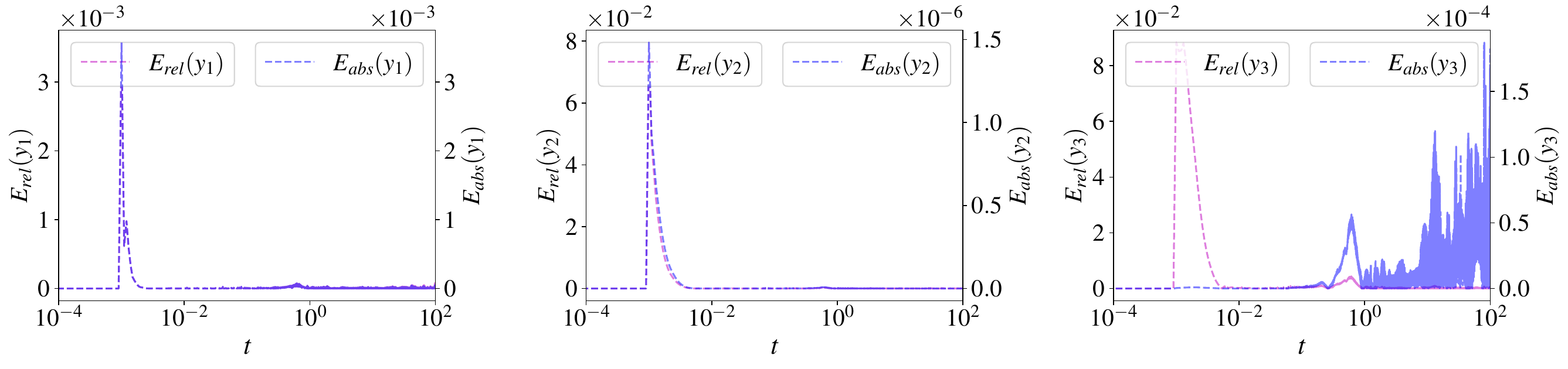}
\caption{Relative and absolute rollout errors for the Robertson problem for one of the $\bm{k}$ used in (a). The errors quantify the distance between the corresponding true and predicted curves shown in (a), and are defined as:  $E_{rel}(y_i)|_{t}= |y_i(t)-\hat{y}_i(t)|/|y_i(t)|$ and $E_{abs}(y_i)|_{t}= |y_i(t)-\hat{y}_i(t)|$ for $i=1,2,3$.}
\label{fig:rober-emu-err}
\end{subfigure}
\caption{Rollout errors for the Robertson problem.}
\label{fig:rober-emu-rollout-err}
\end{figure}

% comment on single step predictions. 
Single-step emulation errors in Table~\ref{tab:rober-emu-single-step} appear negligible. 
% Comment on rollout errors
The rollout errors in Figure~\ref{fig:rober-emu-rollout} are also small with the true and predicted trajectories for the three species practically indistinguishable. However, Figure~\ref{fig:rober-emu-err} shows a spike in both absolute and relative errors at the first prediction step. Since the initial conditions are selected equal for all $\bm{k}$ parameters, and as the $y_{1},y_{3}$ remain initially constant, a minor difference in $y_{2}$ is completely responsible to differentiate the dynamics induced by different coefficients $\bm{k}$. However, as time advances and the trajectories diverge, the LSTM, and in particular the underlying conditioning mechanism is able to correct this initial error.
This means the conditioning becomes ineffective, particularly at the first prediction step, given the nearly identical trajectories over all possible values of $\bm{k}$. 
We confirmed this by switching off the conditioning mechanism. While the error accumulated during rollout increases considerably with time in absence of conditioning, the initial peak remains practically unaltered.

\begin{table}[H]
    \centering
    \begin{tabular}{lc}
    \toprule
    \multicolumn{2}{c}{\bf Robertson problem - Emulation}\\
    \midrule
     Mean RMSE & 3.070464$\times$10$^{-4}$ \\
     Mean Relative RMSE & 3.610673$\times$10$^{-5}$ \\
     \bottomrule
    \end{tabular}
    \caption{Single-step RMSEs for the Robertson problem over a validation dataset with 53000 samples.}
    \label{tab:rober-emu-single-step}
\end{table}

% ===========================
\subsubsection{POLLU problem}
% ===========================

On the time interval $[0,100]$, we generated 2000 time points per simulation, using log-sampling. Additionally, the training dataset for the emulator is created by extracting $n_p+n_f$ solutions at equally spaced time points ($\Delta t=10^{-4}$) around each time suggested by log-sampling, see schematic in Figure~\ref{fig:delta-t-explanation}.
As for the Robertson problem in the previous section, the parameters $\bm{k}$ result from $\pm 50\%$ perturbation around the nominal set $\bm{k}^{*}$ defined in Table~\ref{tab:params_pollu}. 
We still use $n_p=10$ and $n_f=50$.

\begin{figure}[!ht]
\centering
\begin{subfigure}{\textwidth}
\centering
\includegraphics[width=0.95\textwidth]{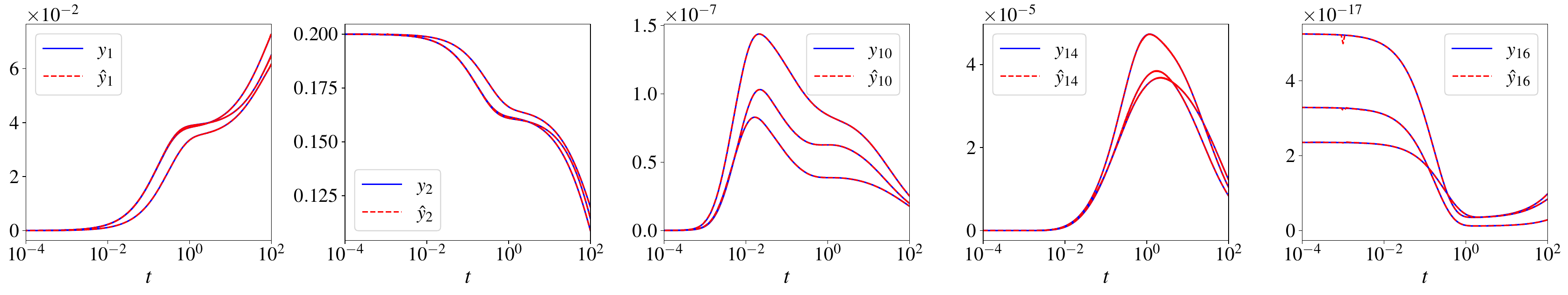}
\caption{Emulator rollout for a subset of the POLLU problem components for three parameter combinations $\bm{k}$ randomly selected from the validation set.}
\label{fig:pollu-emu-rollout}
\end{subfigure}

\begin{subfigure}{\textwidth}
\centering
\includegraphics[width=\textwidth]{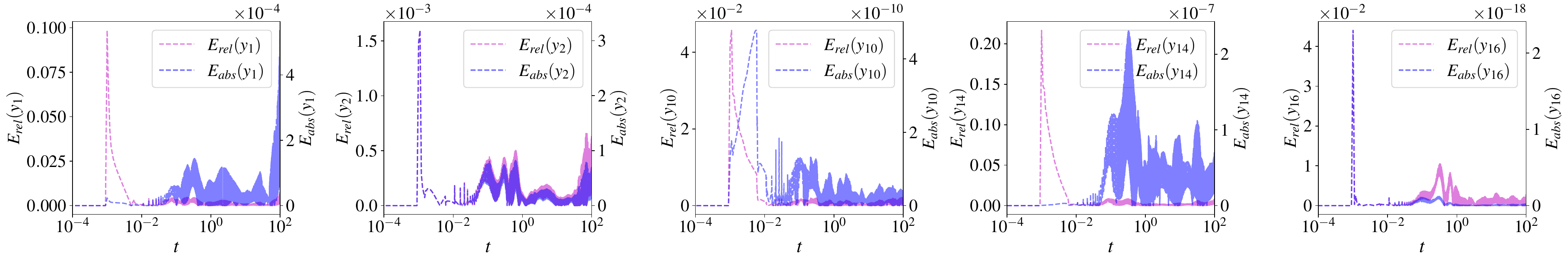}
\caption{Relative and absolute rollout errors for a subset of the species in the POLLU system. We used one of the parameter combinations $\bm{k}$ in (a).}
\label{fig:pollu-emu-err}
\end{subfigure}
\caption{Rollout errors for the POLLU system.}
\label{fig:pollu-emu-rollout-err}
\end{figure}

% comment on single step predictions. 
The higher complexity of the POLLU system leads to single-step emulation errors that are one order of magnitude larger with respect to the Robertson system. This is shown in Table~\ref{tab:pollu-emu-single-step}. 
% Comment on rollout errors
Accumulation errors due to the autoregressive nature of the rollout process are more visible for the POLLU system, but still practically negligible. 
As discussed for the Robertson system, the first prediction step produces large errors (but these are visible only for species with negligible concentrations, as shown in Figure~\ref{fig:pollu-emu-rollout}) which are suddenly reduced as conditioning becomes increasingly effective.

\begin{table}[H]
    \centering
    \begin{tabular}{lc}
    \toprule
    \multicolumn{2}{c}{\bf POLLU problem - Emulation}\\
    \midrule    
     Mean RMSE & 2.089125$\times$10$^{-3}$ \\
     Mean Relative RMSE & 5.260582$\times$10$^{-4}$ \\
     \bottomrule
    \end{tabular}
    \caption{Single-step errors for the POLLU system over a validation dataset with 60000 samples.}
    \label{tab:pollu-emu-single-step}
\end{table}

% ===============================================================
\subsubsection{Systems with Reversible and Irreversible Kinetics}
% ===============================================================

\begin{figure}[!ht]
\centering
\begin{subfigure}{0.39\textwidth}
    \centering
    \includegraphics[width=\textwidth]{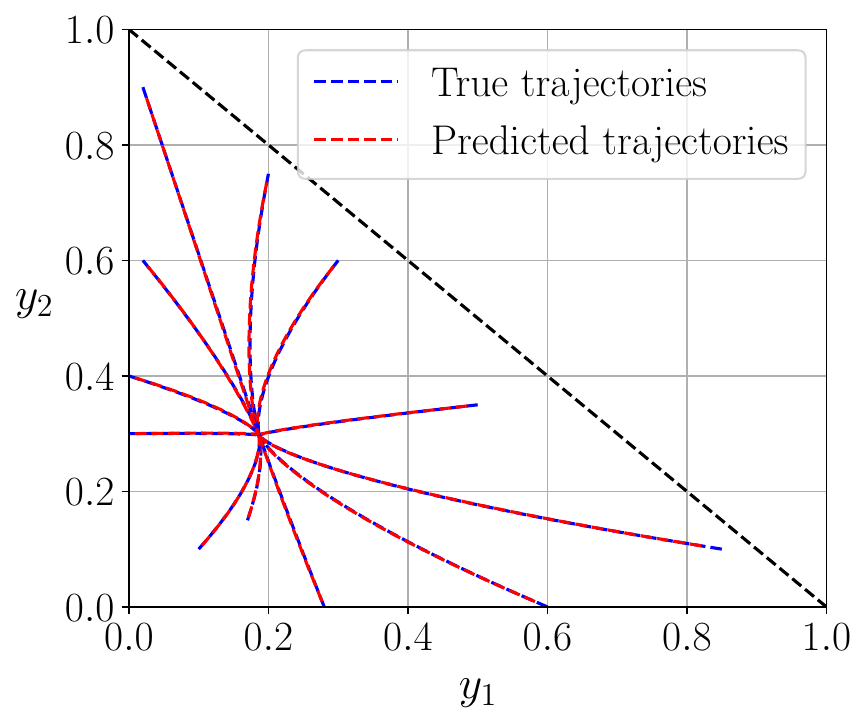}
    \caption{Rollout phase plane trajectories for the reversible system.}
\end{subfigure}
\begin{subfigure}{0.39\textwidth}
    \centering
    \includegraphics[width=\textwidth]{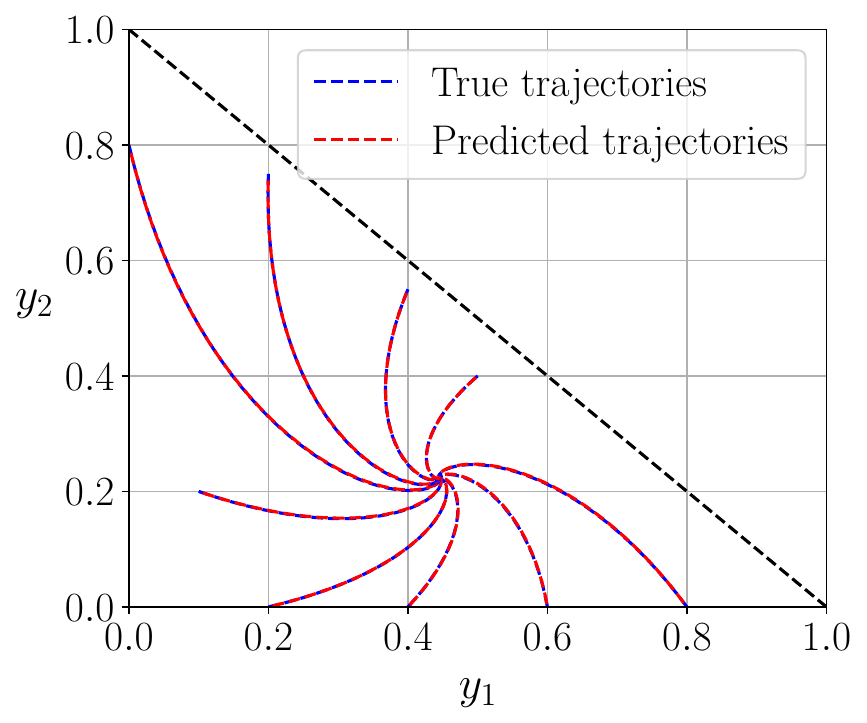}
    \caption{Rollout phase plane trajectories for the irreversible system.}
\end{subfigure}
\caption{Rollout predictions for the reversible and irreversible systems. The figure shows the system's evolution for a randomly selected parameter combination $\bm{k}$ from the validation set, corresponding to different initial conditions. The black dashed line indicates the boundary of the physical domain in the $y_1-y_2$ space where $y_1(t)+y_2(t)\leq 1$, which enables us to have $y_1(t)+y_2(t)+y_3(t)=1$ for non-negative $y_1(t),y_2(t),y_3(t)$'s.}
\label{fig:rev-irr-rollouts}
\end{figure}

The dataset for the emulator was trained using 30 time points in $t\in[0,10]$ per simulation, using log-sampling and $\Delta t=10^{-3}$.
The rate parameters $\bm{k}$ have been generated by $\pm 50\%$ perturbations with respect to the nominal set $\bm{k}^{*}$ defined in \eqref{equ:rev_parameters} and \eqref{equ:irr_parameters}. Additionally, the initial condition $\bm{y}(0)$ was also randomly selected with the constraint of all the components summing to 1.
For this simpler systems we used a ResNet-based emulator with $n_p=10$.
Since the emulator is trained over varying both initial conditions and rate parameters, the input data becomes $\bm{v}=[\bm{k},\,\bm{y}(0),\,t,\, \bm{y}_{\text{aux}}]^{T}$. 
Our model, in this case is trained to handle variations in both $\bm{k}$ and $\bm{y}(0)$ in the respective prior ranges. 
Once a $\bm{k}$ is fixed, by specifying different values of $\bm{y}(0)$, we are able to obtain multiple solution trajectories that start from different locations in the $y_1-y_2$ space. It is thus possible to obtain phase plots at rollout. 
This in turn allows us to determine the steady state of the system given the initial conditions and reaction rate parameters. We present the emulation results in Table~\ref{tab:rev-irr-emu-single-step} and Figure~\ref{fig:rev-irr-rollouts}.

\begin{table}[!ht]
    \centering
    \begin{tabular}{lc}
     \toprule
     \multicolumn{2}{c}{\bf Reversible system - Emulation}\\
     \midrule         
     Mean RMSE & 4.39250$\times$10$^{-6}$\\ 
     Mean Relative RMSE & 1.13695$\times$10$^{-5}$\\
     \midrule         
     \multicolumn{2}{c}{\bf Irreversible system - Emulation}\\
     \midrule         
     Mean RMSE & 2.52039$\times$10$^{-6}$\\
     Mean Relative RMSE & 6.64656$\times$10$^{-6}$\\
     \bottomrule
    \end{tabular}
    \caption{Single-step errors for the reversible and irreversible systems over a validation datasets of 3000 samples.}
    \label{tab:rev-irr-emu-single-step}
\end{table}

% ===========================================
\subsubsection{Hydrogen-air kinetics problem}
% ===========================================

On the time interval $[0,100]$, we have generated 2000 time points per simulation, and constructed the emulator training dataset using a step size $\Delta t=10^{-5}$.
The parameters $\bm{k}$ have been generated by perturbing a nominal set $(\bm{A}, \bm{b}, \bm{E})^{*}$ defined in Table~\ref{tab:water-def} such that $(A, T_0^b, e^{-E/RT_0})$ vary in a $\pm$ 50\% range from the nominal values.
The auxiliary data consists of $n_p=10$ past solutions to predict $n_f=50$ future steps, consistent with the LSTM architecture used for the Robertson and POLLU problems.

\begin{figure}[!ht]
\centering
\includegraphics[width=0.85\textwidth]{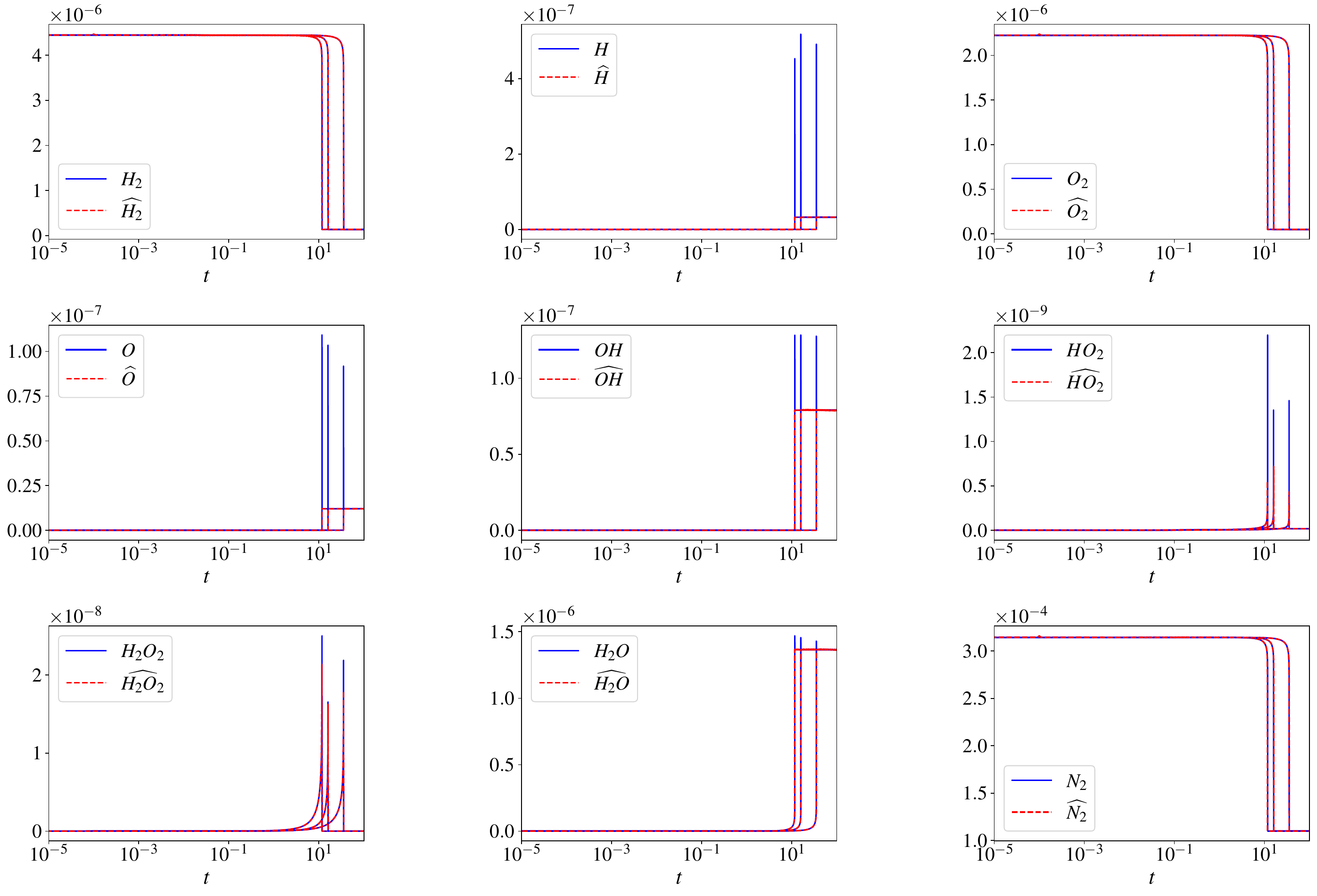}
\caption{LSTM Emulator rollout time traces for the hydrogen-air problem using three different combinations $(\bm{A}, \bm{b}, \bm{E})$, chosen randomly from the validation set.}
\label{fig:water-emu-rollout}
\end{figure}

\begin{figure}[!ht]
\centering
\includegraphics[width=0.85\textwidth]{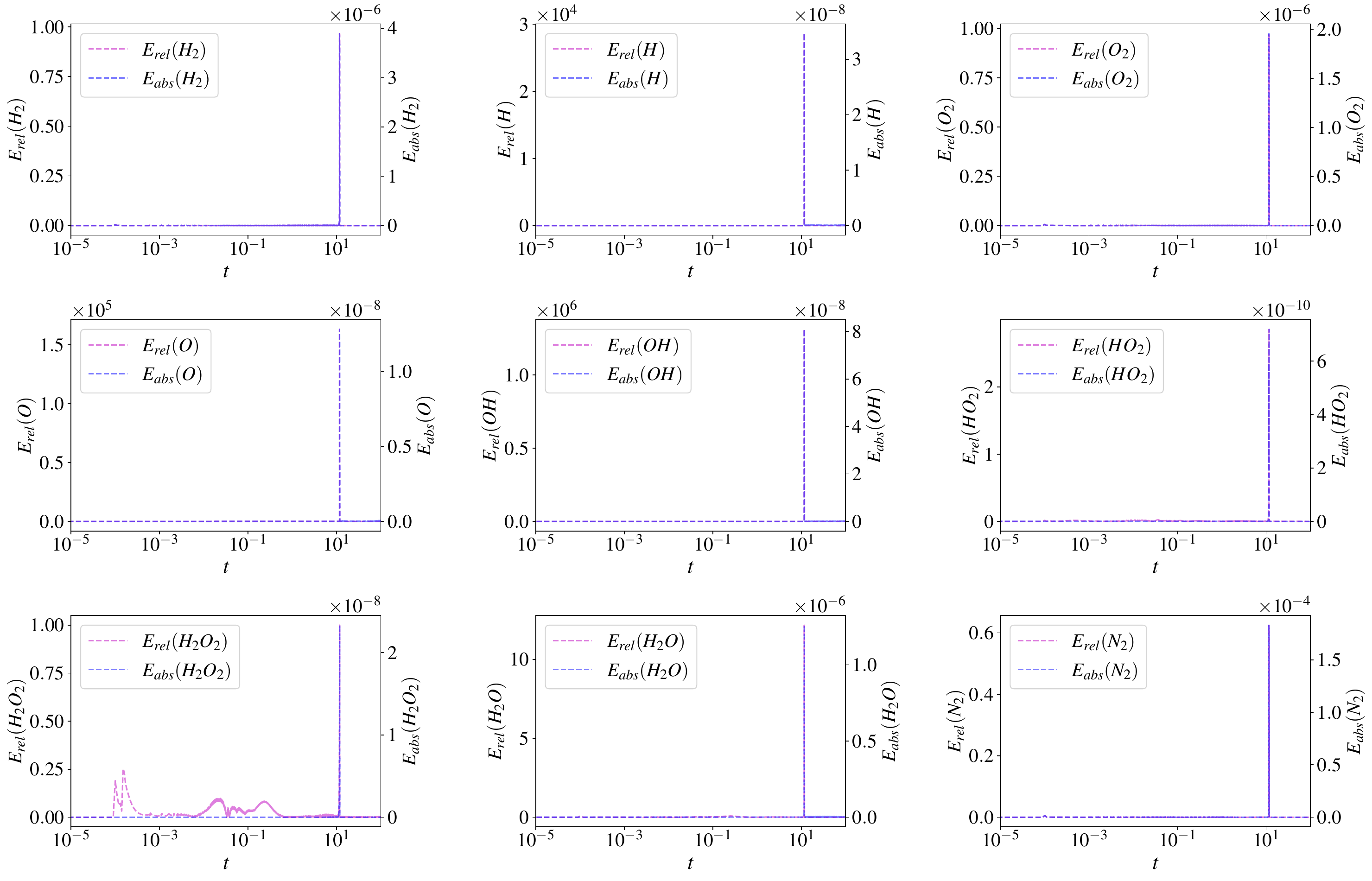}
\caption{Relative and absolute rollout errors in the hydrogen-air system for one of the $(\bm{A}, \bm{b}, \bm{E})$ parameter combinations used in Figure~\ref{fig:water-emu-rollout}.}
\label{fig:water-emu-err}
\end{figure}

Figures~\ref{fig:water-emu-rollout} and \ref{fig:water-emu-err} confirm how the errors are negligible over the entire simulation time. 
However, the concentration time traces are characterized by a sharp peak whose amplitude is not well captured by the LSTM emulator at the point in time where such peak occurs.
As demonstrated in Figure~\ref{fig:water-emu-rollout}, the emulator can correctly predict two quantities of interest for this problem, namely the ignition time and the steady state concentrations, even though it does not have the desired accuracy in capturing the amplitude of each peak.
This relates to the short-lived duration of peak events, much smaller than the selected step size $\Delta t$.

\begin{table}[!ht]
    \centering
    \begin{tabular}{lc}    
    \toprule
    \multicolumn{2}{c}{\bf Hydrogen-air kinetics - Emulation}\\
    \midrule
     Mean RMSE & 3.815623$\times$10$^{-3}$\\
     Mean Relative RMSE & 1.179884$\times$10$^{-3}$\\
     \bottomrule
    \end{tabular}
    \caption{Single-step errors for the hydrogen-air kinetics problem over a validation dataset with 60000 samples.}
    \label{tab:water-emu-single-step}
\end{table}

% ===================================================
\subsection{Amortized Inference}\label{sec:inference}
% ===================================================

In this section, we consider the problem of inferring reaction rate parameters and observation times from measurements of chemical species concentrations in an amortized fashion.
To do so, we have summarized the results from model inversion in this section using multiple plot styles. We begin by choosing a $\bm{y}^*$ from the validation dataset. 
\emph{Trajectory reconstruction} plots are used to verify that we can recover the state $\bm{y}^*$ if we simulate the system with the predicted inputs $\widehat{\bm{v}}$. 
We also use a \emph{parallel chart} to provide a compact representation of the manifold $\mathcal{M}_{\bm{y}^*}$ of non-identifiable parameters, i.e., it shows the variability in different components of the input $\widehat{\bm{v}}$ leading to an identical outcome $\bm{y}^*$. It also shows a complex multi-dimensional manifold in a single plot, and can be used to determine the solution with minimal stiffness for given concentrations. When plotted on an error-based colored scale, it helps you to determine near-identifiability with respect to identifiability.
Additionally, \emph{two-way correlation} plots show the dependence between component pairs from samples in $\mathcal{M}_{\bm{y}^*}$. 
We have also employed a predictor-corrector sampling method (\emph{PC sampling}~\cite{tong2024invaert}, Algorithm~\ref{alg:pc-sampling}) to reduce the number of outliers during inference.
Finally, we perform local identifiability analysis using the \emph{Fisher information matrix (FIM)} and \emph{direct system solution (DSS)} to validate the inverse solutions computed by the inVAErt network and to confirm our intuition.

FIM analysis is conducted using the same $\bm{y}^*$ we selected above for characterizing the accuracy of the inverse problem solution. 
Corresponding to each such $\bm{y}^*$, multiple $\widehat{\bm{v}}$ are generated through the inVAErt network decoder. For each inverse solution, we compute the sensitivity matrix $\bm{J}=\partial\bm{y}/\partial\widehat{\bm{v}}$ using central differences, i.e., observing the relative change in $\bm{y}$ produced by separate perturbations in each component of $\widehat{\bm{v}}$.
The Fisher information matrix $\bm{F}=\bm{J}^T\,\bm{\Sigma}\,\bm{J}$ (with scaling factors $\bm{\Sigma} = \bm{I}_n$) is then assembled, and its eigenvalues plotted in decreasing order, while radar plots are used to visualize the magnitude of each eigenvector. 
For parameters varying considerably in magnitude, we compute the log-log sensitivities with respect to the reaction rate parameters for the $j$-th species $k_j$, $A_j$ or time $t$.

Additionally, DSS identifies inputs $\bm{v}$ in the training dataset whose outputs are sufficiently close to $\bm{y}^*$, and compares such inputs to $\widehat{\bm{v}}$.
To do so, we first identify a tolerance $\varepsilon$. For each parameter combination in the training dataset (sampled from a uniform prior), we then solve the system of ODEs. We then flag a particular parameter combination as a ``solution'' if it belongs to a $\varepsilon$-neighborhood of $\bm{y}^*$, where the distance between trajectories is measured by a relative $L^2$ error expressed as
\begin{equation}
\left.\sqrt{\sum_{i=1}^{N_{\bm{y}}}(y_i - \hat{y}_i)^2} \middle/ \sqrt{\sum_{i=1}^{N_{\bm{y}}}y_i^2}\right. < \varepsilon.
\end{equation}
The results of this analysis are visualized using parallel plots, where ``solutions'' are identified by blue lines and the remaining parameter combinations are in pink.
For a system that is locally identifiable, we would expect to see fewer blue trajectories by decreasing $\varepsilon$ compared to a system with a higher degree of non-identifiability.
Similar analyses can also be carried out by sampling parameters from arbitrary priors, not necessarily those associated with the training dataset.
We have applied DSS to the POLLU, reversible, and hydrogen-air reaction systems in the result section. 
We finally emphasize the \emph{local character} of these analyses, which depend on the selected $\bm{y}^*$, and reiterate that these approaches are used to verify the results obtained from amortized inference.

% ===============================
\subsubsection{Robertson problem}
% ===============================

\begin{figure}[!ht]
    \centering
    \includegraphics[width=\textwidth]{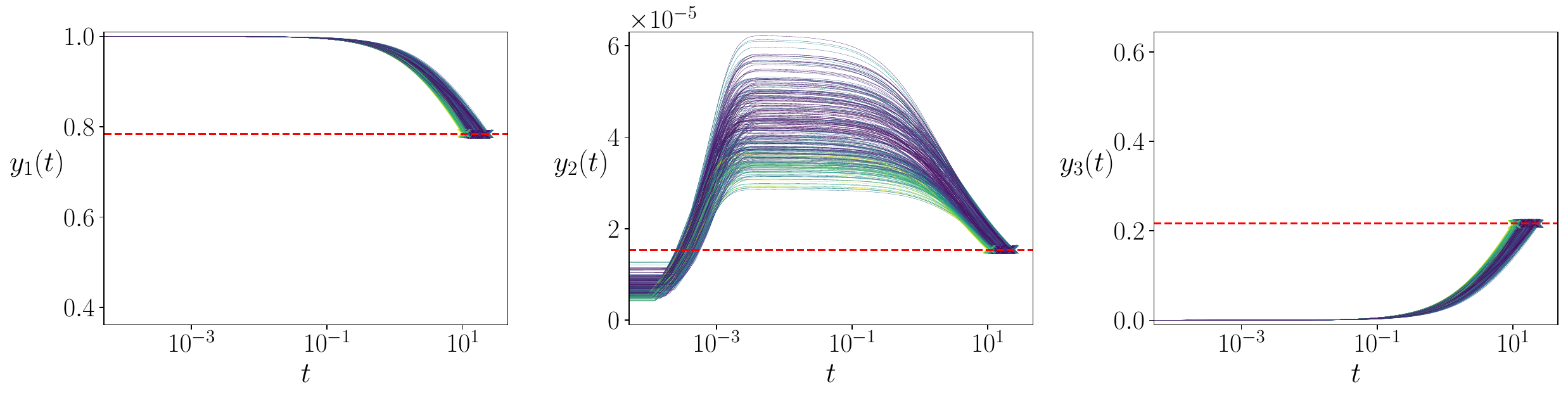}
    \caption{Trajectory reconstructions for the Robertson system at a randomly selected $\bm{y}^*$ from the validation set. The target concentrations $\bm{y}^{*}$ are indicated by a horizontal red line. The trajectories are colored based on the reconstruction RMSE scale shown in Figure~\ref{fig:rober-inv-parallel}}.
    \label{fig:rober-inv-trajectories}
\end{figure}

\begin{figure}[!ht]
    \centering
    \includegraphics[width=\textwidth]{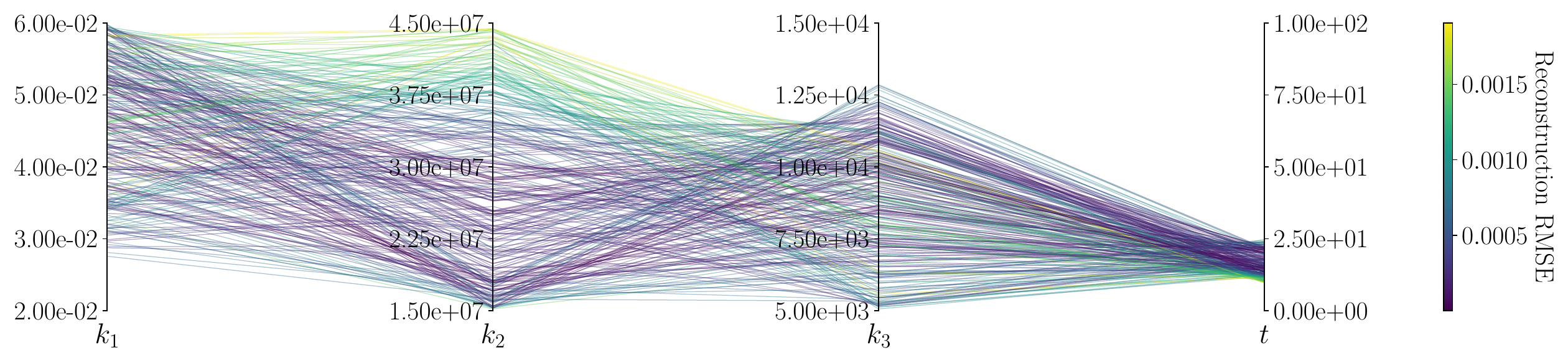}
    \caption{Parallel chart for the Robertson problem parameters $(\bm{k}, t)$ using the same $\bm{y}^*$ as in Figure~\ref{fig:rober-inv-trajectories}, corresponding to the 300 latent space samples. The color of each trajectory quantifies the RMSE error between $\widehat{\bm{y}}$ and $\bm{y}^*$.}
    \label{fig:rober-inv-parallel}
\end{figure}

\begin{figure}[!ht]
    \centering
        \includegraphics[width=\textwidth]{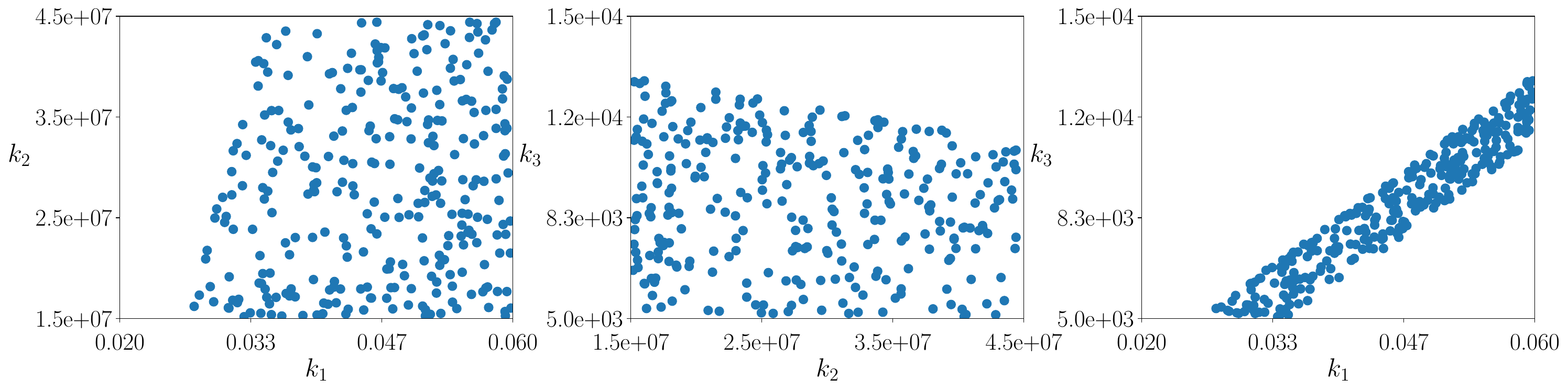}
    \caption{Correlations between the Robertson problem parameters $(\bm{k}, t)$ for the $\bm{y}^*$ used in Figure~\ref{fig:rober-inv-trajectories}, corresponding to the 300 latent space samples.}
    \label{fig:rober-inv-correlations}
\end{figure}

An inVAErt network with a six-dimensional latent space was trained for 5000 epochs, and inference results were generated using two rounds of \emph{PC sampling} (see Algorithm~\ref{alg:pc-sampling}).
Results from the inversion task are computed for the concentrations $\bm{y}^{*}=[$7.83406533$\times$10$^{-1}$, 1.53767928$\times$10$^{-5}$, 2.16578090$\times$10$^{-1}]$.
The inversion task reveals a few interesting patterns. 
% Interesting aspects
% single point in time at which the prescribed concentration are met
Due to the monotonic character of at least one of the three concentrations, there is a single point in time where $\bm{y}^{*}$ is matched, as shown in Figure~\ref{fig:rober-inv-trajectories}.
Additionally, Figure~\ref{fig:rober-inv-parallel} and~\ref{fig:rober-inv-correlations} highlight other two interesting features.
% Reparameterization
To better discuss these features, we rewrite the system using $y_1=1-y_2-y_3$, leading to
\[
\begin{split}
\dot{y}_{2} &= k_{1}(1-y_2-y_3)-k_2\,y^{2}_{2}-k_{3}\,y_2\,y_3\\
\dot{y}_{3} &= k_2\,y^{2}_{2}.
\end{split}
\]
We now introduce a reparameterization in time of the form $\tau=k_{2}\,t$, and define the coefficients $(\varepsilon_{1},\varepsilon_{2})=(k_{1}/k_{2},k_{3}/k_{2})$, leading to
\begin{equation}\label{equ:robert_2dims}
\begin{split}
\dot{y}_{2} &= \varepsilon_{1}(1-y_2-y_3)-y^{2}_{2}-\varepsilon_{2}\,y_2\,y_3\\
\dot{y}_{3} &= y^{2}_{2},
\end{split}
\end{equation}
where the time derivatives are taken with respect to $\tau$. Up to a reparameterization of time, system~\eqref{equ:rober} is equivalent to \eqref{equ:robert_2dims}, see, e.g.,~\cite{baumgartner2025multiparameter}. 
The first feature therefore reflects the fact that $k_2$ can be chosen freely and independently of $k_1$ and $k_3$, provided that $k_{1},k_{3}\ll k_{2}$, as correctly captured by the inVAErt network in Figure~\ref{fig:rober-inv-correlations}.
% linear correlation between k1 and k3
Additionally, this justifies the linear correlation between $k_1$ and $k_3$. For a fixed value of the concentration and their derivatives, the first equation in \eqref{equ:robert_2dims} provides a linear relation between $\varepsilon_{1}$ and $\varepsilon_{2}$.
% limited range of k1
As a second feature, $\bm{y}^{*}$ can be obtained only for a limited range of the $k_{1}$ and $k_{3}$ coefficients. Also, it is evident from \eqref{equ:robert_2dims} that $\varepsilon_{1}=k_{1}/k_{2}$ is the only variable associated with first order terms, hence has a major impact on the dynamics of the entire system.

% ===========================
\subsubsection{POLLU problem}
% ===========================

\begin{figure}[!ht]
\centering
\includegraphics[width=0.8\textwidth]{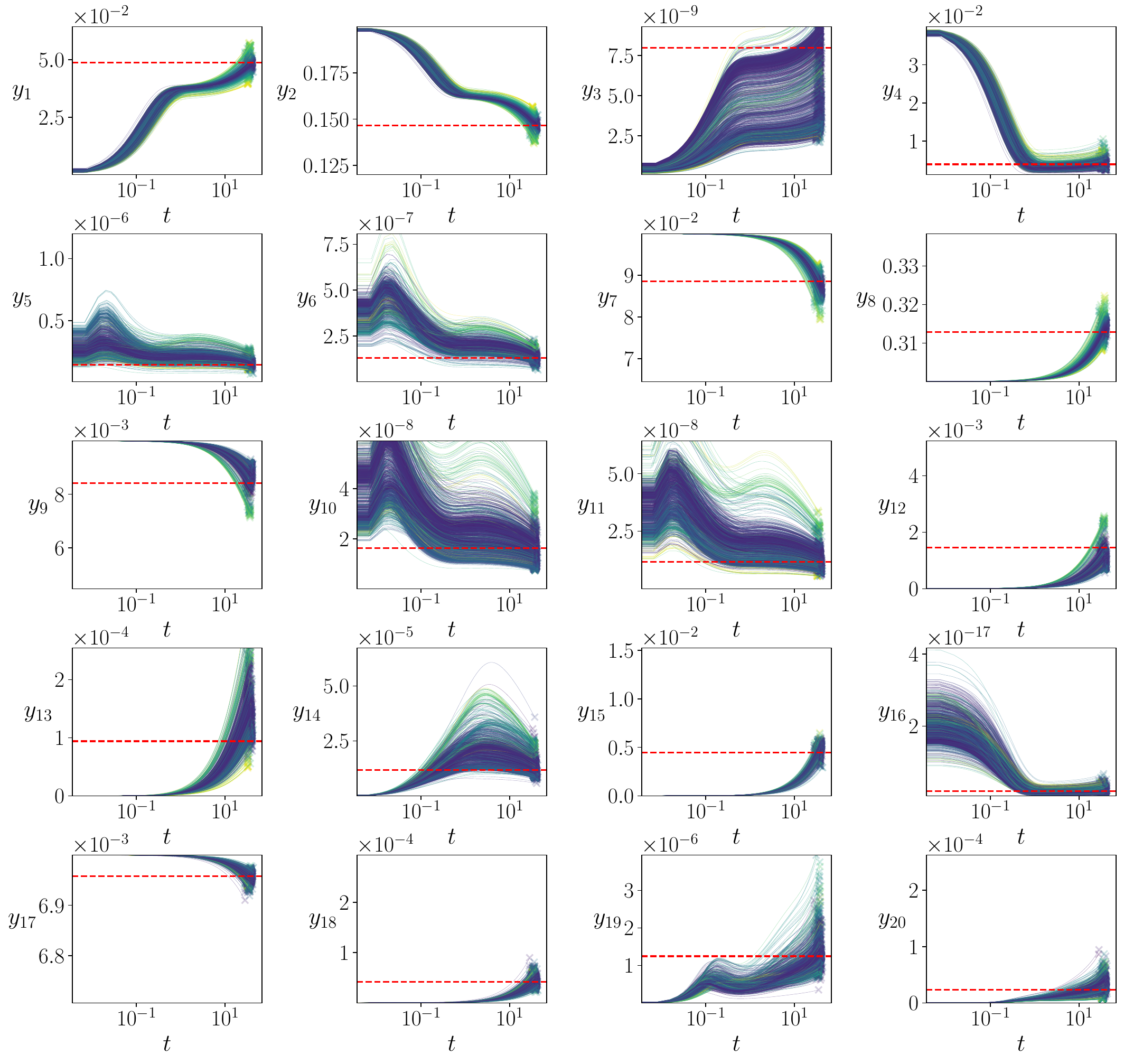}
\caption{POLLU system trajectories inverted from a randomly chosen $\bm{y}^*$ from the validation set. The figure shows an ensemble of trajectories generated by 2000 latent space samples. The trajectories are colored based on the reconstruction RMSE scale shown in Figure~\ref{fig:pollu-inv-parallel-2000}}.
\label{fig:pollu-inv-trajectories-2000}
\end{figure}

\begin{figure}[!ht]
    \centering
    \includegraphics[width=\linewidth]{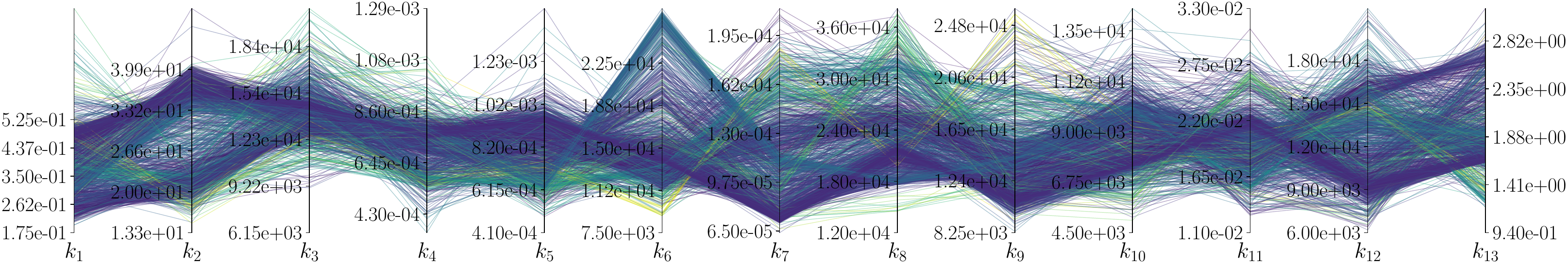}
    \includegraphics[width=\linewidth]{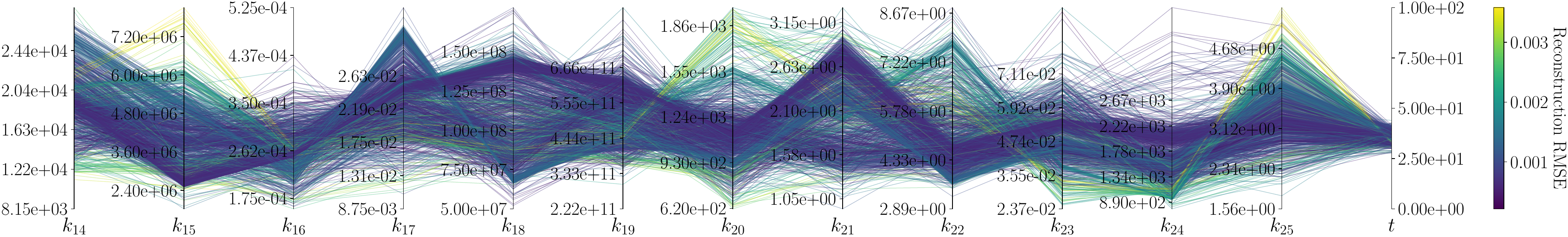}
    \caption{Parallel charts for the POLLU problem parameters $(\bm{k}, t)$ for the same $\bm{y}^*$ used in Figure~\ref{fig:pollu-inv-trajectories-2000}, corresponding to 2000 latent space samples. Parameter trajectories are colored by distance (RMSE) of the corresponding model solution to the targets.}
    \label{fig:pollu-inv-parallel-2000}
\end{figure}

Data-driven inversion of the POLLU system was performed using 
\[
\begin{aligned}
\bm{y}^{*} =[&4.88328162\times10^{-2},\; 1.46545807\times10^{-1},\; 7.97528814\times10^{-9},\; 3.95319461\times10^{-3},\;
1.46661697\times10^{-7},\\ &1.30005375\times10^{-7},\; 8.86035242\times10^{-2},\; 3.12917892\times10^{-1},\;
8.40929169\times10^{-3},\; 1.62704280\times10^{-8},\\
& 1.16691787\times10^{-8},\; 1.46016233\times10^{-3},\;
9.40596559\times10^{-5},\; 1.16793441\times10^{-5},\; 4.47871309\times10^{-3},\\ 
&1.88012741\times10^{-18},\;
6.95776888\times10^{-3},\; 4.22311184\times10^{-5},\; 1.24442820\times10^{-6},\; 2.36797927\times10^{-5}],
\end{aligned}
\]
with a latent space of dimension 8 and 8000 epochs of training, leading to a final inversion accuracy of 99.98\%.
Inference results obtained using 2 rounds of PC sampling are shown in Figure~\ref{fig:pollu-inv-trajectories-2000}, and~\ref{fig:pollu-inv-parallel-2000}.
Given the larger number of variables (20 species) involved, POLLU represents a significantly more difficult inversion problem with respect to the Robertson system. 
However, Figure~\ref{fig:pollu-inv-trajectories-2000} shows that concentration trajectories associated with parameter combination inferred by an inVAErt network, led to concentrations close to $\bm{y}^*$, particularly when such concentrations are higher than $10^{-3}$.

% there is no specific pattern for the points in the manifold. 
% the time is unique
As in the Robertson problem, a monotonic variation in the concentration of at least some species over time should imply a unique time point associated with the observation $\bm{y}^*$.
% similar locations and larger errors in the smaller concentrations
However, due to the the larger number of species, many of which have negligibly small concentrations, relative errors in Figure~\ref{fig:pollu-inv-trajectories-2000} are generally higher for the POLLU system.
% parameter traces are plotted by error. There seems to be identifiability
Figure~\ref{fig:pollu-inv-parallel-2000} shows parameter combinations sampled from the manifold $\mathcal{M}_{\bm{y}^{*}}$ and colored by the distance of the corresponding model output from the target concentrations. 

\begin{figure}[!ht]
\centering
    \includegraphics[width=0.45\textwidth]{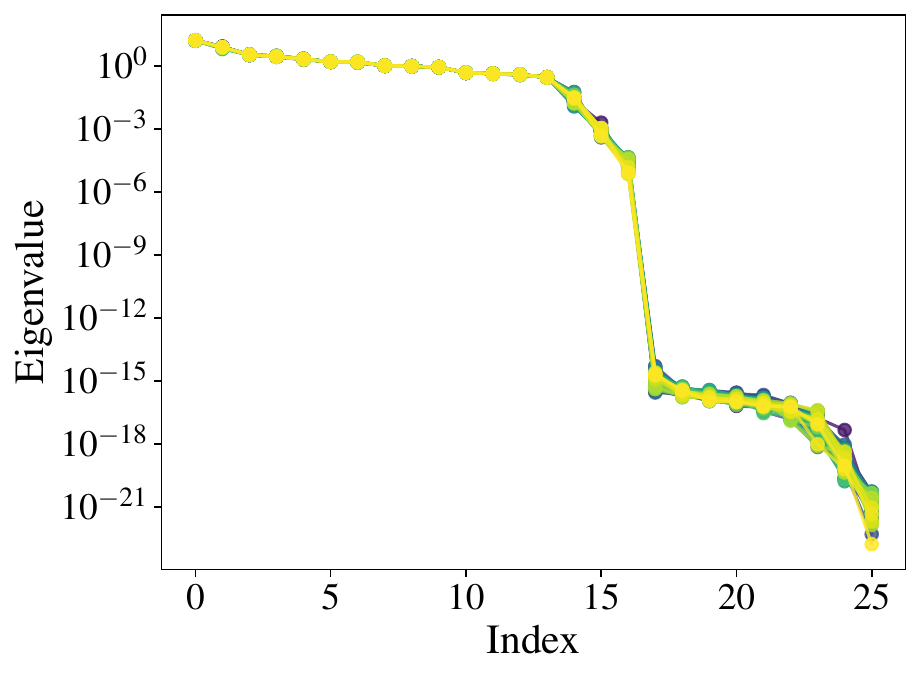}
    \caption{FIM eigenvalue decay for the POLLU system. Different colors are based on $\hat{t}$ associated with 50 $\hat{\bm{v}}$ inferred from $\bm{y}^*$ as shown in Figure~\ref{fig:pollu-inv-trajectories-2000}.}
    \label{fig:pollu-eig}
\end{figure}

\begin{figure}[!ht]
\centering
\begin{subfigure}{0.30\textwidth}
    \centering
    \includegraphics[width=\textwidth]{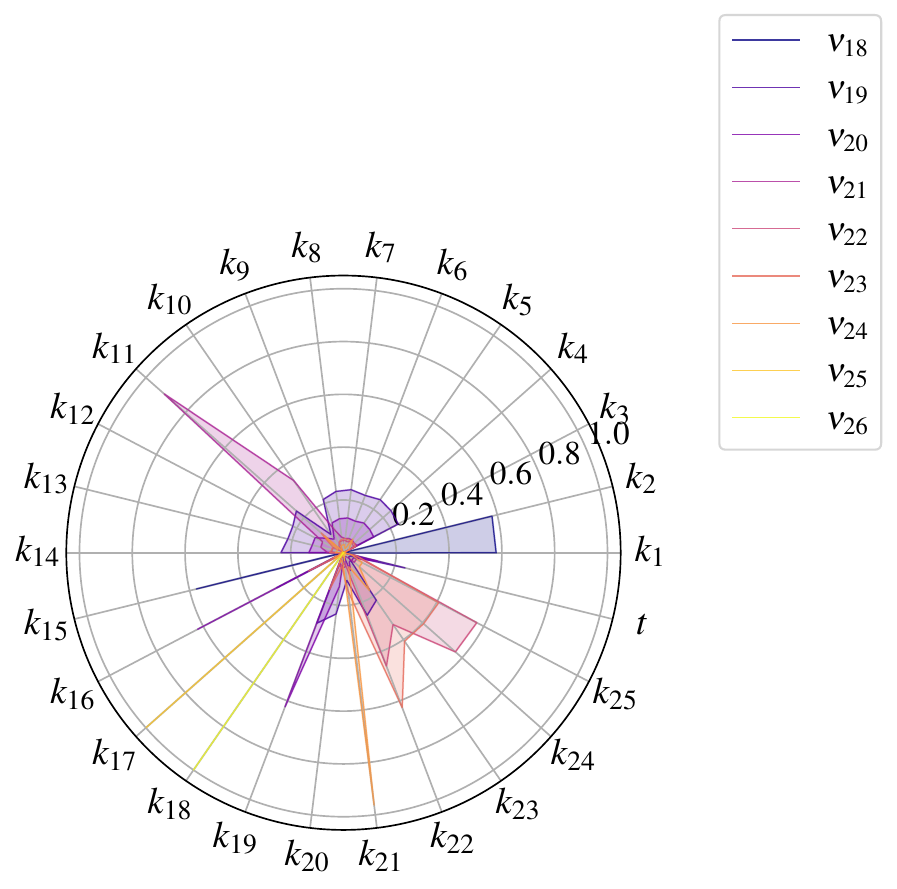}
    \caption{$\hat{t}=32.065370877596294$}
\end{subfigure}
\begin{subfigure}{0.30\textwidth}
    \centering
    \includegraphics[width=\textwidth]{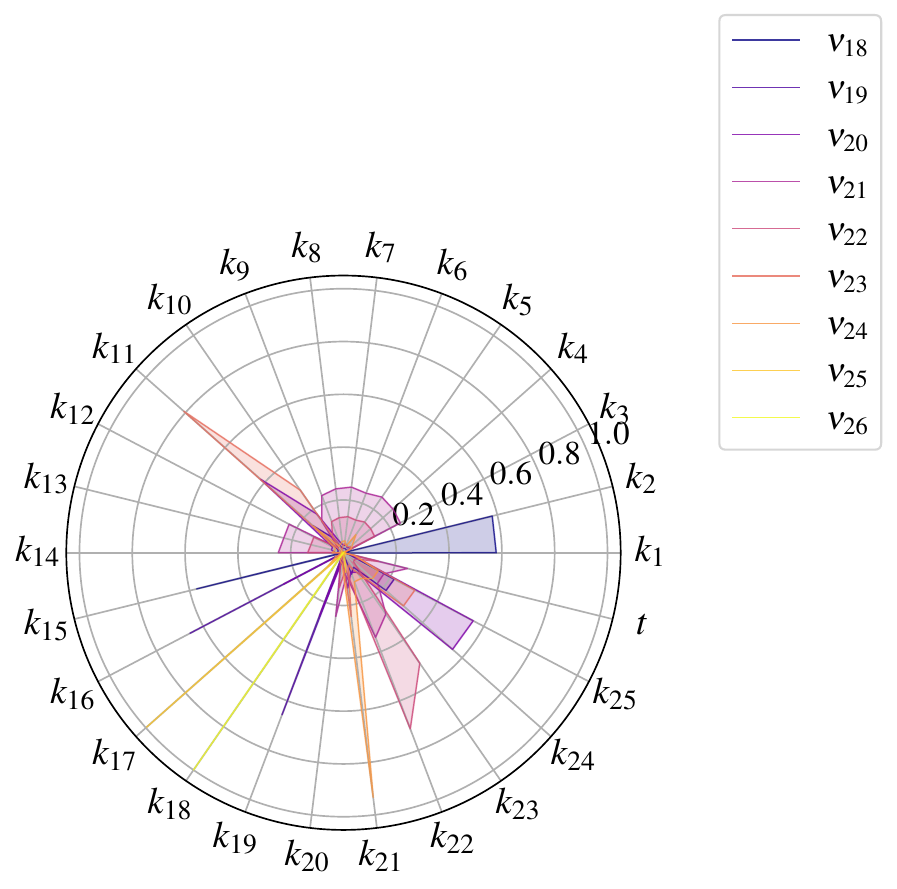}
    \caption{$\hat{t}=33.67715515275458$}
\end{subfigure}
\begin{subfigure}{0.30\textwidth}
    \centering
    \includegraphics[width=\textwidth]{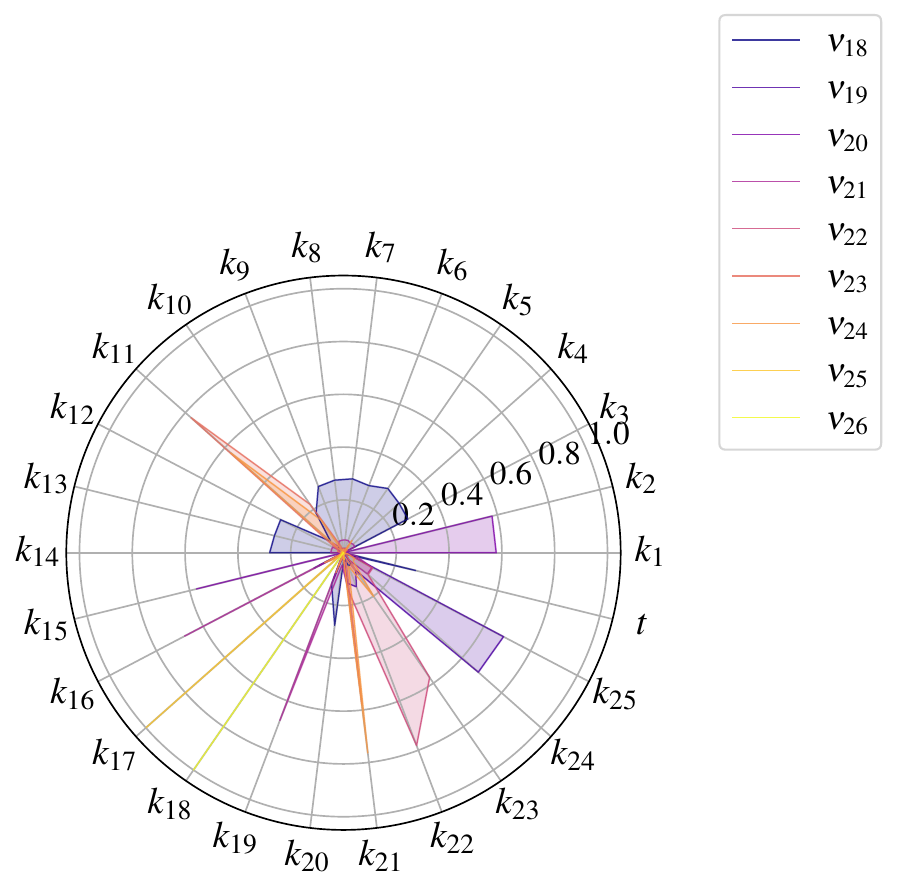}
    \caption{$\hat{t}=34.154995983925126$}
\end{subfigure}
\caption{Radar plots of FIM  singular eigenvectors for different $\hat{t}$ for the POLLU system, where a brighter color corresponds to smaller eigenvalues.}
\label{fig:pollu-radar}
\end{figure}

In the computation of the FIM, we use $1\%$ perturbations of $\widehat{\bm{v}}$ when computing the finite difference approximation of the derivatives. 
Additionally, as the values of $\bm{k}$ and $\bm{y}$ have different order of magnitude, $\partial \bm{y}/\partial\widehat{\bm{v}}$ is replaced by the log-log sensitivities. % for the $k_j$, and log-linear sensitivities with respect to $t$. 
The sensitivity matrix is therefore computed as
\[
J_{[i, j]} =
\begin{cases}
\displaystyle \frac{\partial \log y_i}{\partial \log k_j} = \frac{k_j}{y_i} \frac{\partial y_i}{\partial k_j} , & j < 26 \\
\\
\displaystyle \frac{\partial \log y_i}{\partial \log t} = \frac{t}{y_i} \frac{\partial y_i}{\partial t} , & j = 26.
\end{cases}
\]
The FIM eigenvalue decay in Figure~\ref{fig:pollu-eig} indicates the presence of approximately 17 non-zero eigenvalues. 
The radar plots in Figure~\ref{fig:pollu-radar} illustrate that singular eigenvectors associated with the 9 smallest eigenvalues are consistent across inverse solutions at different times, and emphasize a group of unimportant rate constants $\{k_{15}, k_{16}, k_{17}, k_{18}, k_{19}, k_{21}\}$ of which $\{k_{16}, k_{17}, k_{21}\}$ are consistently associated with the smallest eigenvalues. 
In order to verify this, we have fixed all the important rate constants at their nominal values and randomly sampled those visualized in the radar plots, showing that there is not much variability among the trajectories resulting from 100 such simulations. This is further discussed in Section~\ref{sec:appendix-plots} in the appendix.


% ===============================================================
\subsubsection{Systems with Reversible and Irreversible Kinetics}
% ===============================================================

The inverse problem solutions for the reversible and irreversible reaction systems are analyzed next.
% reversible system
Target concentrations are selected as $\bm{y}^{*} = [0.12153625, 0.09434589, 0.78411786]^{T}$ and $\bm{y}^{*} = [0.40551412, 0.22666444, 0.36782144
]^{T}$ for the reversible and irreversible systems, using latent spaces of dimensions 8 and 6, respectively.
Note that, unlike previous systems, initial conditions are included as inputs for both these cases.

Figure~\ref{fig:rev-inv-parallel} illustrate a parallel plot with two distinct families of trajectories, the first originating from $t=0$, and the second covering a wider spectrum of time values, corresponding to separate solution manifolds containing various combinations of initial conditions and system parameters, respectively. 
For initial conditions, the easiest way to reach the target concentrations is to start with $\bm{y}(0)=\bm{y}^{*}$. This is independent of the choice of the parameters, as confirmed by the spread in values of $\bm{k}$ for the trajectories starting at $t=0$.

Figure~\ref{fig:rev-inverse-correlations} also confirms these two regimes, with nearly identical initial concentrations (or zero time) corresponding to multiple system parameters, and the sum of the concentrations always smaller than $1$.

\begin{figure}[!ht]
\centering
\includegraphics[width=0.5\textwidth]{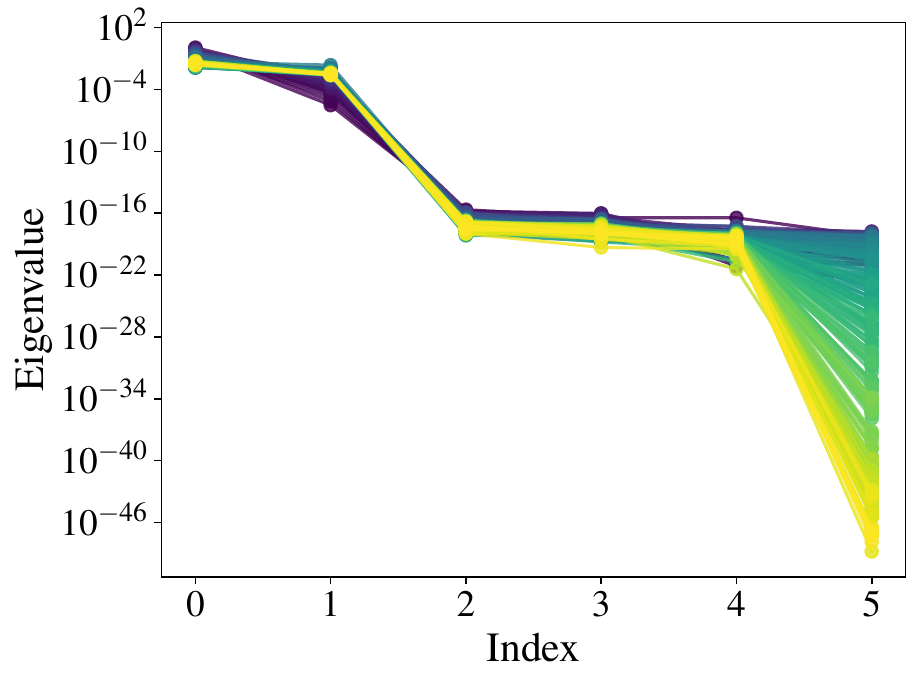}
\caption{FIM eigenvalue decay for the reversible system, corresponding to 300 $\widehat{\bm{v}}$, arranged in increasing order of $\hat{t}$, where a lighter color indicates an increasing time $\hat{t}$.}
\label{fig:rev-eig}
\end{figure}

\begin{figure}[!ht]
\centering
\begin{subfigure}{0.30\textwidth}
    \centering
    \includegraphics[width=\textwidth]{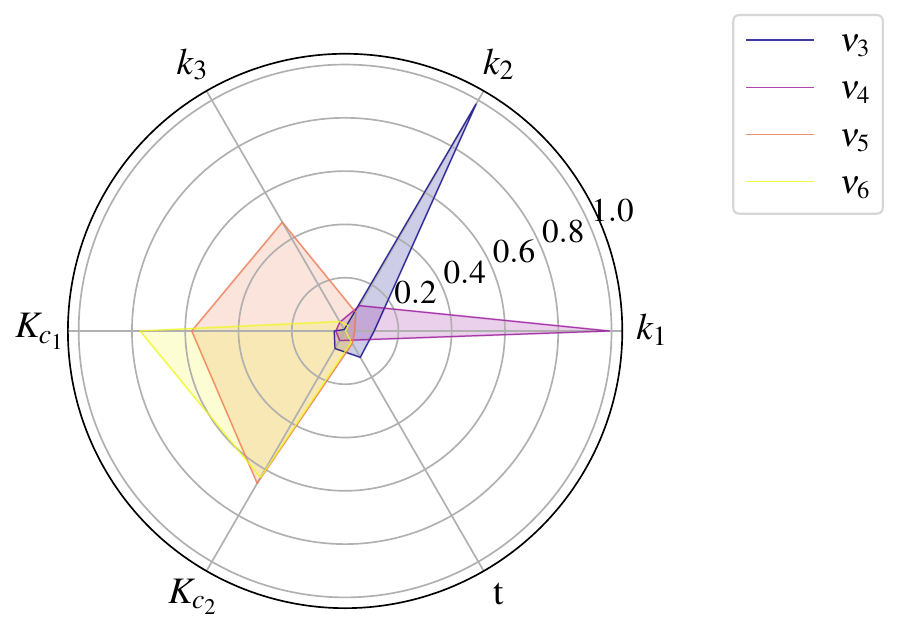}
    \caption{$\hat{t}=0.19212855365818515$}
\end{subfigure}
\begin{subfigure}{0.30\textwidth}
    \centering
    \includegraphics[width=\textwidth]{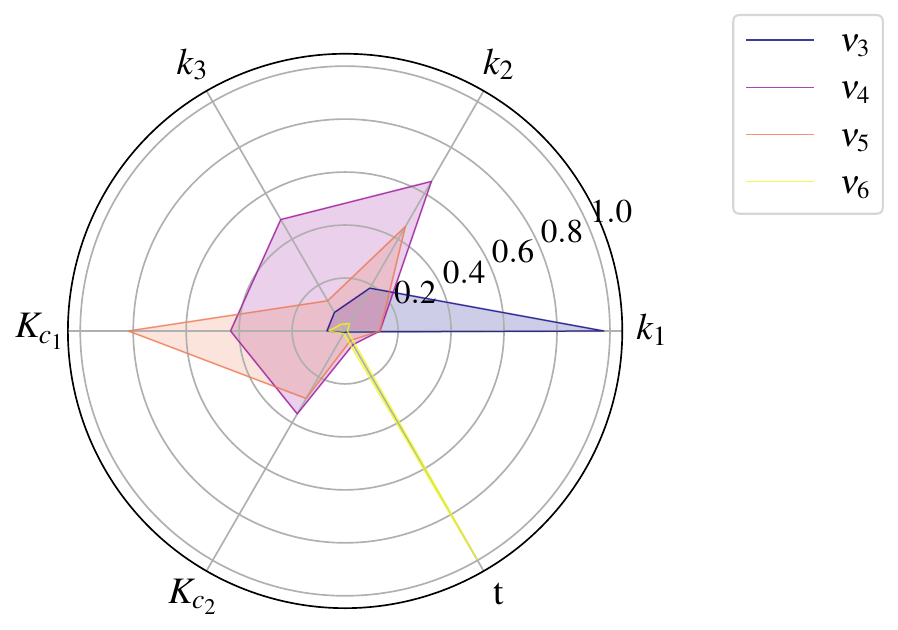}
    \caption{$\hat{t}=0.7850543937908111$}
\end{subfigure}
\begin{subfigure}{0.30\textwidth}
    \centering
    \includegraphics[width=\textwidth]{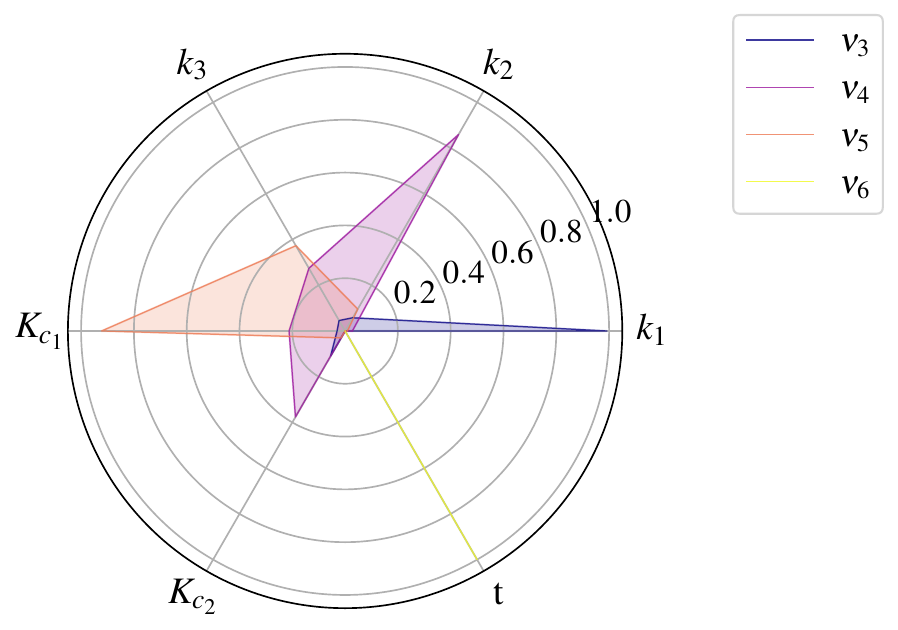}
    \caption{$\hat{t}=4.818707634268933$}
\end{subfigure}
\caption{Radar plots for the singular eigenvectors at different $\hat{t}$ for the reversible system. Colors (from dark to bright) reflect a decreasing magnitude of the respective eigenvalues.}\label{fig:rev_eigenvectors}
\end{figure}

We then analyze the FIM and DSS with inputs $\bm{v}=(\bm{k}, \bm{y}(0), t)$. However, inclusion of the initial conditions $\bm{y}(0)$ in sensitivity analysis, it reveals the obvious correlation between $\bm{y}(0)$ and $\bm{y}(t)$'s where $t\approx0$ and $\bm{y}(0)\approx \bm{y}(t)$. 
To make things simpler, we have fixed $\bm{y}(0)=[0.10982296, 0.0624823, 0.82769474]^T$ (true values from the training dataset for $\bm{y}^*$) and analyzed sensitivities only with relation to $(\bm{k}, t)$.
The input perturbation $\Delta \widehat{\bm{v}}$ for central difference was selected as $1\%$ of $\hat{\bm{v}}$.
It is clear from Figure~\ref{fig:rev-eig} that across all variations in $\hat{t}$, there are 2 non zero eigenvalues, corresponding to the two system equations. 
The radar plots in Figure~\ref{fig:rev_eigenvectors} show eigenvectors in decreasing order of magnitude for the corresponding null eigenvalue. Note how the consistent lack of importance of $k_{1}$ over time is also captured by its large variations in the parallel plot in Figure~\ref{fig:rev-inv-parallel}.

Figure~\ref{fig:rev-inv-parallel} shows evidence of some predictions beyond the prior range from which the training data was obtained. These extrapolated parameter combinations correspond to the values in the parallel chart that go beyond the labeled portions on the axes for $k_2$ and $k_3$. In order to verify the extrapolation capabilities of our model, we performed DSS over 10000 parameter combinations randomly sampled from a larger prior ($k_2\in [1, 4]$ instead of $[1,3]$, and $k_3\in [0.5,2]$ instead of $[0.5, 1.5]$). Figure~\ref{fig:rev-sys-sol} shows results from this test, where we observe that it indeed matches the pattern observed in Figure~\ref{fig:rev-inv-parallel} from model inversion.

\begin{figure}[!ht]
    \centering
        \includegraphics[width=\textwidth]{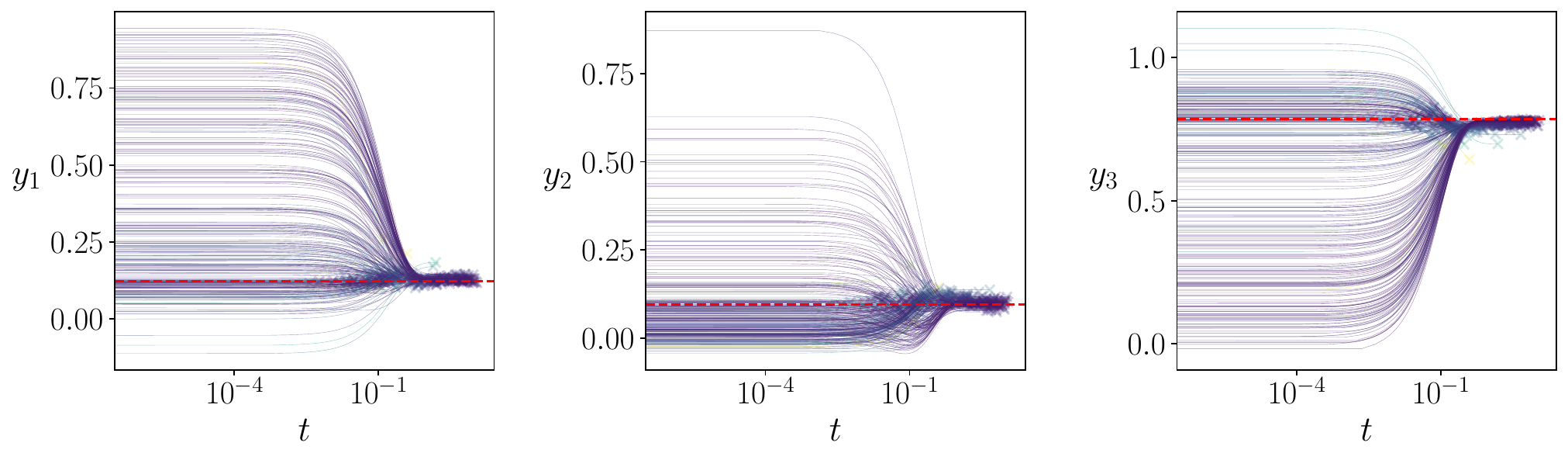}
        \caption{Trajectory reconstructions for the reversible system from 300 latent space samples. We used a randomly chosen $\bm{y}^*$ from the validation set.}
    \label{fig:rev-inv-trajectories}
\end{figure}

\begin{figure}[!ht]
    \centering
        \includegraphics[width=\textwidth]{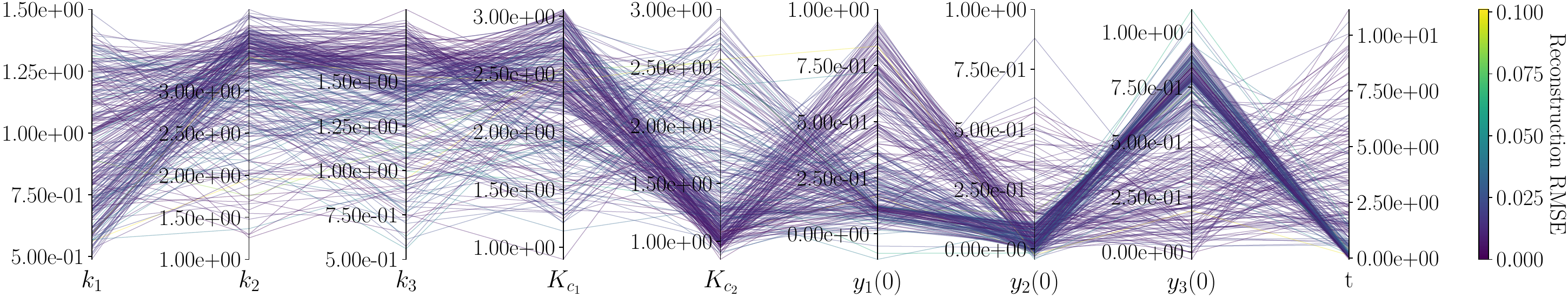}
        \caption{Parallel chart for the reversible system parameters $(\bm{k}, t)$ for the same $\bm{y}^*$ used in Figure~\ref{fig:rev-inv-trajectories}.}
    \label{fig:rev-inv-parallel}
\end{figure}

\begin{figure}[H]
    \centering
        \includegraphics[width=\textwidth]{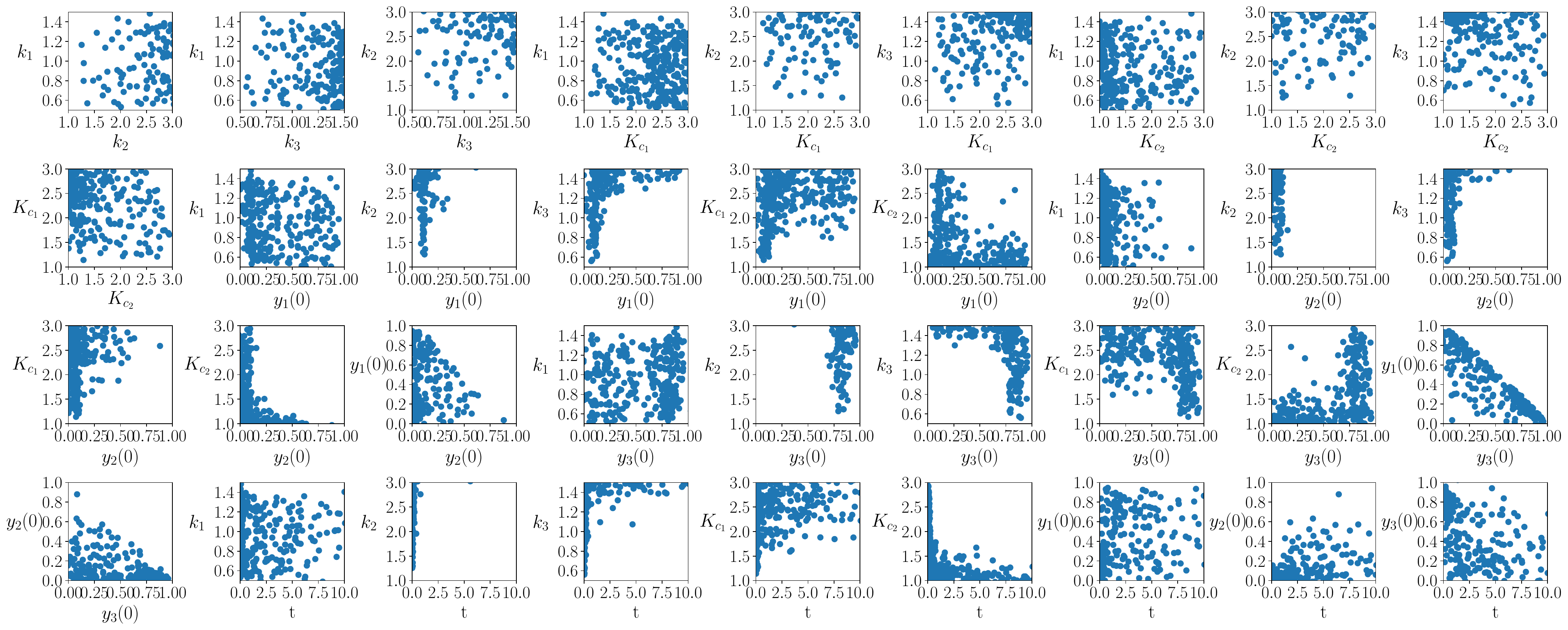}
    \caption{Correlations between the reversible system parameters for the $\bm{y}^*$ used in Figure~\ref{fig:rev-inv-trajectories}.}
    \label{fig:rev-inverse-correlations}
\end{figure}

\begin{figure}[!ht]
    \centering
    \includegraphics[width=0.9\linewidth]{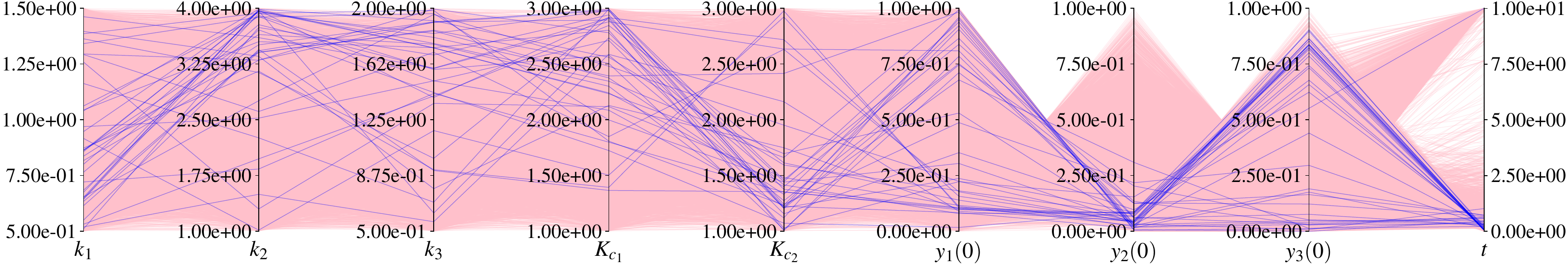}
    \caption{DSS for the reversible system with larger priors ($k_2\in [1, 4]$ instead of $[1,3]$, and $k_3\in [0.5,2]$ instead of $[0.5, 1.5]$) on resampled parameter values for $\varepsilon=5\times10^{-3}$, using the same $\bm{y}^*$ as in Figure~\ref{fig:rev-inv-parallel}.}
    \label{fig:rev-sys-sol}
\end{figure}

% Irreversible system
The manifold of non-identifiable parameters $\mathcal{M}_{\bm{y}^{*}}$ for the irreversible system also appears as the synthesis of two sets of trajectories.
The first set concentrated around $t=0$ is related to initial conditions that coincide with the target concentrations.
Similar to the system with reversible kinetics, the second set corresponds to a one dimensional manifold of non-identifiable parameters which can be seen from the correlations in Figure~\ref{fig:irr-inverse-correlations}, achievable for a wide range of initial conditions, as apparent in Figure~\ref{fig:irr-inv-parallel}. 
This can be seen from~\eqref{equ:irr_eq}, where fixed target concentrations with known derivatives results in a system of two equations and three unknowns.

\begin{figure}[!ht]
    \centering
        \includegraphics[width=\textwidth]{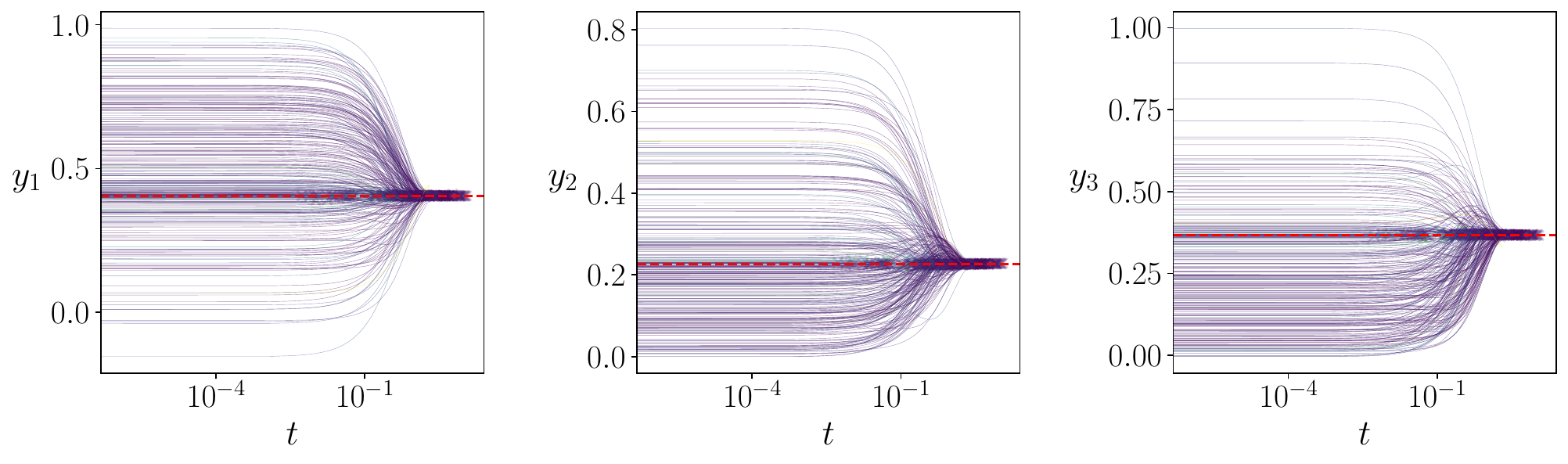}
        \caption{Irreversible system trajectories reconstructed using a randomly chosen $\bm{y}^*$  from the validation set. The figure shows the solutions corresponding to 300 latent space samples.}
    \label{fig:irr-inv-trajectories}
\end{figure}

\begin{figure}[!ht]
    \centering
        \includegraphics[width=\textwidth]{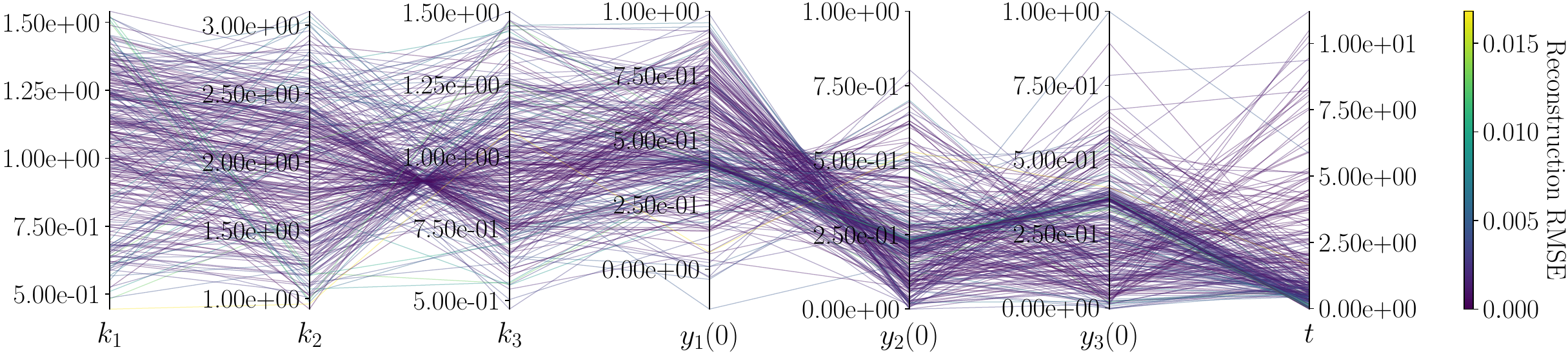}
        \caption{Parallel chart showing the irreversible system parameters $(\bm{k}, t)$ obtained from 300 latent space samples and the same $\bm{y}^*$ used in Figure~\ref{fig:irr-inv-trajectories}.}
    \label{fig:irr-inv-parallel}
\end{figure}

\begin{figure}[!ht]
    \centering
        \includegraphics[width=\textwidth]{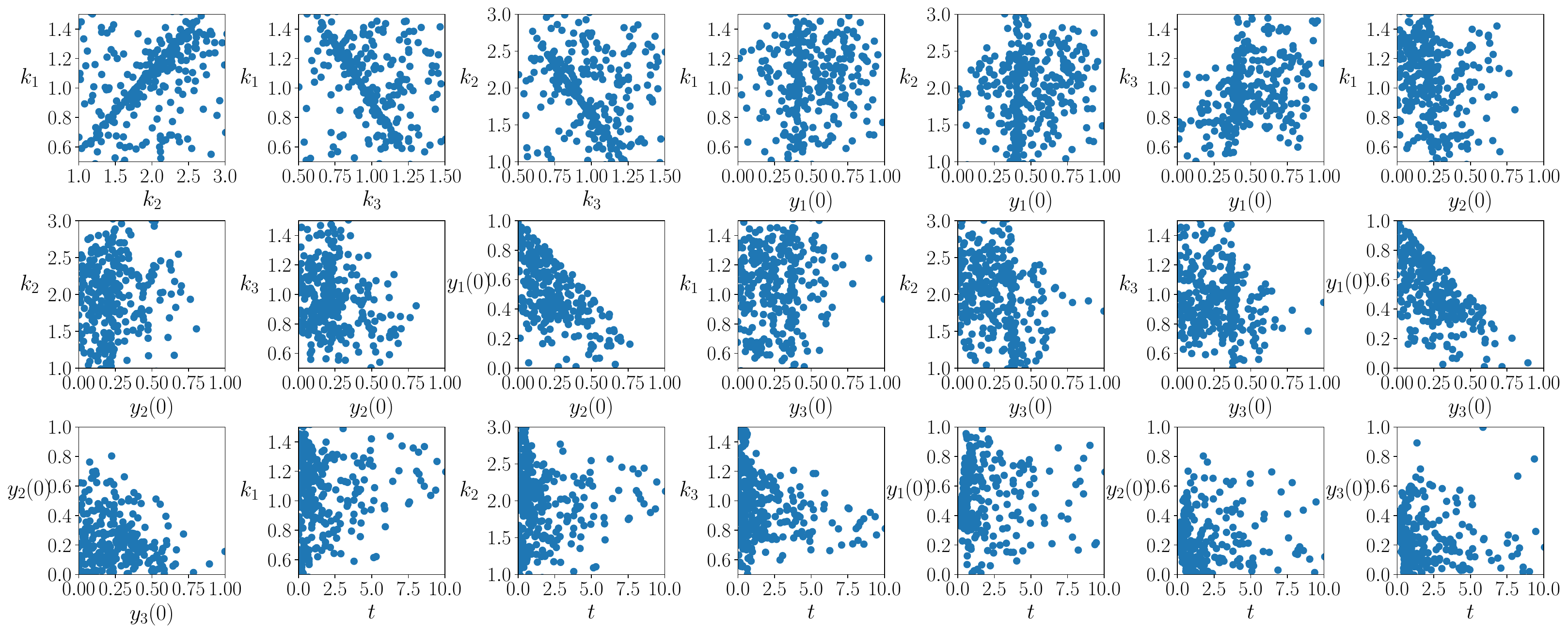}
    \caption{Correlations between parameters of the irreversible system determined by inversion from the same $\bm{y}^*$ used in Figure~\ref{fig:irr-inv-trajectories}.}
    \label{fig:irr-inverse-correlations}
\end{figure}

% ===========================================
\subsubsection{Hydrogen-air kinetics problem}
% ===========================================
  
\begin{figure}[!ht]
    \centering
        \includegraphics[width=0.85\textwidth]{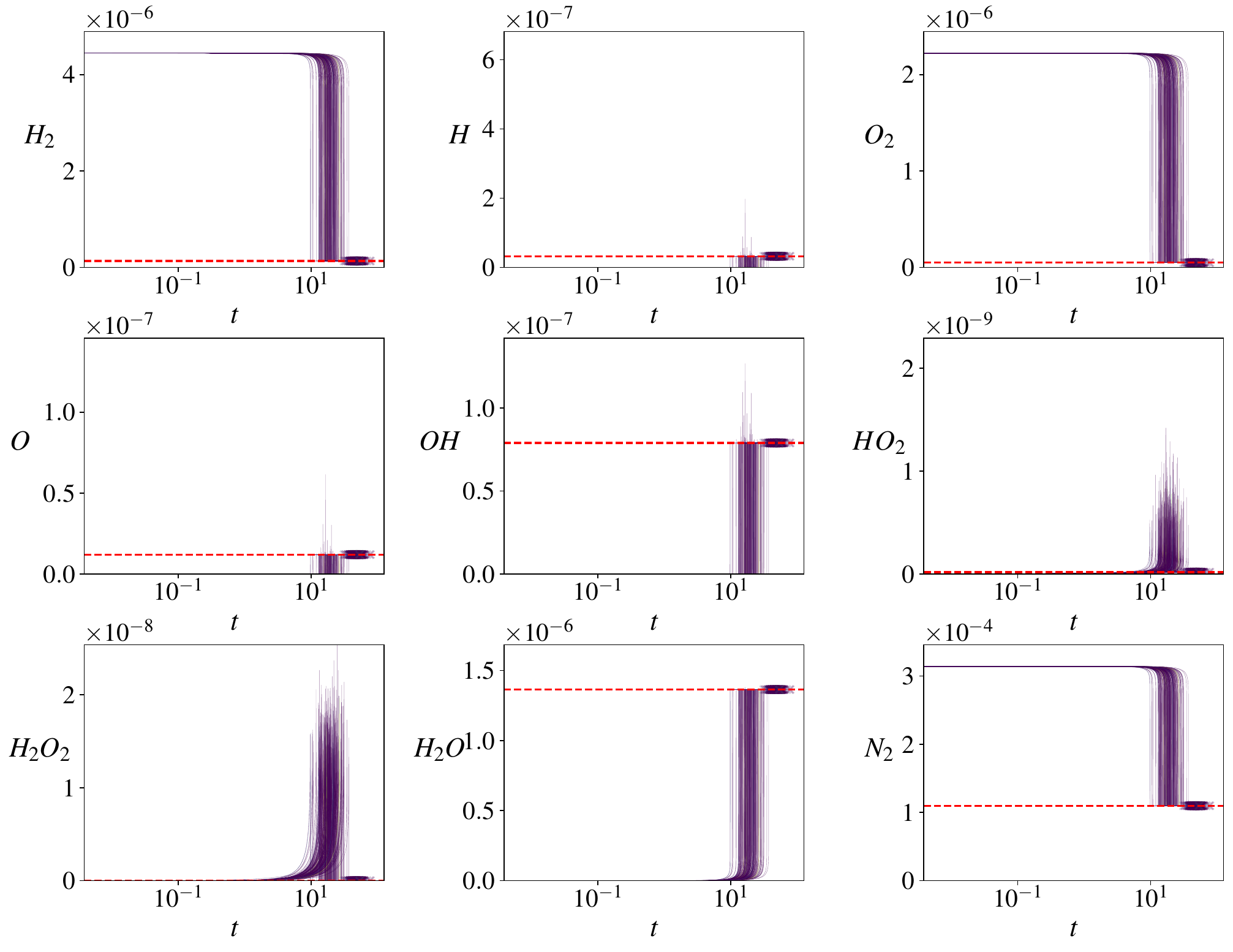}
    \caption{Hydrogen-air system solutions corresponding to 200 latent space samples, and a randomly chosen $\bm{y}^*$ (dashed red line) from the validation dataset.}
    \label{fig:water-inv-trajectories}
\end{figure}

\begin{figure}[!ht]
    \centering
    \includegraphics[width=\textwidth]{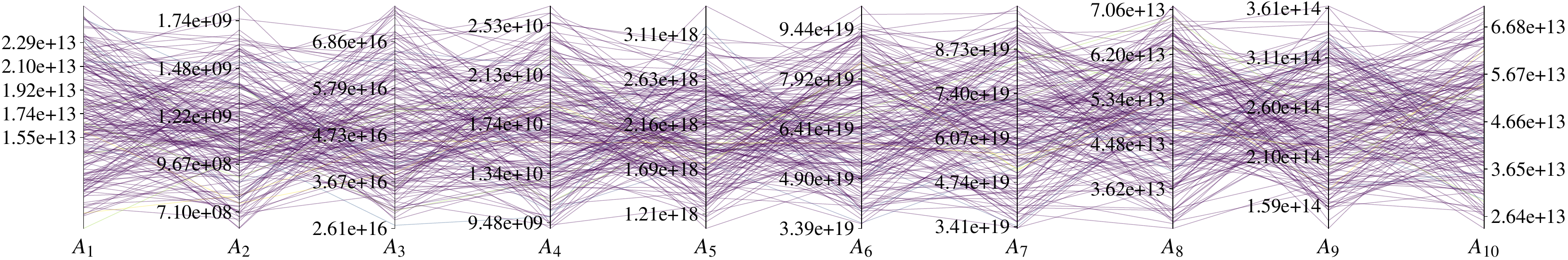}
    \vspace{1cm}
    \includegraphics[width=\textwidth]{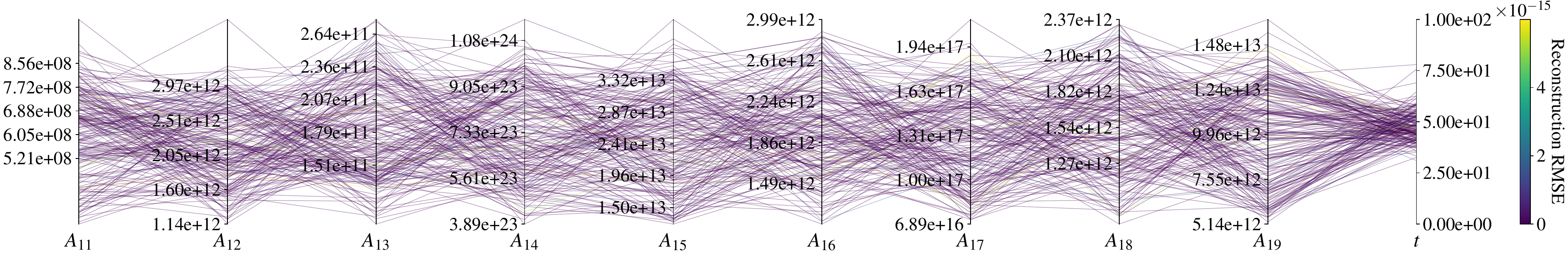}
    \caption{Parallel charts for the hydrogen-air kinetics problem parameters $(\bm{A}, t)$ obtained from 200 latent space samples, and the same $\bm{y}^*$ used in Figure~\ref{fig:water-inv-trajectories}. Note that the overall order of reconstruction errors is $O(10^{-15})$. This is on account of the system approaching a steady state exactly at $\bm{y}^*$.}
    \label{fig:water-inv-parallel}
\end{figure}

\begin{figure}[!ht]
\centering
\begin{subfigure}{0.5\textwidth}
    \centering
    \includegraphics[width=\textwidth]{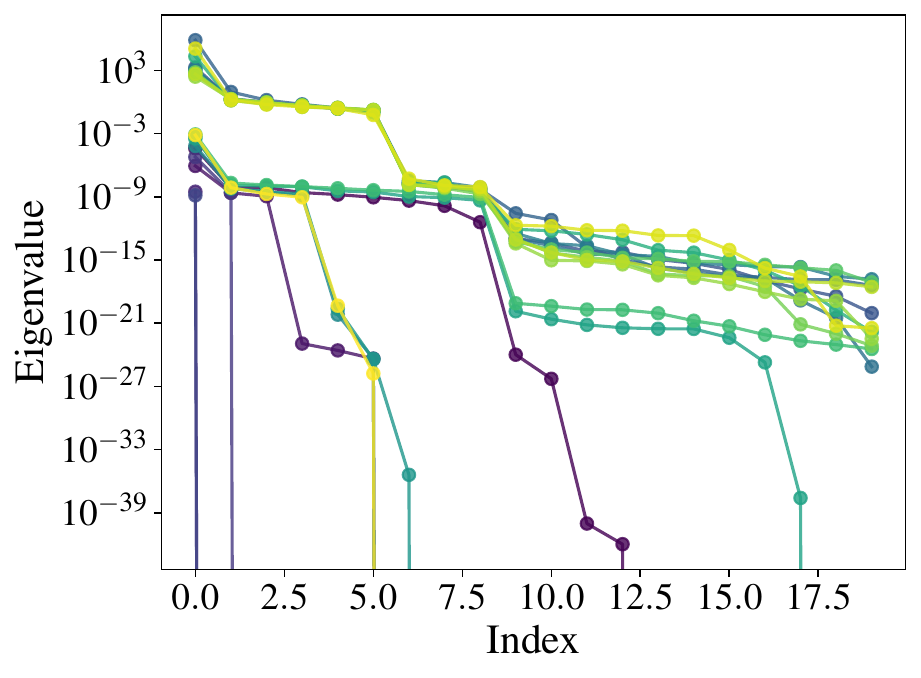}
    \caption{$\bm{y}^+$ (pre ignition)}
\end{subfigure}
\begin{subfigure}{0.49\textwidth}
    \centering
    \includegraphics[width=\textwidth]{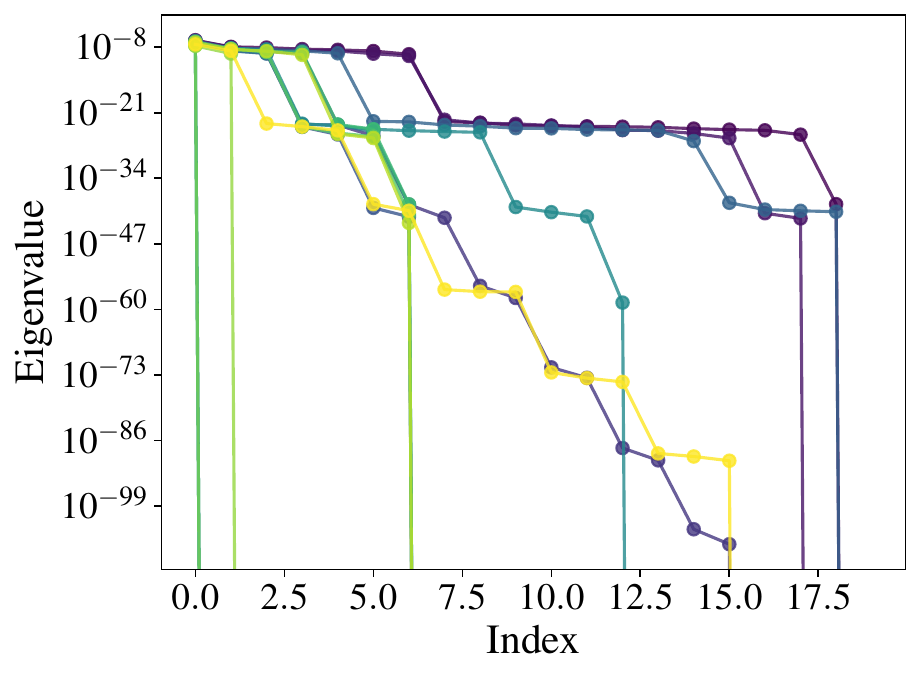}
    \caption{$\bm{y}^*$ (post ignition)}
\end{subfigure}
\caption{FIM eigenvalue decay for the hydrogen-air system corresponding to a \emph{pre-ignition} $\bm{y}^{+}$ and \emph{post-ignition} $\bm{y}^{*}$. Brighter color are used for later times $\hat{t}$.}
\label{fig:water-eig}
\end{figure}

\begin{figure}[!ht]
\centering
\begin{subfigure}{0.24\textwidth}
    \centering
    \includegraphics[width=\textwidth]{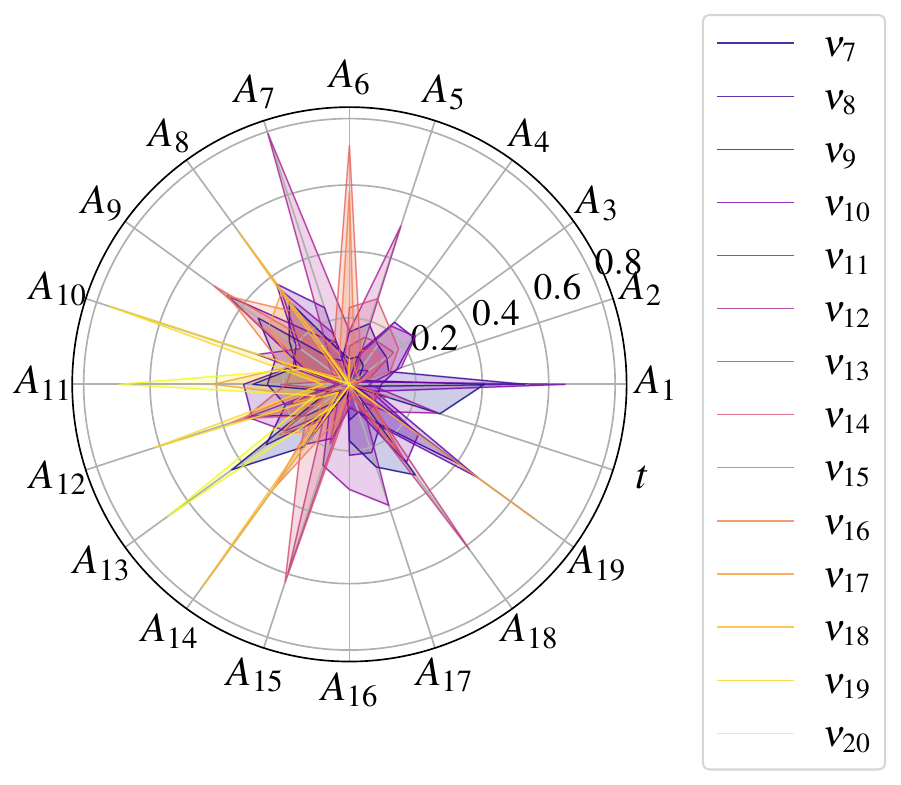}
    \caption{$\hat{t}=12.22145$ for $\bm{y}^+$.}
\end{subfigure}
\begin{subfigure}{0.24\textwidth}
    \centering
    \includegraphics[width=\textwidth]{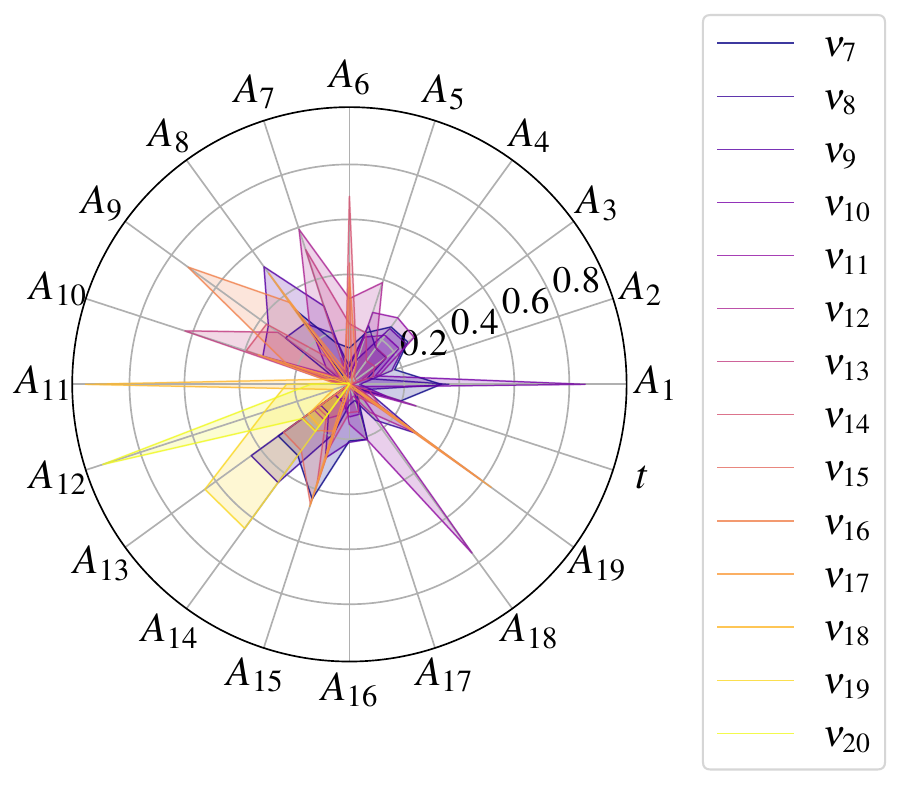}
    \caption{$\hat{t}=13.78993$ for $\bm{y}^+$.}
\end{subfigure}
\begin{subfigure}{0.24\textwidth}
    \centering
    \includegraphics[width=\textwidth]{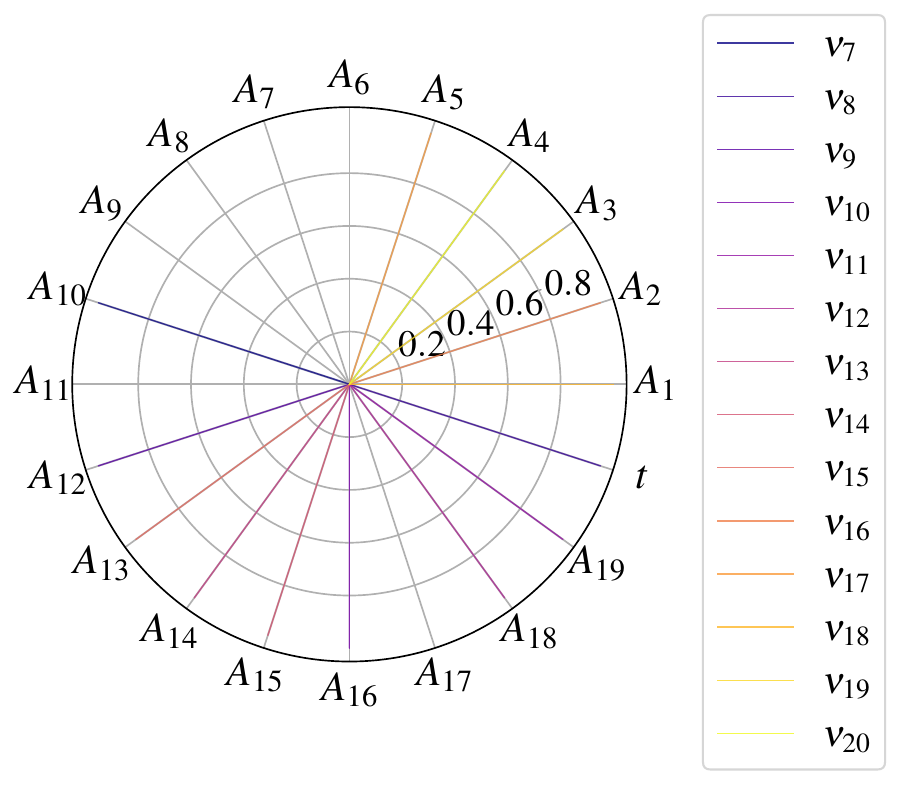}
    \caption{$\hat{t}=35.73403$ for $\bm{y}^*$.}
\end{subfigure}
\begin{subfigure}{0.24\textwidth}
    \centering
    \includegraphics[width=\textwidth]{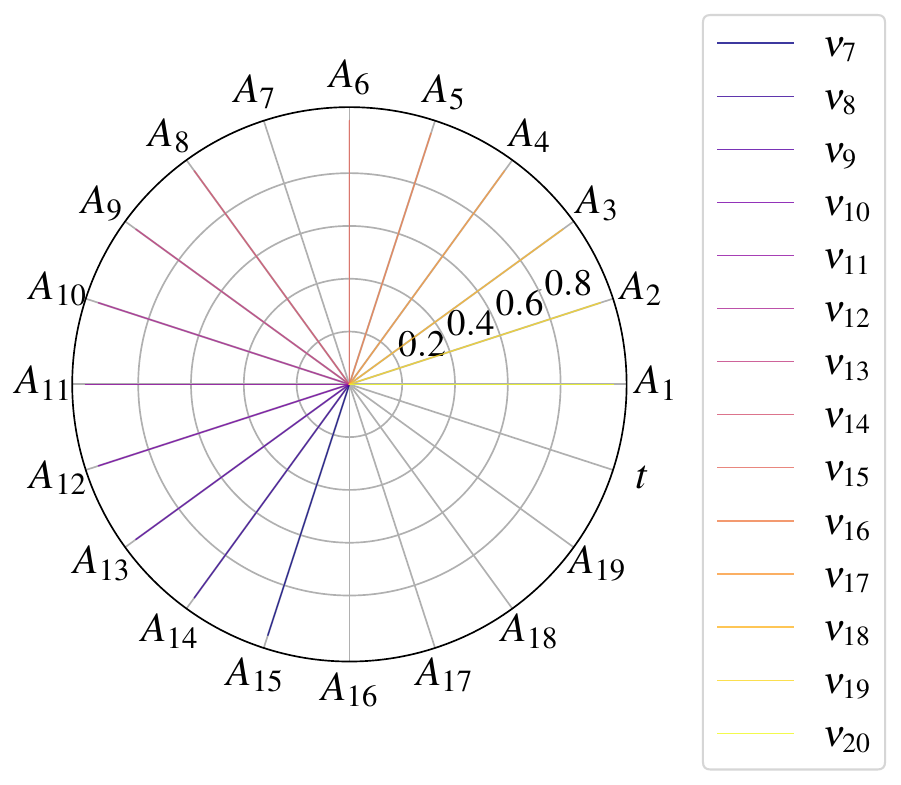}
    \caption{$\hat{t}=38.63715$ for $\bm{y}^*$.}
\end{subfigure}
\caption{Radar plots for the singular eigenvectors of the hydrogen-air system at \emph{pre-ignition} concentrations $\bm{y}^{+}$ and \emph{post-ignition} concentrations $\bm{y}^{*}$. Colors (from dark to bright) reflect a decreasing magnitude of the respective eigenvalues.}
\label{fig:water-radar}
\end{figure}

In the formulation of the inverse problem for the hydrogen-air kinetics system, we have kept the $b_i, E_i$ fixed and varied the $A_i$ in a $\pm 50\%$ range of the nominal values for the $i$-th species. Since the rate constants are effectively $k_i=A_iT^{b_i}e^{-E_i/RT}$, varying $A_i$ significantly changes the magnitudes of the rate constants. 
The InVAErt framework has been applied to the hydrogen-air kinetics system using a latent space of dimension 13 and 6000 training epochs, resulting in an accuracy of 99.97\%.
Inference results in Figure~\ref{fig:water-inv-trajectories} and~\ref{fig:water-inv-parallel} are obtained with 2 rounds of PC sampling, from the concentrations
\[
\begin{aligned}
\bm{y}^{*} =[& 1.34797745\times10^{-7},\; 3.24215215\times10^{-8},\; 4.97295147\times10^{-8},\; 1.20362742\times10^{-8},\;
7.89690731\times10^{-8},\\ & 1.82293677\times10^{-11},\; 1.32417482\times10^{-12},\; 1.36445362\times10^{-6},\;
1.09905202\times10^{-4}].
\end{aligned}
\]
Log-log sensitivities are computed for $A_j$ and $t$, leading to the sensitivity matrix
\[
J_{[i, j]} =
\begin{cases}
\displaystyle \frac{\partial \log y_i}{\partial \log A_j} = \frac{A_j}{y_i} \frac{\partial y_i}{\partial A_j} , & j < 20 \\
\\
\displaystyle \frac{\partial \log y_i}{\partial \log t} = \frac{t}{y_i} \frac{\partial y_i}{\partial t} , & j = 20.
\end{cases}
\]
Here, $\bm{v}=(\bm{A}, t)$ where $\bm{A}=(A_{1},\dots,A_{N_{\bm{y}}})$, and $\Delta \widehat{\bm{v}}$ is selected equal to $0.1\%$ of $\widehat{\bm{v}}$ in the central difference approximation, since the system has very high stiffness. 
Identifiability results are shown for the same $\bm{y}^*$ as in Figure~\ref{fig:water-inv-trajectories}, and for 20 parameters combinations $\widehat{\bm{v}}$ recovered from the inverse problem solution.
Additionally, from Figure~\ref{fig:water-inv-trajectories}, we can see that $\bm{y}^*$ belongs to the \emph{post-ignition} regime, where the system exhibits quick progression towards the steady state.
Hence, the system is \emph{locally} non-identifiable at $\bm{y}^*$, as the same steady state concentrations are obtained over a wide range of kinetic rate parameters, as evident from Figure~\ref{fig:water_perturbed}, whereas knowledge of the concentrations at the ignition peak would instead lead to improved identifiability. 
This is confirmed in Figure~\ref{fig:water-eig} which shows a more pronounced decay of the eigenvalues with increase in $\hat{t}$, i.e., as we move further away from ignition, towards the steady state. 
The radar plots in Figure~\ref{fig:water-radar} also show an increasing dominance of the post-ignition singular eigenvectors, i.e., the parameters are increasingly unimportant. 
Conversely, inversion from concentrations $\bm{y}^{+}$ obtained \emph{before} ignition, lead to larger eigenvalues as we increase $\hat{t}$, as shown in Figure~\ref{fig:water-eig}, hence improved identifiability.

\begin{figure}[!ht]
\centering
\begin{subfigure}{0.9\textwidth}
    \centering
    \includegraphics[width=\textwidth]{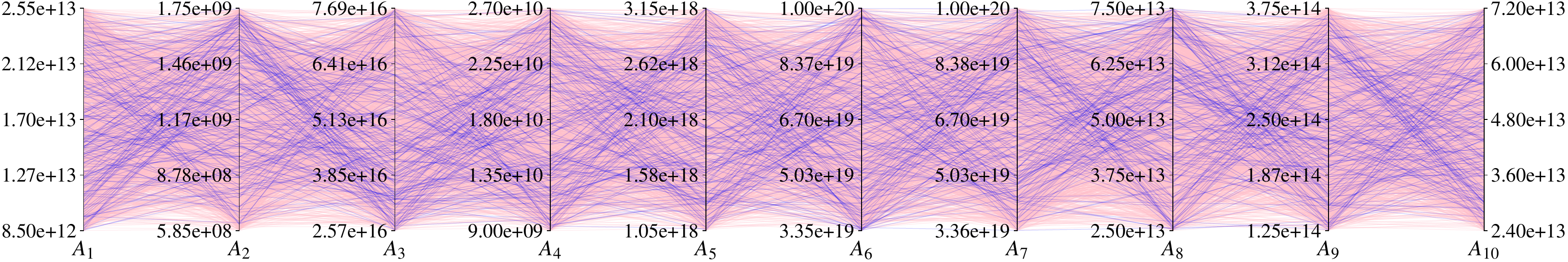}
\end{subfigure}
\begin{subfigure}{0.9\textwidth}
    \centering
    \includegraphics[width=\textwidth]{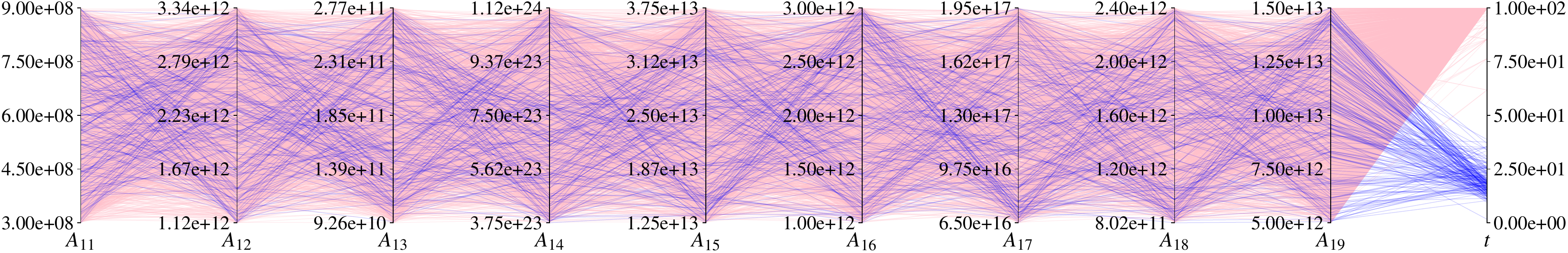}
    \caption{$\bm{y}^{+}$ pre-ignition, $\varepsilon=10^{-4}$.}
    \label{fig:water-epsilon-pre}
\end{subfigure}
\\
\vspace{2mm}
\begin{subfigure}{0.9\textwidth}
    \centering
    \includegraphics[width=\textwidth]{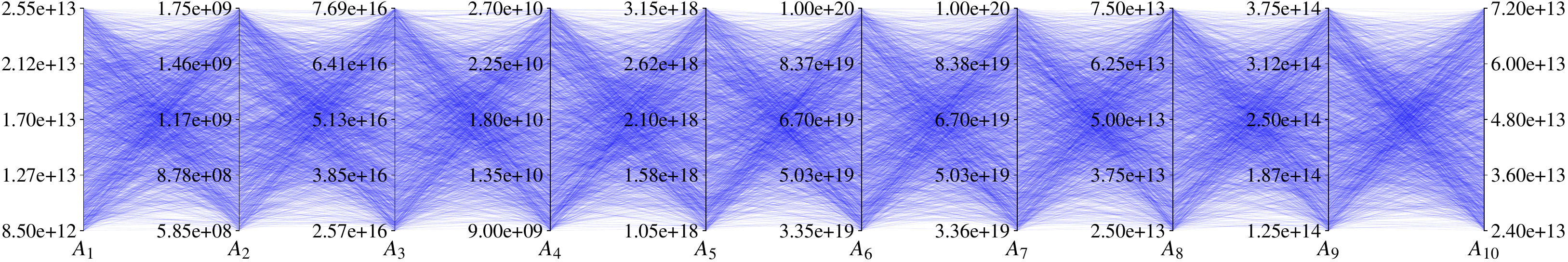}
\end{subfigure}
\begin{subfigure}{0.9\textwidth}
    \centering
    \includegraphics[width=\textwidth]{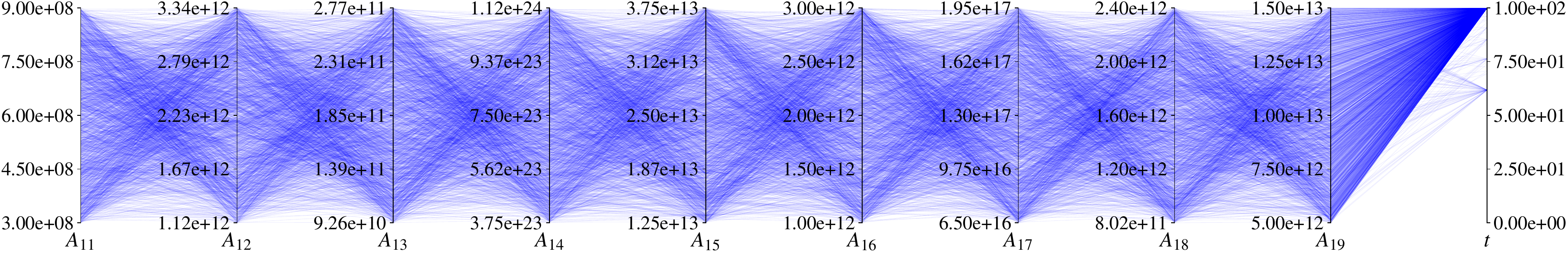}
    \caption{$\bm{y}^*$ post-ignition, $\varepsilon=10^{-8}$.}
    \label{fig:water-epsilon-post}
\end{subfigure}
\caption{DSS of hydrogen-air system using target concentrations selected from pre- and post-ignition conditions.}
\label{fig:water-epsilon}
\end{figure}

Finally, DSS results confirms this intuition. For a \emph{pre-ignition} $\bm{y}^{+}$, we were able to extract multiple trajectories satisfying the $\varepsilon$-bound ($\varepsilon=10^{-4}$), as shown in Figure~\ref{fig:water-epsilon-pre}.
However, for \emph{post-ignition} $\bm{y}^*$, Figure~\ref{fig:water-epsilon-post} shows how almost all parameter combinations in the dataset satisfy even a bound as tight as $10^{-8}$.


% ============================================================
\subsubsection{Identifiability under finite solution accuracy}
% ============================================================

\begin{figure}[!ht]
    \centering
    \includegraphics[width=0.7\linewidth]{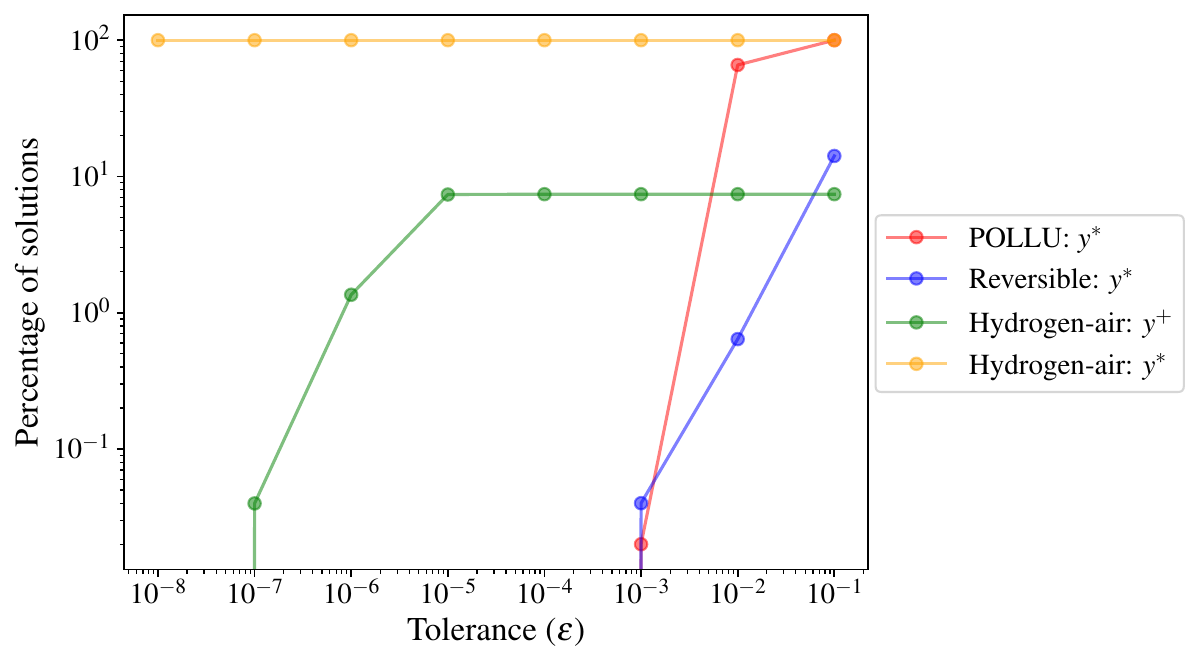}
    \caption{Percentage of nearly-identifiable prior parameter combinations identified by DSS vs. $L^2$ relative error tolerance ($\varepsilon$).}
    \label{fig: identifiability-consolidated}
\end{figure}

We finally consider conditions under which the systems analyzed in the previous sections are only \emph{nearly-identifiable}, i.e., there exist identifiable solutions $\widehat{\bm{v}}$ that are only \emph{close} (e.g. in $L^{2}$ sense) to the sought $\bm{y}^{*}$.
To analyze such conditions, we applied DSS by first sampling input parameters uniformly from their respective prior, then counting all solutions found within a distance $\varepsilon$ of $\bm{y}^{*}$, while progressively increasing $\varepsilon$.
Results are presented in Figure~\ref{fig: identifiability-consolidated}.

For the hydrogen-air system under post-ignition conditions, all reaction rates lead to the same steady state concentrations regardless of $\varepsilon$, resulting in a horizontal line indicating that 100\% of prior samples reproduce the desired output $\bm{y}^{*}$.
Conversely, under pre-ignition conditions, the number of solutions in the manifold $\mathcal{M}_{\bm{y}^{*}}$ \emph{gradually shrinks} to zero as the threshold is reduced, while eventually approaching 100\% as $\varepsilon$ is increased.
Figure~\ref{fig: identifiability-consolidated} thus provides a more complete picture for the varying degree of identifiability between pre- and post-ignition inverse solutions for the hydrogen-air mechanism, while also characterizing the density of nearly identifiable solutions within the selected prior. 
The POLLU problem and the system with reversible kinetics exhibit a different behavior: no solution is detected up to approximately $\varepsilon=10^{-3}$, beyond which a steep increase in the number of solutions is observed.
The rate at which nearly identifiable solutions accumulate with increasing $\varepsilon$ can therefore serve as an additional indicator to differentiate the (nearly-) identifiability properties of chemical kinetics models.

% ========================================================
\section{Discussion and Conclusions}\label{sec:discussion}
% ========================================================

This study demonstrates the effectiveness of inVAErt networks with ResNet and LSTM emulators in providing a unified apparatus for forward emulation and inverse parameter recovery for stiff chemical reaction systems.

We show that emulators can achieve near-perfect accuracy over families of stiff dynamical systems parametrized by the respective rate constants, varying over the predetermined prior ranges. 
While ResNet emulators work perfectly well for low-dimensional systems with limited stiffness, error accumulation during rollout can be excessive for stiff systems with more complex dynamics, requiring a finer discretization in time to resolve sharp concentration gradients.
This limitation is mitigated by introducing LSTM emulators. In particular, the addition of a conditioning mechanism to the LSTM encoder has the most significant impact in reducing errors accumulated during rollout.

Furthermore, we focus on model inversion, leveraging the variational autoencoder component of the inVAErt network to generate samples from the non-identifiable manifold of reaction rate constants associated with given system concentrations.
For systems with relatively low dimensionality, we are able to formally verify the correctness for the inverse problem solutions computed through the proposed approach. 
An inVAErt network is also capable to suggest inverse solutions going beyond prior input combinations observed during training.
For higher dimensional systems the results obtained through an inVAErt network are consistent with those obtained from various approaches for local identifiability analysis proposed in the literature.

In conclusion, our approach provides an effective new paradigm leveraging amortization to solve forward and inverse problems in the context of stiff dynamical systems, circumventing the need for repeated numerical integration, while simultaneously providing almost instantaneous parameter retrieval without retraining.

% ========================================
\section*{Acknowledgments}\label{sec:ackw}
% ========================================

The authors acknowledge the Center for Research Computing at the University of Notre Dame for providing the computational resources essential to this study. They also acknowledge support from NIH grant \#1R01HL167516. The authors would like to thank Dr. Joseph M. Powers for providing working examples, offering guidance throughout the course of this study, and providing feedback on this work and Dr. Shivam Barwey for his feedback on earlier versions of this manuscript.

\bibliography{main}

@book{law-of-mass-action,
  title={Mathematical Models of Chemical Reactions: Theory and Applications of Deterministic and Stochastic Models},
  author={{\'E}rdi, P. and Toth, J. and T{\'o}th, J.},
  isbn={9780691085326},
  lccn={88039781},
  series={Nonlinear science : theory and applications},
  year={1989},
  publisher={Princeton University Press}
}

@article{rober,
  title={The solution of a set of reaction rate equations},
  author={Robertson, H.H.},
  journal={Numerical Analysis: An Introduction},
  volume={178182},
  year={1966}
}

@article{verwer1994gauss,
  title={Gauss--{S}eidel iteration for stiff {ODE}s from chemical kinetics},
  author={Verwer, Jan G.},
  journal={SIAM Journal on Scientific Computing},
  volume={15},
  number={5},
  pages={1243--1250},
  year={1994},
  publisher={SIAM}
}

@article{jmp-water,
  title={Analysis of the spatio-temporal scales of laminar premixed flames near equilibrium},
  author={Al-Khateeb, Ashraf N. and Powers, Joseph M. and Paolucci, Samuel},
  journal={Combustion Theory and Modelling},
  volume={17},
  number={1},
  pages={76--108},
  year={2013},
  publisher={Taylor \& Francis}
}

@article{cond-lstm,
  title={Conditional recurrent neural networks for broad applications in nonlinear optics},
  author={Lauria, Simone and Saleh, Mohammed F.},
  journal={Optics Express},
  volume={32},
  number={4},
  pages={5582--5591},
  year={2024},
  publisher={Optica Publishing Group}
}

@misc{lstm-encoder-decoder,
  author       = {Kulowski, Lucas },
  title        = {LSTM\_encoder\_decoder },
  year         = {2020},
  howpublished = {GitHub repository: \url{https://github.com/lkulowski/LSTM_encoder_decoder}},
  note         = {}
}

@article{grayson-lstm,
  title={Data-driven synchronization-avoiding algorithms in the explicit distributed structural analysis of soft tissue},
  author={Tong, Guoxiang Grayson and Schiavazzi, Daniele E.},
  journal={Computational Mechanics},
  volume={71},
  number={3},
  pages={453--479},
  year={2023},
  publisher={Springer}
}

@article{owoyele2022chemnode,
  title={{ChemNODE}: A neural ordinary differential equations framework for efficient chemical kinetic solvers},
  author={Owoyele, Opeoluwa and Pal, Pinaki},
  journal={Energy and AI},
  volume={7},
  pages={100118},
  year={2022},
  publisher={Elsevier},
  doi={10.1016/j.egyai.2021.100118}
}

@article{kim2021stiff,
  title={Stiff neural ordinary differential equations},
  author={Kim, Suyong and Ji, Weiqi and Deng, Sili and Ma, Yingbo and Rackauckas, Christopher},
  journal={Chaos: An Interdisciplinary Journal of Nonlinear Science},
  volume={31},
  number={9},
  pages={093122},
  year={2021},
  publisher={AIP Publishing},
  doi={10.1063/5.0060697}
}

@article{ji2021stiff,
  title={{Stiff-PINN}: Physics-informed neural network for stiff chemical kinetics},
  author={Ji, Weiqi and Qiu, Weilun and Shi, Zhiyu and Pan, Shaowu and Deng, Sili},
  journal={The Journal of Physical Chemistry A},
  volume={125},
  number={36},
  pages={8098--8106},
  year={2021},
  publisher={ACS Publications},
  doi={10.1021/acs.jpca.1c05102}
}

@article{weng2022multiscale,
  title={Multiscale physics-informed neural networks for stiff chemical kinetics},
  author={Weng, Yuting and Zhou, Dezhi},
  journal={The Journal of Physical Chemistry A},
  volume={126},
  number={45},
  pages={8534--8543},
  year={2022},
  publisher={ACS Publications},
  doi={10.1021/acs.jpca.2c06513}
}

@article{kumar2024physics,
  title={A posteriori evaluation of a physics-constrained neural ordinary differential equations approach coupled with {CFD} solver for modeling stiff chemical kinetics},
  author={Kumar, Tadbhagya and Kumar, Anuj and Pal, Pinaki},
  journal={arXiv preprint arXiv:2312.00038},
  year={2024},
}

@article{goswami2024learning,
  title={Learning stiff chemical kinetics using extended deep neural operators},
  author={Goswami, Somdatta and Jagtap, Ameya D. and Babaee, Hessam and Susi, Bryan T. and Karniadakis, George Em},
  journal={Computer Methods in Applied Mechanics and Engineering},
  volume={419},
  pages={116674},
  year={2024},
  publisher={Elsevier},
  doi={10.1016/j.cma.2023.116674}
}

@article{echekki2024combustion,
  title={Combustion chemistry acceleration with {DeepONets}},
  author={Echekki, Tarek and Farjoo, Majid and Owoyele, Opeoluwa},
  journal={Fuel},
  volume={365},
  pages={131212},
  year={2024},
  publisher={Elsevier},
  doi={10.1016/j.fuel.2024.131212}
}

@article{weng2025extended,
  title={Extended {Fourier} neural operators to learn stiff chemical kinetics under unseen conditions},
  author={Weng, Yuting and Li, Han and Zhang, Hao and Chen, Zhi X and Zhou, Dezhi},
  journal={Combustion and Flame},
  volume={272},
  pages={113847},
  year={2025},
  publisher={Elsevier},
  doi={10.1016/j.combustflame.2024.113847}
}

@article{wang2025enforcing,
  title={Enforcing physical conservation in neural network surrogate models for complex chemical kinetics},
  author={Wang, Tinghao and Yi, Yuxiao and Yao, Junjie and Xu, Zhi-Qin John and Zhang, Tianhan and Chen, Zheng},
  journal={Combustion and Flame},
  volume={275},
  pages={114105},
  year={2025},
  publisher={Elsevier},
  doi={10.1016/j.combustflame.2025.114105}
}

@article{zhang2022multiscale,
  title={A multi-scale sampling method for accurate and robust deep neural network to predict combustion chemical kinetics},
  author={Zhang, Tianhan and Shu, Yuxiao and Jiang, Xi and Yi, Pengfei and Xu, Chengxi and Wang, Ziming and Wang, Huangwei},
  journal={Combustion and Flame},
  volume={245},
  pages={112319},
  year={2022},
  publisher={Elsevier},
  doi={10.1016/j.combustflame.2022.112319}
}

@article{doppel2022efficient,
  title={Efficient machine learning based surrogate models for surface kinetics by approximating the rates of the rate-determining steps},
  author={D{\"o}ppel, Felix A and Votsmeier, Martin},
  journal={Chemical Engineering Science},
  volume={262},
  pages={117964},
  year={2022},
  publisher={Elsevier},
  doi={10.1016/j.ces.2022.117964}
}

@article{jagtap2024amore,
  title={{AMORE}: Adaptive multi-output operator network for stiff chemical kinetics},
  author={Jagtap, Ameya D and Mao, Zhiping and Karniadakis, George Em},
  journal={arXiv preprint arXiv:2510.12999},
  year={2024},
}

@article{christo1996,
  title={An integrated {PDF}/neural network approach for simulating turbulent reacting systems},
  author={Christo, Farid C. and Masri, Assaad R. and Nebot, Eduardo Mario and Pope, Stephen B.},
  journal={Symposium (International) on Combustion},
  volume={26},
  number={1},
  pages={43--48},
  year={1996},
  publisher={Elsevier},
  doi={10.1016/S0082-0784(96)80198-6}
}

@article{blasco1998,
  title={Modelling the temporal evolution of a reduced combustion chemical system with an artificial neural network},
  author={Blasco, J.A. and Fueyo, N. and Dopazo, C. and Ballester, J.},
  journal={Combustion and Flame},
  volume={113},
  number={1-2},
  pages={38--52},
  year={1998},
  publisher={Elsevier},
  doi={10.1016/S0010-2180(97)00211-3}
}

@article{schwaller2019molecular,
  title={Molecular transformer: A model for uncertainty-calibrated chemical reaction prediction},
  author={Schwaller, Philippe and Laino, Teodoro and Gaudin, Th{\'e}ophile and Bolgar, Peter and Hunter, Christopher A. and Bekas, Costas and Lee, Alpha A.},
  journal={ACS Central Science},
  volume={5},
  number={9},
  pages={1572--1583},
  year={2019},
  publisher={ACS Publications},
  doi={10.1021/acscentsci.9b00576}
}

@article{heid2022machine,
  title={Machine learning of reaction properties via learned representations of the condensed graph of reaction},
  author={Heid, Esther and Green, William H},
  journal={Journal of Chemical Information and Modeling},
  volume={62},
  number={9},
  pages={2101--2110},
  year={2022},
  publisher={ACS Publications},
  doi={10.1021/acs.jcim.1c00975}
}

@article{kircher2023global,
  title={Learning kinetics from non-ideal reactors by implicitly solved finite volumes and global reaction neural networks},
  author={Kircher, Tim and Votsmeier, Martin},
  journal={Chemical Engineering Journal},
  volume={522},
  pages={166749},
  year={2025},
  publisher={Elsevier},
  doi={10.1016/j.cej.2025.166749}
}

@book{powers2016combustion,
  title={Combustion Thermodynamics and Dynamics},
  author={Powers, Joseph M.},
  year={2016},
  publisher={Cambridge University Press}
}

@article{tong2024invaert,
  title={In{VAE}rt networks: A data-driven framework for model synthesis and identifiability analysis},
  author={Tong, Guoxiang Grayson and Sing Long, Carlos A. and Schiavazzi, Daniele E.},
  journal={Computer Methods in Applied Mechanics and Engineering},
  volume={423},
  pages={116846},
  year={2024},
  publisher={Elsevier}
}

@inproceedings{dinh2017density,
title={Density estimation using {R}eal {NVP}},
author={Laurent Dinh and Jascha Sohl-Dickstein and Samy Bengio},
booktitle={International Conference on Learning Representations},
year={2017},
url={https://openreview.net/forum?id=HkpbnH9lx}
}

@article{kobyzev2020normalizing,
  title={Normalizing flows: An introduction and review of current methods},
  author={Kobyzev, Ivan and Prince, Simon J.D. and Brubaker, Marcus A.},
  journal={IEEE transactions on pattern analysis and machine intelligence},
  volume={43},
  number={11},
  pages={3964--3979},
  year={2020},
  publisher={IEEE}
}

@article{papamakarios2021normalizing,
  title={Normalizing flows for probabilistic modeling and inference},
  author={Papamakarios, George and Nalisnick, Eric and Rezende, Danilo Jimenez and Mohamed, Shakir and Lakshminarayanan, Balaji},
  journal={Journal of Machine Learning Research},
  volume={22},
  number={57},
  pages={1--64},
  year={2021}
}

@inproceedings{kingma2013auto,
  author = {Kingma, Diederik P. and Welling, Max},
  title = {{Auto-Encoding Variational Bayes}},
  booktitle = {2nd International Conference on Learning Representations, {ICLR} 2014, Banff, AB, Canada, April 14-16, 2014, Conference Track Proceedings},
  year = {2014},
  url = {http://arxiv.org/abs/1312.6114}
}

@article{kiefer2011laser,
  title={Laser diagnostics and minor species detection in combustion using resonant four-wave mixing},
  author={Kiefer, Johannes and Ewart, Paul},
  journal={Progress in Energy and Combustion Science},
  volume={37},
  number={5},
  pages={525--564},
  year={2011},
  publisher={Elsevier}
}

@article{gasparyan2023parameter,
  title={Parameter estimation for models of chemical reaction networks from experimental data of reaction rates},
  author={Gasparyan, Manvel and Van Messem, Arnout and Rao, Shodhan},
  journal={International Journal of Control},
  volume={96},
  number={2},
  pages={392--407},
  year={2023},
  publisher={Taylor \& Francis}

}

@inproceedings{he2016deep,
  title={Deep residual learning for image recognition},
  author={He, Kaiming and Zhang, Xiangyu and Ren, Shaoqing and Sun, Jian},
  booktitle={Proceedings of the IEEE conference on Computer Vision and Pattern Recognition},
  pages={770--778},
  year={2016}
}

@book{wanner1996solving,
  title={Solving Ordinary Differential Equations II},
  author={Wanner, Gerhard and Hairer, Ernst},
  volume={375},
  year={1996},
  publisher={Springer Berlin Heidelberg New York}
}

@techreport{kee1989chemkin,
  title={Chemkin-{II}: A Fortran chemical kinetics package for the analysis of gas-phase chemical kinetics},
  author={Kee, Robert J. and Rupley, Fran M. and Miller, James A.},
  year={1989},
  institution={Sandia National Laboratories (SNL-CA), Livermore, CA (United States)}
}

@article{devroye2006nonuniform,
  title={Nonuniform random variate generation},
  author={Devroye, Luc},
  journal={Handbooks in Operations Research and Management Science},
  volume={13},
  pages={83--121},
  year={2006},
  publisher={Elsevier}
}

@article{de2025quantification,
  title={Quantification of total uncertainty in the physics-informed reconstruction of {CVSim-6} physiology},
  author={De Florio, Mario and Zou, Zongren and Schiavazzi, Daniele E. and Karniadakis, George Em},
  journal={Philosophical Transactions A},
  volume={383},
  number={2292},
  pages={20240221},
  year={2025},
  publisher={The Royal Society}
}

@article{hochreiter1996lstm,
  title={{LSTM} can solve hard long time lag problems},
  author={Hochreiter, Sepp and Schmidhuber, J{\"u}rgen},
  journal={Advances in Neural Information Processing Systems},
  volume={9},
  year={1996}
}

@article{baumgartner2025multiparameter,
  title={A multiparameter singular perturbation analysis of the {R}obertson model},
  author={Baumgartner, Lukas and Szmolyan, Peter},
  journal={Studies in Applied Mathematics},
  volume={154},
  number={2},
  pages={e70020},
  year={2025},
  publisher={Wiley Online Library}
}

@article{qin2019data,
  title={Data driven governing equations approximation using deep neural networks},
  author={Qin, Tong and Wu, Kailiang and Xiu, Dongbin},
  journal={Journal of Computational Physics},
  volume={395},
  pages={620--635},
  year={2019},
  publisher={Elsevier}
}

@article{law2006combustion,
  title={Combustion Physics.},
  author={Law, C.K.},
  journal={Cambridge Univ. Press, N. J},
  year={2006}
}

@article{koenig2025chemkans,
  title={Chem{KAN}s for combustion chemistry modeling and acceleration},
  author={Koenig, Benjamin C and Kim, Suyong and Deng, Sili},
  journal={Physical Chemistry Chemical Physics},
  volume={27},
  number={33},
  pages={17313--17330},
  year={2025},
  publisher={Royal Society of Chemistry}
}

@article{vijayarangan2024data,
  title={A data-driven reduced-order model for stiff chemical kinetics using dynamics-informed training},
  author={Vijayarangan, Vijayamanikandan and Uranakara, Harshavardhana A and Barwey, Shivam and Galassi, Riccardo Malpica and Malik, Mohammad Rafi and Valorani, Mauro and Raman, Venkat and Im, Hong G},
  journal={Energy and AI},
  volume={15},
  pages={100325},
  year={2024},
  publisher={Elsevier}
}

\appendix

% ======================================================================
\section{Network architecture and hyperparameters}\label{sec:arch_hyper}
% ======================================================================

Hyperparameter choices for the Robertson, POLLU, reversible, irreversible and hydrogen-air systems are reported in Table~\ref{tab:hyp-rober}, \ref{tab:hyp-pollu}, \ref{tab:hyp-rev}, \ref{tab:hyp-irr} and~\ref{tab:hyp-water}, respectively. 

\begin{table}[H]
    \centering
    \begin{tabular}{ll}
    \toprule
       $n_p, n_f$ for $\mathcal{NN}_e$ (LSTM) & $\{10, 50\}$ \\
       Number of Encoder and Decoder layers for $\mathcal{NN}_e$ & $\{2,1\}$\\
       Hidden layer size for $\mathcal{NN}_e$ & 50 \\
       Parameters for conditioning in $\mathcal{NN}_e$ & $[50, 4, ReLU]$ \\
       Initial learning rate for $\mathcal{NN}_e$, $(\mathcal{NN}_v, \mathcal{NN}_d)$  & $\{5\times 10^{-4}, 10^{-3}\}$ \\
       Learning rate decay rate for $\mathcal{NN}_e$, $(\mathcal{NN}_v, \mathcal{NN}_d)$  &  $\{0.998, 0.985\}$ \\
        Mini-batch size for $\mathcal{NN}_e$, $(\mathcal{NN}_v, \mathcal{NN}_d)$ &  $\{512, 512\}$ \\
        Parameters of $\mathcal{NN}_v$ & $[32, 8, SiLU]$\\
        Parameters of $\mathcal{NN}_d$ & $[64, 10, SiLU]$\\
        Latent space dimension & 6 \\
        $L^2$-weight decay rate for $(\mathcal{NN}_v, \mathcal{NN}_d)$ & 0 \\
        Loss penalties $\beta_v, \beta_d, \beta_r$ & $[400, 1, 50]$\\
        \bottomrule
    \end{tabular}
    \caption{InVAErt network hyperparameters used for the Robertson system.}
    \label{tab:hyp-rober}
\end{table}

\begin{table}[H]
    \centering
    \begin{tabular}{ll}
    \toprule
       $n_p, n_f$ for $\mathcal{NN}_e$ (LSTM) & $\{10, 50\}$ \\
       Number of Encoder and Decoder layers for $\mathcal{NN}_e$ & $\{2,1\}$\\
       Hidden layer size for $\mathcal{NN}_e$ & 50 \\
       Parameters for conditioning in $\mathcal{NN}_e$ & $[50, 4, ReLU]$ \\
       Initial learning rate for $\mathcal{NN}_e$, $(\mathcal{NN}_v, \mathcal{NN}_d)$  & $\{5\times 10^{-4}, 10^{-3}\}$ \\
       Learning rate decay rate for $\mathcal{NN}_e$, $(\mathcal{NN}_v, \mathcal{NN}_d)$  &  $\{0.998, 0.98\}$ \\
        Mini-batch size for $\mathcal{NN}_e$, $(\mathcal{NN}_v, \mathcal{NN}_d)$ &  $\{512, 512\}$ \\
        Parameters of $\mathcal{NN}_v$ & $[32, 6, SiLU]$\\
        Parameters of $\mathcal{NN}_d$ & $[156, 12, SiLU]$\\
        Latent space dimension & 8\\
        $L^2$-weight decay rate for $(\mathcal{NN}_v, \mathcal{NN}_d)$ & 0 \\
        Loss penalties $\beta_v, \beta_d, \beta_r$ & $[1, 6000, 0]$\\
        \bottomrule
    \end{tabular}
    \caption{InVAErt network hyperparameters used for the POLLU system.}
    \label{tab:hyp-pollu}
\end{table}

\begin{table}[H]
    \centering
    \begin{tabular}{ll}
    \toprule
       $n_p$ for $\mathcal{NN}_e$ (ResNet) & $10$ \\
       Initial learning rate for $\mathcal{NN}_e$, $(\mathcal{NN}_v, \mathcal{NN}_d)$  & $\{10^{-3}, 10^{-3}\}$ \\
       Learning rate decay rate for $\mathcal{NN}_e$, $(\mathcal{NN}_v, \mathcal{NN}_d)$  &  $\{0.9,0.985\}$ \\
        Mini-batch size for $\mathcal{NN}_e$, $(\mathcal{NN}_v, \mathcal{NN}_d)$ &  $\{1024, 512\}$ \\
        Parameters of $\mathcal{NN}_e$ & $[16, 10, SiLU]$ \\
        Parameters of $\mathcal{NN}_v$ & $[32, 6, SiLU]$\\
        Parameters of $\mathcal{NN}_d$ & $[64, 8, SiLU]$\\
        Latent space dimension & 8 \\
        $L^2$-weight decay rate for $(\mathcal{NN}_v, \mathcal{NN}_d)$ & $\{0, 0\}$\\
        Loss penalties $\beta_v, \beta_d, \beta_r$ & $[1, 1000, 0]$\\
        \bottomrule
    \end{tabular}
    \caption{InVAErt network hyperparameters used for the reversible system.}
    \label{tab:hyp-rev}
\end{table}

\begin{table}[H]
    \centering
    \begin{tabular}{ll}
    \toprule
       $n_p$ for $\mathcal{NN}_e$ (ResNet) & $10$ \\
       Initial learning rate for $\mathcal{NN}_e$, $(\mathcal{NN}_v, \mathcal{NN}_d)$  & $\{10^{-3}, 10^{-3}\}$ \\
       Learning rate decay rate for $\mathcal{NN}_e$, $(\mathcal{NN}_v, \mathcal{NN}_d)$  &  $\{0.9,0.985\}$ \\
        Mini-batch size for $\mathcal{NN}_e$, $(\mathcal{NN}_v, \mathcal{NN}_d)$ &  $\{1024, 512\}$ \\
        Parameters of $\mathcal{NN}_e$ & $[16, 10, SiLU]$ \\
        Parameters of $\mathcal{NN}_v$ & $[32, 6, SiLU]$\\
        Parameters of $\mathcal{NN}_d$ & $[64, 8, SiLU]$\\
        Latent space dimension & 6 \\
        $L^2$-weight decay rate for $(\mathcal{NN}_v, \mathcal{NN}_d)$ & $\{0, 0\}$\\
        Loss penalties $\beta_v, \beta_d, \beta_r$ & $[1, 1000, 0]$\\
        \bottomrule
    \end{tabular}
    \caption{InVAErt network hyperparameters used for the irreversible system.}
    \label{tab:hyp-irr}
\end{table}

\begin{table}[H]
    \centering
    \begin{tabular}{ll}
    \toprule
       $n_p, n_f$ for $\mathcal{NN}_e$ (LSTM) & $\{10, 50\}$ \\
       Number of Encoder and Decoder layers for $\mathcal{NN}_e$ & $\{2,1\}$\\
       Hidden layer size for $\mathcal{NN}_e$ & 50 \\
       Parameters for conditioning in $\mathcal{NN}_e$ & $[50, 4, ReLU]$ \\
       Initial learning rate for $\mathcal{NN}_e$, $(\mathcal{NN}_v, \mathcal{NN}_d)$  & $\{5\times 10^{-4}, 10^{-3}\}$ \\
       Learning rate decay rate for $\mathcal{NN}_e$, $(\mathcal{NN}_v, \mathcal{NN}_d)$  &  $\{0.998, 0.98\}$ \\
        Mini-batch size for $\mathcal{NN}_e$, $(\mathcal{NN}_v, \mathcal{NN}_d)$ &  $\{512, 512\}$ \\
        Parameters of $\mathcal{NN}_v$ & $[32, 8, SiLU]$\\
        Parameters of $\mathcal{NN}_d$ & $[256, 8, SiLU]$\\
        Latent space dimension & 13 \\
        $L^2$-weight decay rate for $(\mathcal{NN}_v, \mathcal{NN}_d)$ & $\{0, 0\}$\\
        Loss penalties $\beta_v, \beta_d, \beta_r$ & $[1, 6000, 100]$\\
        \bottomrule
    \end{tabular}
    \caption{InVAErt network hyperparameters used for the hydrogen-air system.}
    \label{tab:hyp-water}
\end{table}

% =====================================================
\section{Additional figures}\label{sec:appendix-plots}
% =====================================================

\begin{figure}[!ht]
\centering
\includegraphics[width=0.83\textwidth]{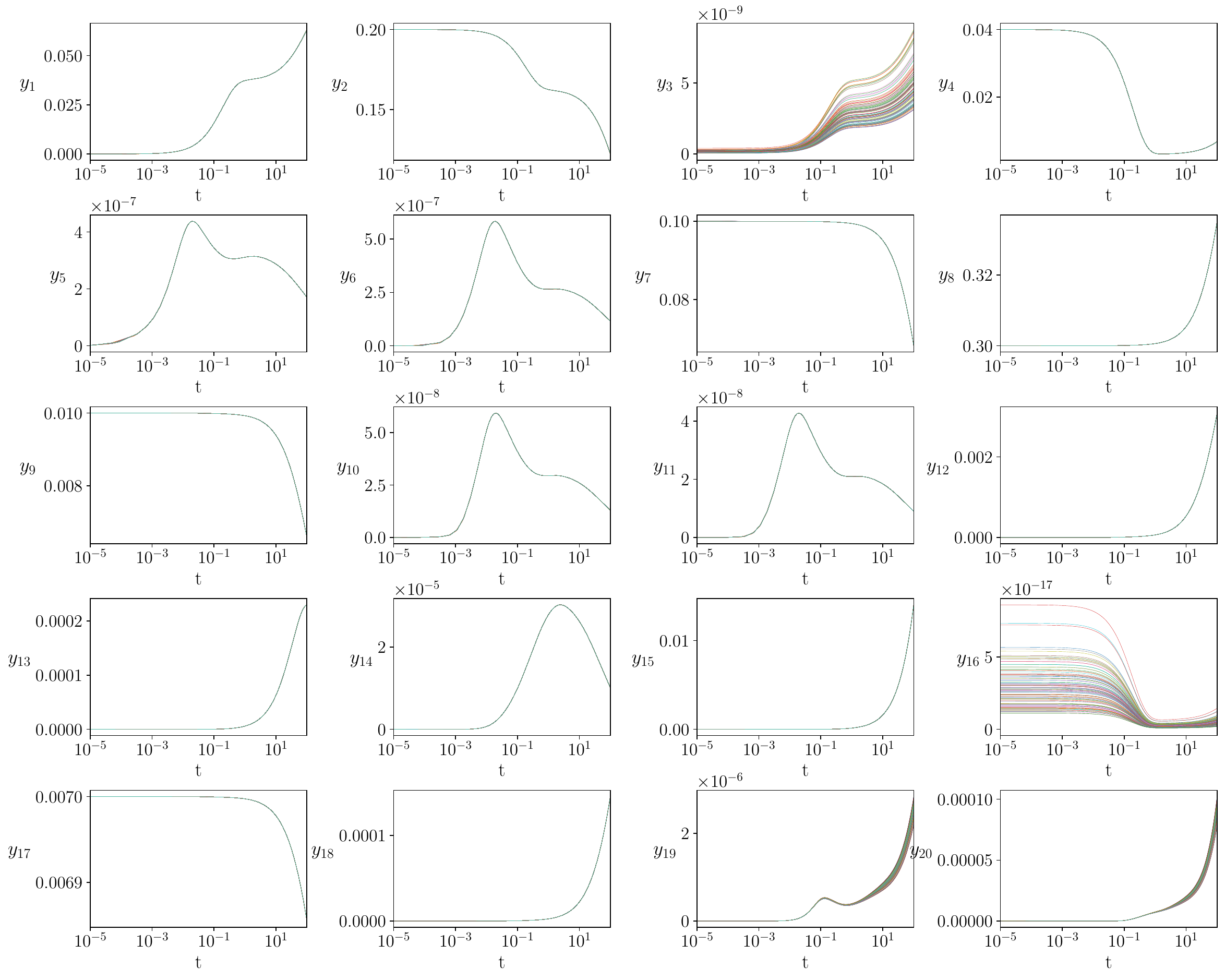}
\caption{Trajectories of the POLLU system obtained by simulating 100 instances of the system corresponding to the nominal values (Table~\ref{tab:params_pollu}) of all parameters except $\{k_{15}, k_{16}, k_{17}, k_{18}, k_{19}, k_{21}\}$, which are instead sampled from the associated prior.}\label{fig:pollu-fix-perturb}
\end{figure}

In order to verify the contribution of each rate parameter in the POLLU problem towards the identifiability of the system, we have fixed all the ``important" rate constants at their nominal values and randomly sampled the ones identified as ``unimportant" through FIM analysis to show that there is not much variability among the trajectories resulting from 100 such simulations. 
Figure~\ref{fig:pollu-fix-perturb} demonstrates this. Observe that the species which have significant variability in their concentrations are present in the reaction mixture in very small amounts compared to the rest. 
Hence the behavior of the system is almost entirely explained by the ``important'' parameters. 

\end{document}